\def\year{2022}\relax
\documentclass[letterpaper]{article} 
\usepackage{aaai22}  
\usepackage{times}  
\usepackage{helvet}  
\usepackage{courier}  
\usepackage[hyphens]{url}  
\usepackage{graphicx} 
\usepackage{natbib}  
\usepackage{caption} 
\DeclareCaptionStyle{ruled}{labelfont=normalfont,labelsep=colon,strut=off} 
\usepackage{algorithm}
\usepackage{algorithmic}
\usepackage{times}
\usepackage{epsfig}
\usepackage{graphicx}
\usepackage{amsmath}
\usepackage{amssymb}
\usepackage{mathrsfs}
\usepackage{bm}
\usepackage{booktabs}
\usepackage{color}
\usepackage{tablefootnote}
\usepackage{epigraph}
\usepackage{multirow}
\usepackage{bbding}
\usepackage{pdfpages}
\usepackage{pifont}
\usepackage{soul}
\newcommand{\cmark}{\ding{51}}
\newcommand{\xmark}{\ding{55}}
  
\usepackage{color}
\usepackage[table]{xcolor}
\definecolor{mygray}{gray}{0.91}
\usepackage{hyperref}
\hypersetup{breaklinks=true}

\usepackage{newfloat}
\usepackage{listings}
\floatstyle{ruled}
\newfloat{listing}{tb}{lst}{}
\floatname{listing}{Listing}

\title{Free Lunch for Co-Saliency Detection: Context Adjustment}

\author{
    Lingdong Kong$^{1}$,
    Prakhar Ganesh$^{2}$,
    Tan Wang$^{1}$,
    Junhao Liu$^{3}$,
    Le Zhang$^{4}$,
    Yao Chen$^{5}$
}

\affiliations{
    \textsuperscript{\rm 1}Nanyang Technological University, Singapore\\
    \textsuperscript{\rm 2}National University of Singapore\\
    \textsuperscript{\rm 3}Shenzhen Institute of Advanced Technology, Chinese Academy of Sciences\\
    \textsuperscript{\rm 4}University of Electronic Science and Technology of China\\
    \textsuperscript{\rm 5}Advanced Digital Sciences Center, University of Illinois\\

    \small {\texttt{\{lingdong001,tan317\}@e.ntu.edu.sg}},~~
    \small {\texttt{pganesh@comp.nus.edu.sg}},\\
    \small {\texttt{jh.liu@siat.ac.cn}},~~
    \small {\texttt{lezhang@uestc.edu.cn}},~~
    \small {\texttt{yao.chen@adsc-create.edu.sg}}
}

\begin{document}

\maketitle

\begin{abstract}
We unveil a long-standing problem in the prevailing co-saliency detection systems: there is indeed inconsistency between training and testing. Constructing a high-quality co-saliency detection dataset involves time-consuming and labor-intensive pixel-level labeling, which has forced most recent works to rely instead on semantic segmentation or saliency detection datasets for training. However, the lack of proper co-saliency and the absence of multiple foreground objects in these datasets can lead to spurious variations and inherent biases learned by models. To tackle this, we introduce the idea of counterfactual training through context adjustment and propose a ``cost-free" group-cut-paste (GCP) procedure to leverage off-the-shelf images and synthesize new samples. Following GCP, we collect a novel dataset called Context Adjustment Training (CAT). CAT consists of 33,500 images, which is four times larger than the current co-saliency detection datasets. All samples are automatically annotated with high-quality mask annotations, object categories, and edge maps. Extensive experiments on recent benchmarks are conducted, show that CAT can improve various state-of-the-art models by a large margin (5$\%$ $\sim$ 25$\%$). We hope that the scale, diversity, and quality of our dataset can benefit researchers in this area and beyond. Our dataset will be publicly accessible through our project page\footnote{\url{http://ldkong.com/data/sets/cat/home}.}.
\end{abstract}

\section{Introduction}
\label{sec:intro}
\setlength{\epigraphwidth}{.9\columnwidth}
\renewcommand{\epigraphflush}{center}
\renewcommand{\textflush}{flushepinormal}


\begin{figure}[t]
\begin{center}
\includegraphics[width=0.474\textwidth]{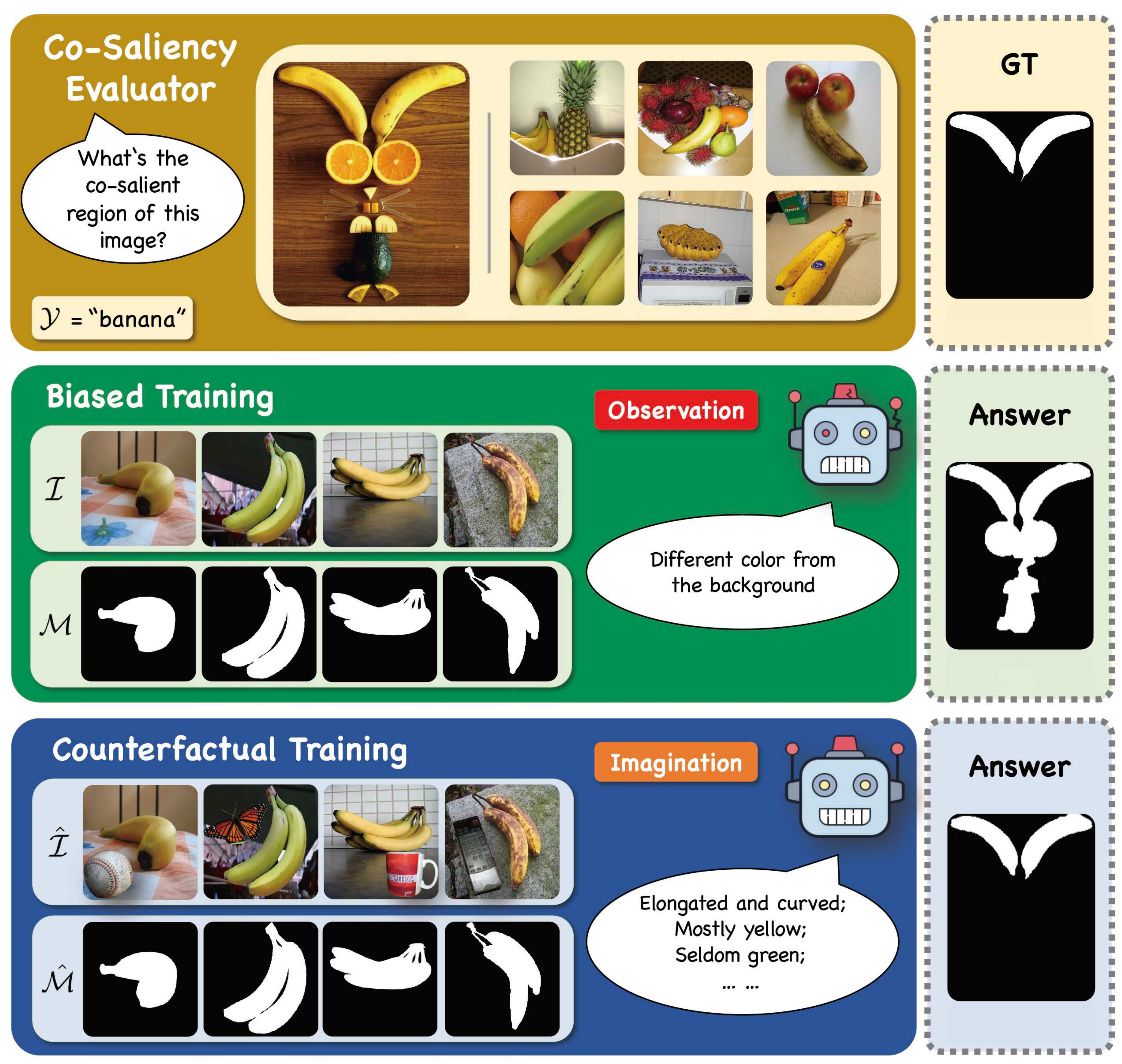}
\end{center}
\vspace{-0.37cm}
\caption{Conceptual illustration of counterfactual training versus biased training. \textbf{Top block}: Evaluations for the \textit{banana} group in \textit{CoSOD3k} \cite{Dataset-CoSOD3k}; \textbf{Middle block}: Biased training samples from current co-saliency detection datasets; \textbf{Bottom block}: Samples generated under the idea of counterfactual training. \textbf{GT} denotes the ground-truth. \textbf{Answers} are generated by \textit{GICD} \cite{Model-GICD}.}
\label{fig:concept}
\end{figure}

Saliency detection attempts to mimic the human visual system by automatically detecting and segmenting out object regions that attract the most attention in an image \cite{Survey-SOD}.
Today's saliency detection systems, especially those equipped with deep neural networks \cite{Survey-SOD-DL} are good at identifying salient objects from images \cite{Survey-SOD}. However, the setting of detecting a single object is less practical. For real-world applications like {image/video retrieval} \cite{Application-Retrieval}, surveillance \cite{Application-Surveillance}, video analysis \cite{Application-Video-Co-localization}, \textit{etc.}, there are always scenes with multiple objects co-occurring in or across frames. 
This motivates our community to explore co-saliency detection \cite{Model-CoSaliency}. 
Sharing a similar rationale with saliency detection, co-saliency detection automatically detects and segments out the common object regions that attract the most attention within image groups \cite{Survey-SOD-DL}. 
Such an imitation can be achieved by the learning paradigm hiding behind artificial neural networks, especially convolutional neural networks \cite{CNN}. However, ``\textit{there is no such thing as a free lunch}". The success of deep learning systems depends heavily on large-scale datasets, which take huge effort and time to collect. The conventional way \cite{Dataset-CoSOD3k,Model-GICD} of constructing a specialized co-saliency detection dataset involves time-consuming and labor-intensive data selection and pixel-level labeling. 
As shown in Figure \ref{fig:concept}, the uni-object distribution of the current training data (middle block) has deviated heavily from that of the evaluation data (top block), which consists of complex context with multiple foreground objects. 
Models encoded with such spurious variations and biases fail to capture the co-salient signals and thus make incorrect predictions \cite{Unbiased-Look-Bias,Scene-Recognition-Bias}.

\begin{figure}[htbp]
\vspace{-0.em}
\begin{minipage}[c]{0.665\linewidth}
\includegraphics[width=1\linewidth]{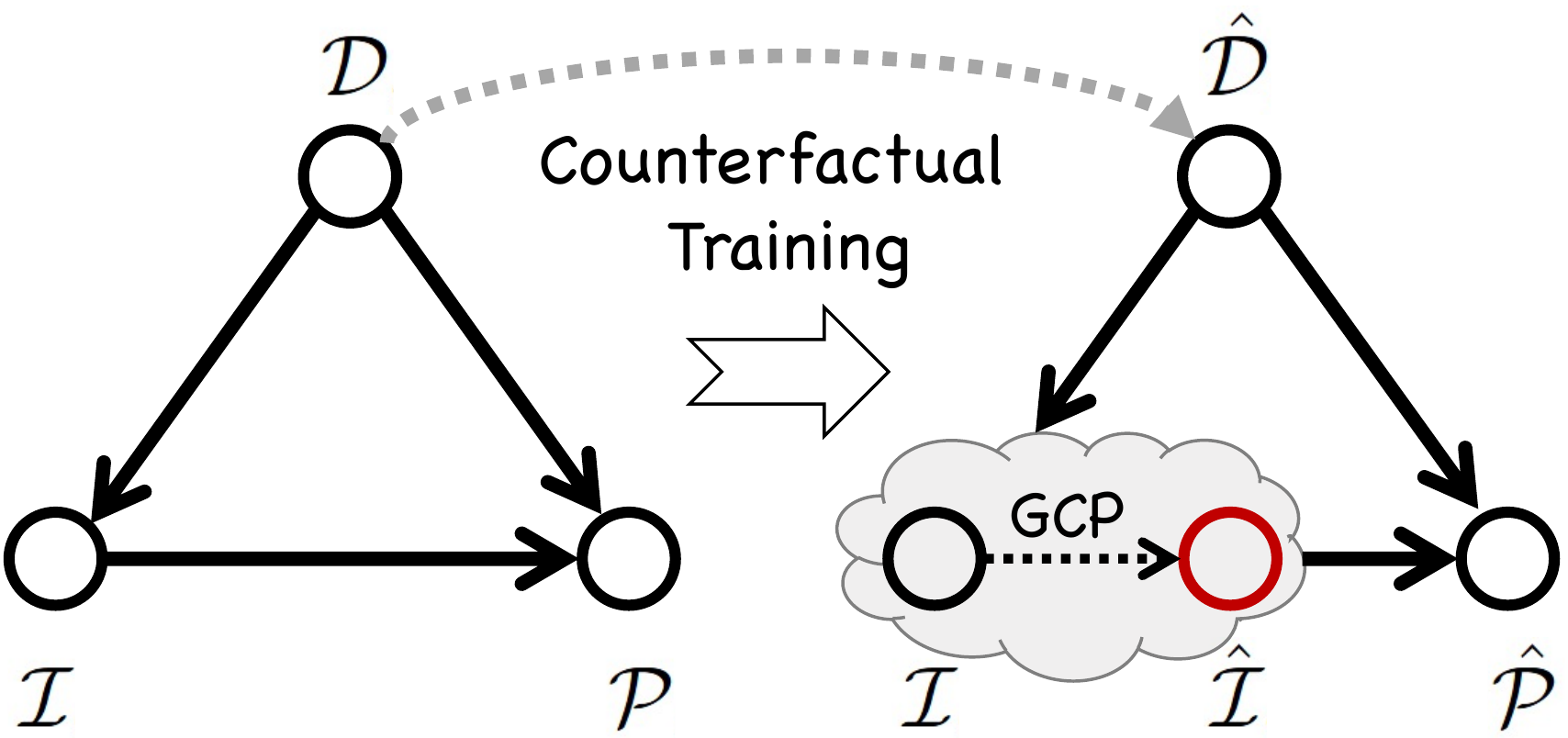}
\end{minipage}\hfill
\begin{minipage}[c]{0.295\linewidth}
\caption{\footnotesize Causal graph for co-saliency detection. Circle nodes denote variables and arrows denote direct causal effects.}
\label{fig:causal_graph}
\end{minipage}
\end{figure}

In the context of co-saliency detection, we denote the training, testing, and true (real-world) distributions as $\mathcal{D}$, $\mathcal{E}$, and $\mathcal{T}$, respectively. Recent works on evaluation \cite{Dataset-CoSOD3k,Model-GICD} make $\mathcal{E}$ one-step closer towards $\mathcal{T}$ by identifying co-salient signals from multi-foreground clusters in a group-by-group manner. However, a dataset that can well-define $\mathcal{D}$ is still missing. As we will discuss in the next section, most recent works borrow semantic segmentation datasets like \textit{MS-COCO} \cite{Dataset-COCO} and saliency detection datasets like \textit{DUTS} \cite{Dataset-DUTS} to train their models, which exacerbate the inconsistency among $\mathcal{D}$, $\mathcal{E}$, and $\mathcal{T}$. Not surprisingly, as models only see naive examples during training, they will inevitably cater to the seen idiosyncrasies, and thus are stuck in the unseen world with biased assumptions. In this paper, we aim at finding a ``cost-free" way to handle the distribution inconsistency in co-saliency detection. Intrigued by causal effect \cite{Causality-Why,Causality-Primer} and its extensions in vision \& language \cite{Causality-VC-RCNN,Causality-CFVQA,Causality-IFSL}, we introduce \textbf{counterfactual training} with regard to the gap between current training distribution $\mathcal{D}$ and true distribution $\mathcal{T}$ as the direct cause \cite{Causality-Effect,Causality-Long-Tailed} of incorrect co-saliency predictions. As shown in Figure \ref{fig:causal_graph}, the quality of prediction $\mathcal{P}$ made by a learning-based model is dependent on the quality of input data $\mathcal{I}$ under distribution $\mathcal{D}$. The goal of counterfactual training is to synthesize ``imaginary" data sample $\hat{\mathcal{I}}$, whose distribution $\hat{\mathcal{D}}$ --- also originates from $\mathcal{D}$ --- can mimic $\mathcal{T}$. In this way, models can capture the true effect in terms of co-saliency from $\hat{\mathcal{I}}$ and make prediction $\hat{\mathcal{P}}$ properly.

Under the instruction of counterfactual training, we propose using context adjustment \cite{Aug-Empirical-Study} to augment off-the-shelf saliency detection datasets, and introduce a novel \textbf{group-cut-paste (GCP)} procedure to improve the distribution of the training dataset. Taking a closer look at Figure \ref{fig:concept}. $\mathcal{M}$ and $\mathcal{Y}$ are the corresponding pixel-level annotation and category label of sample $\mathcal{I}$. GCP turns $\mathcal{I}$ into a canvas to be completed and paint the remaining part through the following steps: (1) classifying candidate images into a semantic group $\mathcal{Z}$ (e.g., \textit{banana}) by reliable pretrained models; (2) cutting out candidate objects (e.g., \textit{baseball}, \textit{butterfly}, etc.); and (3) pasting candidate objects into image samples. We will revisit this procedure formally in the following sections. Different from the plain ``observation" (e.g., \textit{different color from the background}) made by biased training, counterfactual training opens the door of ``imagination" and allows models to think comprehensively \cite{Causality-CFVQA}. A better prediction can be made possibly because of features, such as the \textit{elongated and curved} shape (instead of the \textit{round} shape of \textit{baseball} and \textit{orange}) or \textit{yellow-green} color (instead of the \textit{dark} color of \textit{remote control} and \textit{avocado}), are captured. In this way, models can focus more on the true causal effects rather than spurious variations and biases caused by the distribution gap \cite{Causality-CSS}.

Following GCP, we collect a novel dataset called \textbf{Context Adjustment Training (CAT)}. Our dataset consists of $280$ subclasses affiliated to $15$ superclasses, which cover common items in daily life, as well as animals and plants in nature. While CAT is diverse in semantics, it also has a large scale. A total number of 33,500 training samples making it the current largest in co-saliency detection. Every sample in CAT is equipped with sophisticated mask annotation, category, and edge information. It is worth noting that, unlike expensive data selection and pixel-by-pixel labeling, all the images and their corresponding masks in our dataset are automatically synthesized and annotated, making the cost virtually ``free". Extensive experiments verify the effectiveness of our dataset. Without bells and whistles, CAT helps both one-stage and two-stage models to achieve significant improvements for on average 5$\%$ $\sim$ 25$\%$ in conventionally-adopted metrics on the challenging evaluation datasets \textit{CoSOD3k} \cite{Dataset-CoSOD3k} and \textit{CoCA} \cite{Model-GICD}.
\section{Related Work}
\label{sec:related_work}

\begin{table*}[t]
\centering
\scalebox{0.803}{
\begin{tabular}{p{0.4cm}<{\centering}|p{2cm}<{\raggedleft}|p{1.18cm}<{\centering}|p{1.18cm}<{\centering}p{1.18cm}<{\centering}p{1.18cm}<{\centering}p{1.18cm}<{\centering}p{1.18cm}<{\centering}|p{0.8cm}<{\centering}p{0.8cm}<{\centering}p{0.8cm}<{\centering}p{0.8cm}<{\centering}|p{1cm}<{\centering}|p{2.3cm}<{\centering}}
\toprule[1.25pt]
\textbf{\#} & \textbf{Dataset}  & \textbf{Year} & \textbf{\#Img.} & \textbf{\#Cat.} & \textbf{\#Avg.} & \textbf{\#Max.} & \textbf{\#Min.} & \textbf{Mul.} & \textbf{Sal.} & \textbf{Larg.} & \textbf{H.Q.} & \textbf{Type} & \textbf{Inputs}
\\ \hline\hline
1 & MSRC & 2005 & 240 & 8 & 30.0 & 30 & 30 & \xmark & \cmark & \xmark & \xmark & CO & Group images
\\
2 &iCoseg  & 2010 & 643 & 38 & 16.9 & 41 & 4 & \xmark & \cmark & \xmark & \cmark & CO & Group images
\\
3 & CoSal2015 & 2015 & 2,015 & 50 & 40.3 & 52 & 26 & \cmark & \cmark & \xmark & \cmark & CO & Group images
\\
4 & DUTS-TR & 2017 & 10,553 & - & -  & - & - & \xmark & \cmark & \cmark & \cmark & SD & Single image
\\
5 & COCO9213 & 2017 & 9,213 & 65 & 141.7 & 468 & 18 & \cmark & \cmark & \xmark & \xmark & SS & Group images
\\
6 & COCO-GWD & 2019 & 9,000 & 118 & 76.2 & - & - & \cmark & \cmark & \xmark & \xmark & SS & Group images
\\
7 & COCO-SEG & 2019 & 200,932 & 78 & 2576.1  & 49,355 & 201 & \cmark  & \xmark & \cmark & \xmark & SS & Group images
\\
8 & WISD & 2019 & 2,019 & - & - & - & - & \cmark & - & \xmark & - & CO & Single image
\\
9 & DUTS-Class & 2020 & 8,250 & 291 & 28.3 & 252 & 5 & \xmark & \cmark & \cmark & \cmark & CO & Group images
\\ \hline
10 & \textbf{CAT (Ours)} & \textbf{2021} & \textbf{33,500} & \textbf{280} & \textbf{119.6} & \textbf{824} & \textbf{28} & \cmark & \cmark & \cmark & \cmark & \textbf{CO} & \textbf{Group images}
\\ 
\bottomrule[1.25pt]
\end{tabular}}
\vspace{-0.2cm}
\caption{Datasets used for training co-saliency detection models. \textbf{\#Img.}: Number of images; \textbf{\#Cat.}: Number of categories; \textbf{\#Avg.}: Average number of images per category; \textbf{\#Max.}: Maximum number of images per group; \textbf{\#Min.}: Minimum number of images per group. \textbf{Mul.}: Whether contains multiple foreground objects or not; \textbf{Sal.}: Whether maintains saliency or not; \textbf{Larg.}: Whether large-scale (more than 10k images) or not; \textbf{H.Q.}: Whether has high-quality annotations or not. \textbf{CO}: Co-saliency detection dataset; \textbf{SD}: Saliency detection dataset; \textbf{SS}: Semantic segmentation dataset. ``-" denotes ``not available".}
\label{tab:training_datasets}
\end{table*}

\noindent\textbf{Context Adjustment.}
Modeling visual contexts liberates a lot of computer vision tasks from the impediment caused by the need for sophisticated annotations \cite{Aug-Cut-Paste-Learn}, such as bounding boxes and pixel-level masks. \cite{Aug-CutMix} leverages training pixels and regularization effects of regional dropout by cutting and pasting image patches. To avoid overfitting, \cite{Aug-CutOut} randomly masks out square regions of an image during training. The generic vicinal distribution introduced in \cite{Aug-MixUp} helps to synthesize the groundtruth of new samples by the linear interpolation of one-hot labels. Unfortunately, the patches introduced by both \cite{Aug-CutMix} and \cite{Aug-CutOut} may cause severe occlusions for the original objects, which is not compatible with the idea of saliency detection \cite{Saliency}. For \cite{Aug-MixUp}, the local statistics and semantic are destroyed. Since the foreground and background are blended, the saliency cannot be appropriately defined anymore. \cite{Aug-AugMix} fixes this problem by mixing several transformed versions of an image together, i.e., augmentation chain. Although such a composition improves the model robustness, it cannot introduce co-salient signals. \cite{Model-GICD} concatenates two samples together to form a multi-foreground image. However, for backbones that require fixed-size inputs \cite{VGG}, the resize of such jigsawed images leads to severe shape distortion. In contrast, our GCP preserves both the multi-foreground requirement and object shapes and maintains saliency \cite{Saliency}.

\noindent\textbf{Co-Saliency Detection.}
The origin of this task can be dated back to the last decade \cite{Model-CoSaliency}, where co-saliency was defined as the regions of an object which occurs in a pair of images \cite{Dataset-ImagePair}. A more formal definition given in \cite{Model-MI} exploits both intra- and inter-image saliency in a group-by-group manner. Since then, researchers in this area have been working on identifying co-salient signals across image groups \cite{Dataset-THUR-15K}. Representative works from early years  
\cite{Model-PCSD,Model-CBCS,Model-CSHS} rely on certain heuristics like background priors \cite{Model-MVSRCC} and color contrasts \cite{Model-MCFM}. Let alone the post-processing like CRF \cite{CRF}, most modern co-saliency detection models can be divided into two-stage and one-stage models. Building upon saliency detection frameworks, two-stage models leverage the intra-saliency with inter-saliency generated by specially designed modules \cite{Model-ICNet,Model-CoEGNet}. There are also some one-stage models which do not require saliency preparation. \cite{Model-CoADNet}, \cite{Model-GICD}, and \cite{Model-GCoNet} introduce consensus embedding procedures to explore group representation and use it to guide soft attentions \cite{CAM}. Associating with our dataset, both the state-of-the-art two-stage and one-stage models achieve better performances.

\noindent\textbf{Datasets.}
Due to the lack of a suitable training dataset, most previous works use existing saliency detection datasets or semantic segmentation datasets for training. Table \ref{tab:training_datasets} summarizes datasets adopted to train co-saliency detection models. Small datasets  \cite{Dataset-MSRC,Dataset-iCoseg,Dataset-Coseg-Rep} are popular among works in early years only. Some large-scale saliency detection datasets \cite{Dataset-DUT-OMRON,Dataset-MSRA10K,Dataset-DUTS} are adopted by two-stage models to extract intra-saliency cues. However, such datasets do not have class information and multiple foreground objects, hence it is impossible for them to train end-to-end co-saliency detection models. Some works even borrow datasets from semantic segmentation, e.g., \textit{COCO-SEG} \cite{Dataset-COCO-SEG} and \textit{COCO9213} \cite{Dataset-COCO}. Unfortunately, although such datasets are large in scales, their annotations are coarse. Recently, \textit{DUTS-Class} \cite{Model-GICD} and its jigsawed version \cite{Model-GICD} have been proposed as a ``transition plan" towards co-saliency. However, the former is just a ``grouped" saliency detection dataset, while the latter introduces issues like shape distortion and independent boundaries. Evidence shows that these inappropriate training paradigms have caused serious biases for co-saliency detection models \cite{Dataset-CoSOD3k,Model-GICD}. Our dataset addresses the aforementioned problems. CAT is currently the largest co-saliency detection dataset and offers diverse semantics and high-quality annotations.


\begin{figure*}[!thp]
\begin{center}
\includegraphics[width=0.9999\textwidth]{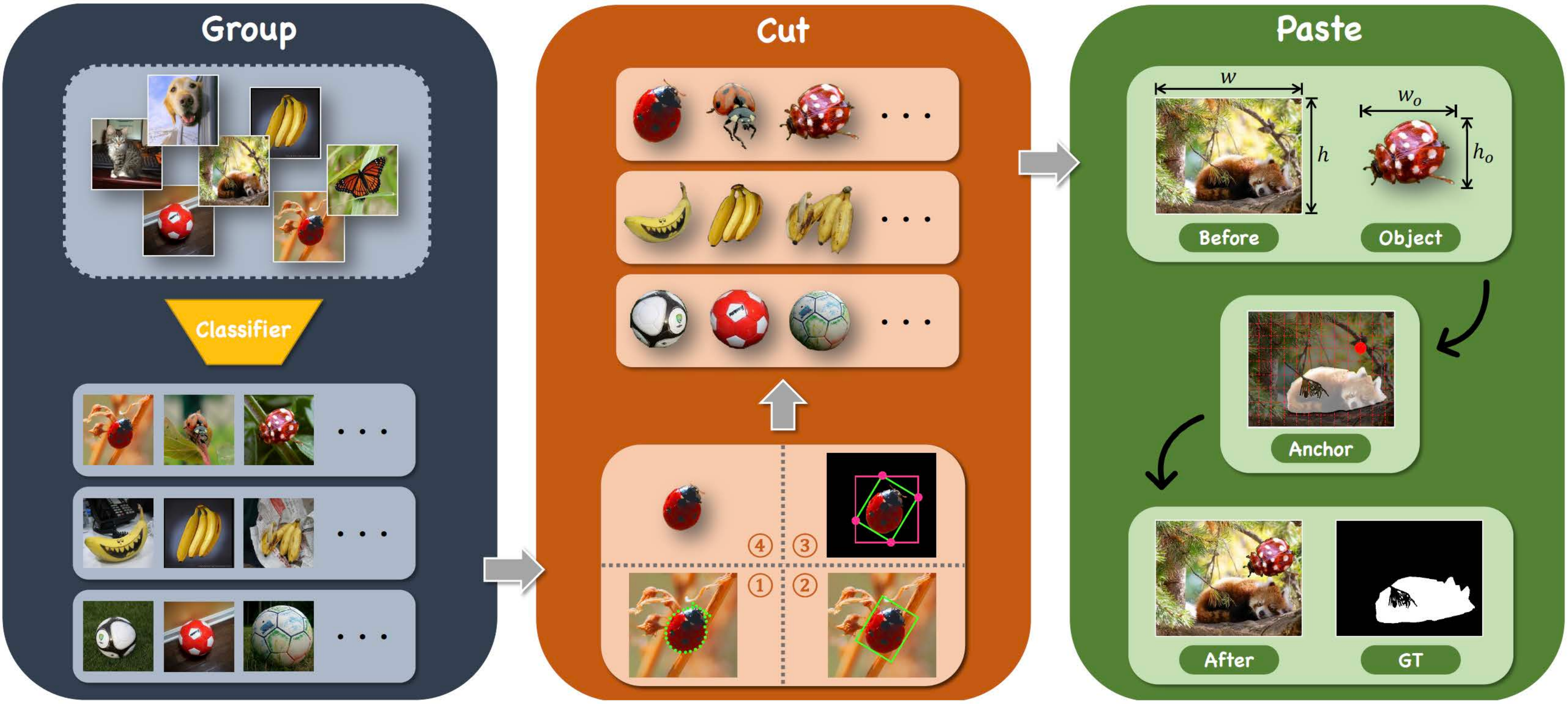}
\end{center}
\vspace{-0.3cm}
\caption{Overview of our group-cut-paste (GCP) procedure. \textbf{Left block}: Classifier $\mathcal{F(\cdot)}$ generates category label $\mathcal{Y}$ for all raw images and ``group" them accordingly; \textbf{Middle block}: Object $o$ is ``cut" out from the original image based on contours $\{(p, q)_{n}\}$ and bounding boxes; \textbf{Right block}: New sample $\hat{\mathcal{I}}$ is synthesized by ``pasting" randomly sampled object $o$ into canvas $\mathcal{I}$.}
\label{fig:gcp}
\end{figure*}
\section{Proposed Method} 
\label{sec:proposed_method}

\begin{figure}[t]
\begin{center}
\includegraphics[width=0.47\textwidth]{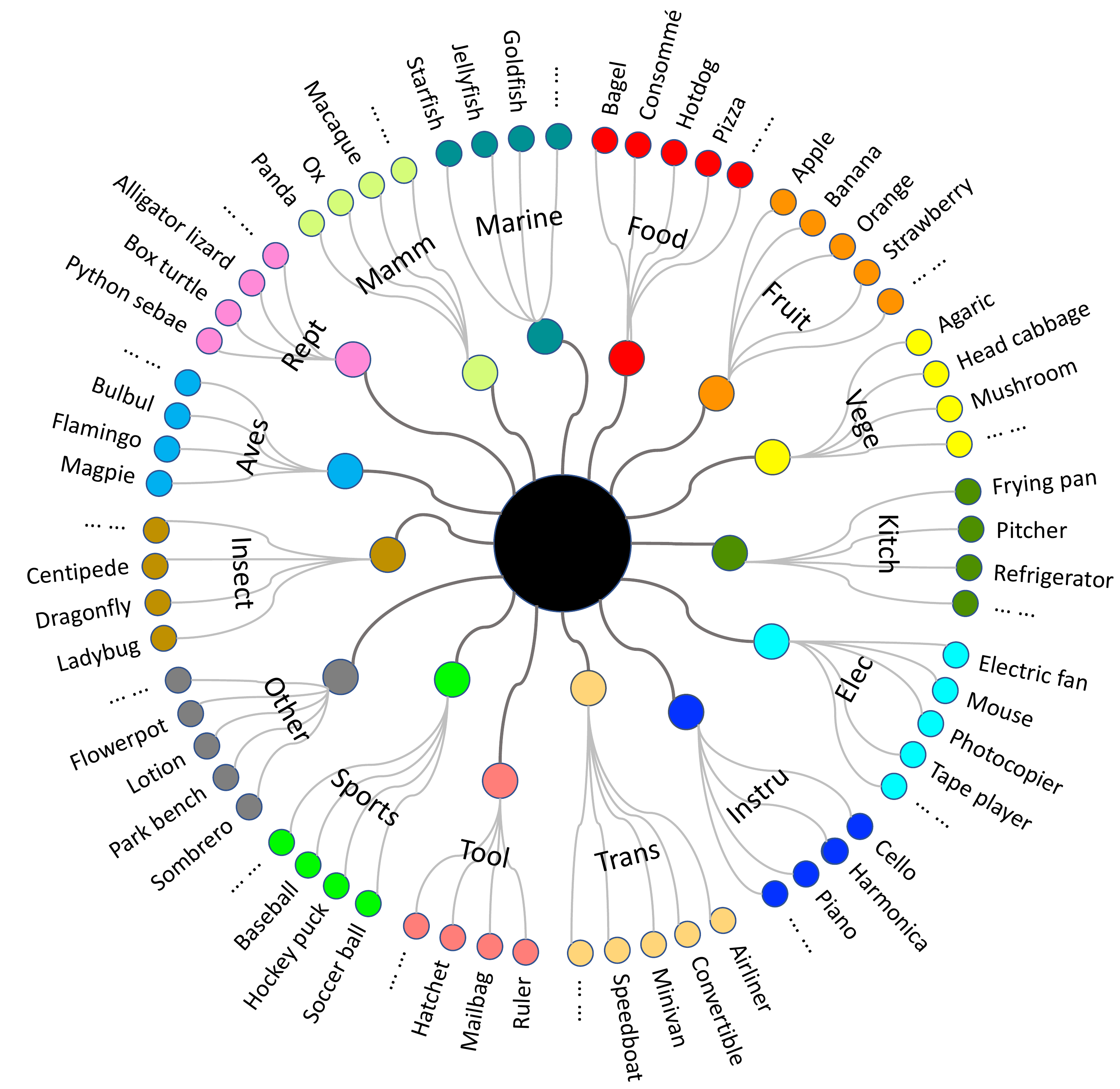}
\end{center}
\vspace{-0.37cm}
\caption{The taxonomic illustration for our dataset, of which consists of 15 superclasses and 280 subclasses. Zoom in for details.}
\label{fig:superclasses}
\end{figure}

\begin{figure}[!t]
\begin{center}
\includegraphics[width=0.473\textwidth]{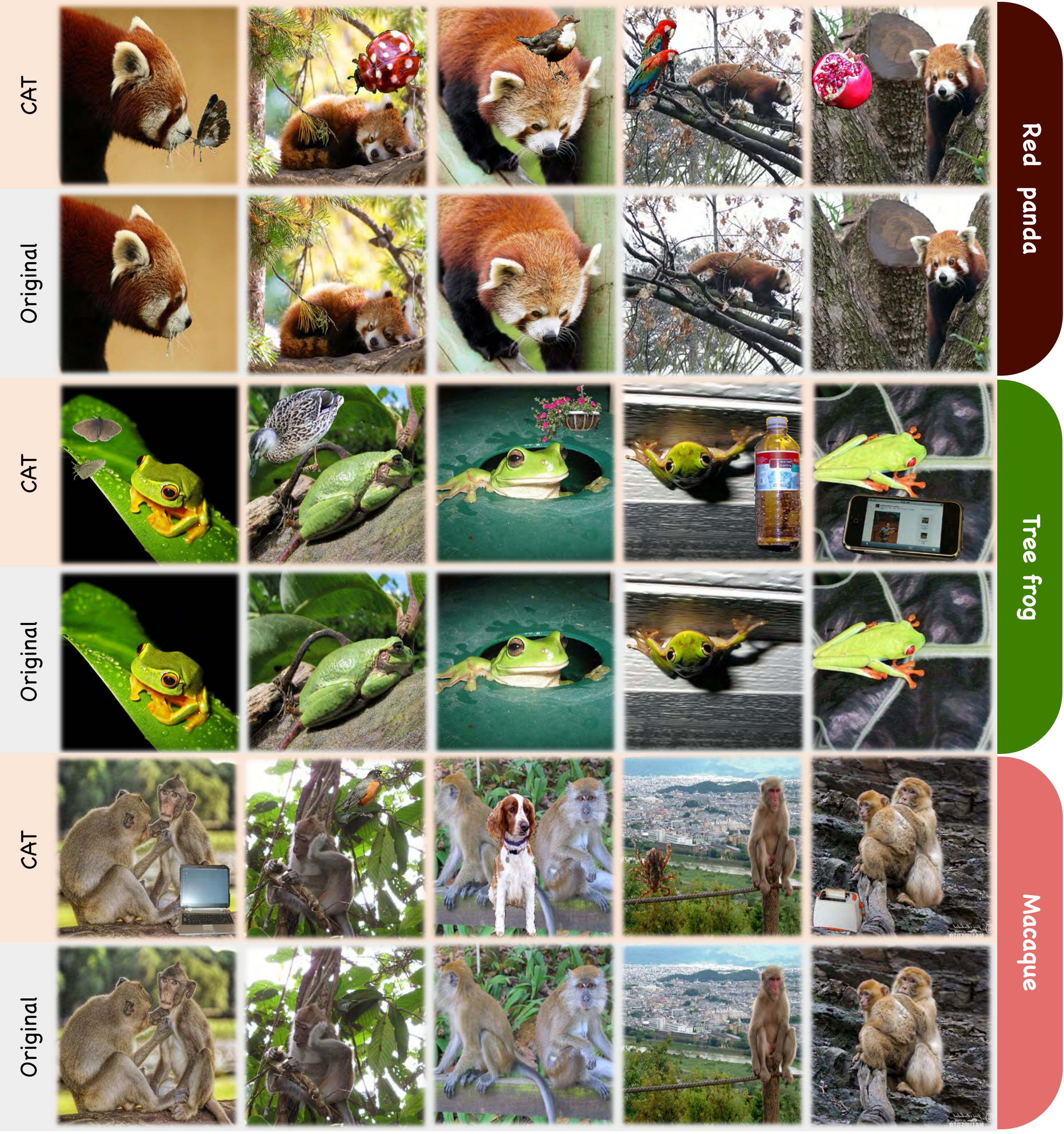}
\end{center}
\vspace{-0.37cm}
\caption{Samples from our dataset. Zoom in for details.}
\label{fig:example}
\vspace{-0.2cm}
\end{figure}

In this section, we first introduce the idea of counterfactual training in co-saliency detection. Under this high-level instruction, we propose a group-cut-paste (GCP) procedure to adjust visual context and generate training samples. Finally, we construct a novel dataset by following GCP.

\subsection{Counterfactual Training}
Let $\mathcal{I}\in\mathbb{R}^{w\times h\times c}$ and $\mathcal{M}\in\mathbb{R}^{w\times h}$ denote a training image and its mask annotation with width $w$, height $h$, and channel $c$. $\mathcal{Z}$ is an image group whose members include $\mathcal{I}$. $\mathcal{Y}$ denotes the category label which contains the semantic information for both $\mathcal{Z}$ and $\mathcal{I}$. The counterfactual \cite{Causality} $\hat{\mathcal{I}}=\mathcal{I}_{\mathcal{M}}[\hat{\mathcal{Z}}(\hat{\mathcal{I}})]$ of sample $\mathcal{I}$, is read as:

\begin{center}
\fbox
{\shortstack[l]{
    $~~~~~~~~~$$\mathcal{I}$ would be $\hat{\mathcal{I}}$, had $\mathcal{M}$ been $\hat{\mathcal{M}}$,$~~~$\\
    $~~~~~~~$given the fact that $\mathcal{Z}=\hat{\mathcal{Z}}(\mathcal{I}=\hat{\mathcal{I}})$.$~~~~~$}
}
\end{center}
where $\hat{\mathcal{Z}}\supset\hat{\mathcal{I}}$ denotes the new image group; New mask is restricted as $\hat{\mathcal{M}}\approx \mathcal{M}$ to maintain the saliency \cite{Saliency} of the co-occurring object. In other words, label $\mathcal{Y}$ will not change during the whole process. The given ``fact" $\mathcal{Z}=\hat{\mathcal{Z}}(\hat{\mathcal{I}})$ means that the group training paradigm remains the same, while the ``counter-fact" (what if) $\mathcal{M}\neq \hat{\mathcal{M}}(\hat{\mathcal{I}})$ is indicating that the generated $\hat{\mathcal{M}}(\hat{\mathcal{I}})$ clashes with $\hat{\mathcal{M}}$.

The conceptual meanings of counterfactual training can be substantiated by the following instructions \cite{Causality-Primer}:

\noindent\textbf{Abduction} - ``\textit{Given the fact that $\mathcal{Z}=\hat{\mathcal{Z}}(\mathcal{I}=\hat{\mathcal{I}})$}". When constructing a new image group $\hat{\mathcal{Z}}$ for $\mathcal{Z}$, the co-saliency should remain unchanged. That is to say, the ``fact" that the object in $\mathcal{I}\subset \mathcal{Z}$ is salient still holds for it in $\hat{\mathcal{I}}\subset \hat{\mathcal{Z}}$.

\noindent\textbf{Action} - ``\textit{Had $\mathcal{M}$ been $\hat{\mathcal{M}}$}". After the ``imagination", the mask annotation $\hat{\mathcal{M}}$ should not have a significant change. Here we intervene $\mathcal{M}$ by keeping the semantic label $\mathcal{Y}$ the same both beforehand and afterhand.

\noindent\textbf{Prediction} - ``\textit{$\mathcal{I}$ would be $\hat{\mathcal{I}}$}". Conditioning on the ``fact" $\mathcal{Z}=\hat{\mathcal{Z}}(\hat{\mathcal{I}})$ and the intervention objective 
$\mathcal{M}\approx \hat{\mathcal{M}}$, we can generate the counterfactual sample $\hat{\mathcal{I}}$ and its mask $\hat{\mathcal{M}}$ from the following probability distribution function: 
$P(\mathcal{I}\mid\mathcal{Z}=\hat{\mathcal{Z}}(\hat{\mathcal{I}}), \mathcal{M}\approx \hat{\mathcal{M}})$.

\subsection{Group-Cut-Paste}

Following the counterfactual training instructions, we design the group-cut-paste (GCP) procedure. Specifically, the object map $\mathcal{O}\in\mathbb{R}^{w\times h\times c}$ for image $\mathcal{I}$ can be generated via $\mathcal{M}$, i.e., $\mathcal{O} =\mathcal{I} \odot \mathcal{M}$, where $\odot$ denote element-wise multiplication. The goal of GCP is to automatically generate new training sample $(\hat{\mathcal{I}}, \hat{\mathcal{M}})$ by combing $\mathcal{I}_{a}$ and $\mathcal{M}_{a}$ with $\mathcal{O}_{b}$, where $a$ and $b$ are category indexes of the target and source groups, respectively. The generated sample $(\hat{\mathcal{I}}, \hat{\mathcal{M}})$ is then used to train candidate models with their original loss functions. Figure \ref{fig:gcp} gives an overview of our method.

\noindent\textbf{Group.}
The first step is to build image groups. To effectively classify candidate images, we adopt one of the current state-of-the-art classifiers \textit{WSL-Images} \cite{Pretrain-WSL-Images}, which achieves $85.4\%$ Top-1 accuracy on \textit{ImageNet} \cite{Dataset-ImageNet}. The category label $\mathcal{Y}$ can be generated via $\mathcal{Y} = \mathcal{F}(\mathcal{I})$, where $\mathcal{F(\cdot)}$ denotes the pretrained classifier. We manually pick out the misclassified examples after grouping. This results in $\{(\mathcal{I}, \mathcal{Y}), \mathcal{M}\}$ sample pairs distributed in $z$ semantic groups. 

\noindent\textbf{Cut.}
Both object map $\mathcal{O}$ and mask $\mathcal{M}$ are used to prepare candidate object $o$. We adopt the \textit{border following algorithm} \cite{Board-Following-Algorithm} to extract $n$ external contour points $\{(p, q)_{n}\}$ by mask $\mathcal{M}$. Based on contours, the bounding rectangle with minimum area can be drawn. The area to be cut can then be defined by the four vertices of this rotated box. This gives us the final object $o\in\mathbb{R}^{w_{o}\times h_{o}\times c}$.

\noindent\textbf{Paste.}
Under the counterfactual training instructions, object $o$ can be pasted properly into $\mathcal{I}$ to form a new sample $\hat{\mathcal{I}}$. Here we maintain the ``fact" by preserving group consistency, i.e., conduct this procedure in a group-by-group manner. Denote the regions outside $\mathcal{M}$ as $\bar{\mathcal{M}}$. To avoid severe occlusion, coordinate $(x, y)$ is randomly sampled in $\bar{\mathcal{M}}$ as the position to be pasted. The center of $o$ is chosen as the anchor for pasting. We execute the ``action" by restricting the size of candidate object $o$ and maintaining label $\mathcal{Y}$ before and after paste. The mask annotations $\hat{\mathcal{M}}$ are automatically synthesized by dyadic operations between $o$ and $\mathcal{M}$. Finally, under uniform probability distribution $P=1/(z-1)$,  ``prediction" gives us $\{(\hat{\mathcal{I}}, \mathcal{Y}), \hat{\mathcal{M}\}}$ pairs for further model training. Our GCP operation can be easily applied on CPUs in parallel with the main GPU training tasks, which hides the computation cost and boosts the performance for virtually free.

\subsection{Constructing the Dataset}
\begin{figure}[t]
\begin{center}
\includegraphics[width=0.474\textwidth]{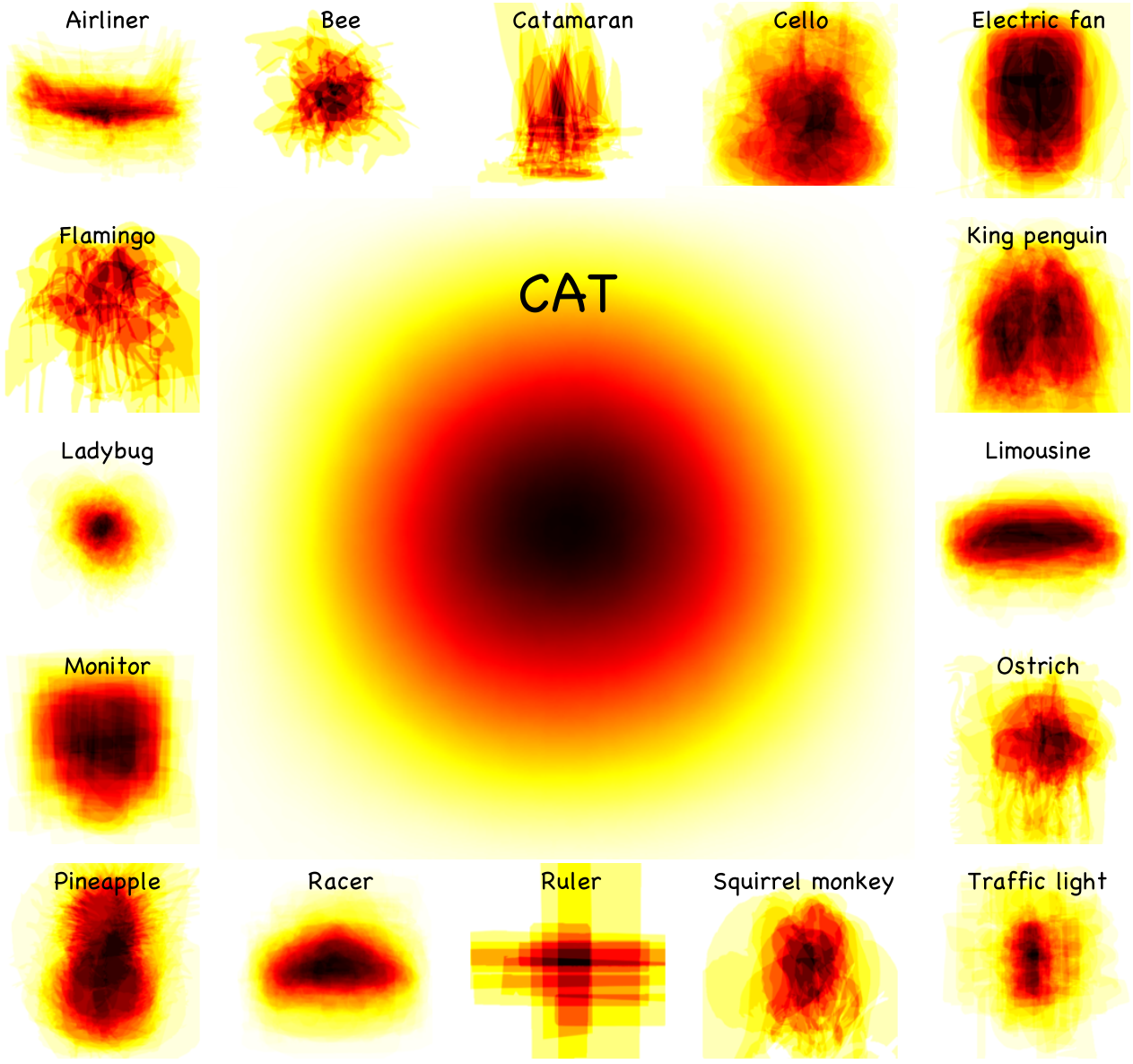}
\end{center}
\vspace{-0.37cm}
\caption{Patterns for representative groups and the overall dataset. Best viewed in color and zoomed-in for details.}
\label{fig:patterns}
\end{figure}

For the purpose of stabilization and to ensure high quality, we collect a dataset called Context Adjustment Training (CAT). We pick 8,375 images with clear saliency and sophisticated annotations as our canvas from existing saliency detection datasets \cite{Dataset-DUTS,Saliency}. Following GCP, 280 semantic groups affiliated to 15 superclasses are built after the ``\textit{grouping}", i.e., aves, electronic product (elec.), food, fruit, insect, instrument (instru.), kitchenware (kitch.), mammal (mamm.), marine, other, reptile (rept.), sports, tool, transportation (trans.), and vegetable (vege.). See Figure \ref{fig:superclasses} for the taxonomy. We then ``\textit{cut}" all the objects $o$ out and discard those with incomplete shapes. During the ``\textit{paste}" stage, the object size is restricted as $\frac{w_{o}}{w}=\frac{h_{o}}{h}=(0.1 \sim 0.8)$. Candidate objects are randomly flipped horizontally to increase diversity. We sample twice for each candidate image to generate two samples and do a re-sample for unsatisfied cases. This gives us 16,750 augmented images. Their corresponding masks are automatically generated. As shown in Figure \ref{fig:example}, samples synthesized by counterfactual training and GCP can model realistic visual context and offer proper co-salient signals. Following \cite{Model-GICD}, we expand the scale of our dataset by supplementing twice as many samples without augmentation. In total, CAT consists of 33,500 images, which is ten times and four times bigger than the current largest co-saliency detection evaluation set and training set, respectively.  See Table \ref{tab:training_datasets} for more dataset statistics. Several group patterns (calculated by averaging overlapping masks) of our dataset are shown in Figure \ref{fig:patterns}. The overall pattern of our dataset tends to be a ``round" shape, while categories with unique shapes (e.g., \textit{cello} and \textit{limousine}) result in shape-biased patterns, which is consistent with the definition of saliency \cite{Saliency}. More details of our dataset can be found in the supplementary material.

\section{Benchmark Experiment}
\label{sec:experiment}
In this section, we conduct a comprehensive benchmark study for co-saliency detection to verify the effectiveness and superiority of our proposed methods and dataset.

\subsection{Implementation Details}

\noindent\textbf{Competitors.}
We compare our dataset with six representative datasets and four popular augmentation techniques, i.e., \textit{COCO-SEG} \cite{Dataset-COCO-SEG}, \textit{DUTS-TR} \cite{Dataset-DUTS}, \textit{COCO9213} \cite{Dataset-COCO}, \textit{CoSal2015} \cite{Dataset-CoSal2015}, \textit{DUTS-Class} \cite{Model-GICD}, \textit{AutoAug} \cite{AutoAug}, \textit{CutOut} \cite{Aug-CutOut}, \textit{CutMix} \cite{CutMix}, and \textit{Jigsaw} \cite{Model-GICD}.

\noindent\textbf{Models.}
Five state-of-the-art models with open-sourced code are selected for our benchmark experiments, i.e., \textit{PoolNet} \cite{Model-PoolNet}, \textit{EGNet} \cite{Model-EGNet}, \textit{ICNet} \cite{Model-ICNet}, \textit{GICD} \cite{Model-GICD}, and \textit{GCoNet} \cite{Model-GCoNet}. All training configurations w.r.t. model architectures, hyperparameters, loss functions, and other tricks are kept default except for learning rate. For each dataset, we tune the learning rate a bit with multiple runs to ensure the convergence and report the average results to avoid extreme cases. NVIDIA GeForce RTX 2080Ti graphics cards are used through our experiments. Due to space limits, please refer to the supplementary for more configuration details.

\begin{table}[t]
\centering
\scalebox{0.66}{
\begin{tabular}{c|cccc|cccc}
\toprule[1pt]
\small{Dataset} & \small{\textit{MAE} $\downarrow$} & \small{$F_{max}$ $\uparrow$} & \small{$E_{avg}$ $\uparrow$} & \small{$S_{\alpha}$ $\uparrow$} & \small{\textit{MAE} $\downarrow$} & \small{$F_{max}$ $\uparrow$} & \small{$E_{avg}$ $\uparrow$} & \small{$S_{\alpha}$ $\uparrow$}
\\ \hline\hline
\small{CoSal2015} & 0.181 & 0.510 & 0.629 & 0.578 & 0.199 & 0.315 & 0.577 & 0.508
\\
\small{DUTS-TR}$^{\ddag}$ & 0.209 & 0.518 & 0.646 & 0.620 & 0.235 & 0.299 & 0.558 & 0.525
\\
\small{DUTS-Class} & 0.205 & 0.508 & 0.649 & 0.591 & 0.210 & 0.323 & 0.594 & 0.535
\\
\small{COCO9213} & 0.225 & 0.488 & 0.639 & 0.601 & 0.256 & 0.283 & 0.541 & 0.509
\\
\cellcolor{mygray}\small{\textbf{CAT}} & \cellcolor{mygray}\textbf{0.171} & \cellcolor{mygray}\textbf{0.561} & \cellcolor{mygray}\textbf{0.689} & \cellcolor{mygray}\textbf{0.636} & \cellcolor{mygray}\textbf{0.181} & \cellcolor{mygray}\textbf{0.346} & \cellcolor{mygray}\textbf{0.625} & \cellcolor{mygray}\textbf{0.561}
\\
\bottomrule[1pt]
\end{tabular}}
\vspace{-0.25cm}
\caption{Benchmarking results of different datasets trained by \textit{PoolNet} \cite{Model-PoolNet} and evaluated on \textit{CoSOD3k} ({left}) and \textit{CoCA} ({right}).}
\label{tab:poolnet}
\end{table}

\begin{table}[t]
\centering
\scalebox{0.66}{
\begin{tabular}{c|cccc|cccc}
\toprule[1pt]
\small{Dataset} & \small{\textit{MAE} $\downarrow$} & \small{$F_{max}$ $\uparrow$} & \small{$E_{avg}$ $\uparrow$} & \small{$S_{\alpha}$ $\uparrow$} & \small{\textit{MAE} $\downarrow$} & \small{$F_{max}$ $\uparrow$} & \small{$E_{avg}$ $\uparrow$} & \small{$S_{\alpha}$ $\uparrow$}
\\ \hline\hline
\small{CoSal2015} & 0.103 & 0.698 & 0.791 & 0.754 & 0.159 & 0.430 & 0.649 & 0.615
\\
\small{DUTS-TR}$^{\ddag}$ & 0.118 & 0.696 & 0.771 & 0.760 & 0.182 & 0.425 & 0.610 & 0.604
\\
\small{DUTS-Class} & 0.113 & 0.704 & 0.793 & 0.770 & 0.166 & 0.440 & 0.640 & 0.620
\\
\small{COCO9213} & 0.122 & 0.702 & 0.781 & 0.762 & 0.178 & 0.430 & 0.635 & 0.613
\\
\cellcolor{mygray}\small{\textbf{CAT}} & \cellcolor{mygray}\textbf{0.088} & \cellcolor{mygray}\textbf{0.745} & \cellcolor{mygray}\textbf{0.830} & \cellcolor{mygray}\textbf{0.791} & \cellcolor{mygray}\textbf{0.143} & \cellcolor{mygray}\textbf{0.449} & \cellcolor{mygray}\textbf{0.678} & \cellcolor{mygray}\textbf{0.626}
\\
\bottomrule[1pt]
\end{tabular}}
\vspace{-0.25cm}
\caption{Benchmarking results of different datasets trained by \textit{EGNet} \cite{Model-EGNet} and evaluated on \textit{CoSOD3k} ({left}) and \textit{CoCA} ({right}).}
\label{tab:egnet}
\end{table}

\begin{table}[t]
\centering
\scalebox{0.66}{
\begin{tabular}{c|cccc|cccc}
\toprule[1pt]
\small{Dataset} & \small{\textit{MAE} $\downarrow$} & \small{$F_{max}$ $\uparrow$} & \small{$E_{avg}$ $\uparrow$} & \small{$S_{\alpha}$ $\uparrow$} & \small{\textit{MAE} $\downarrow$} & \small{$F_{max}$ $\uparrow$} & \small{$E_{avg}$ $\uparrow$} & \small{$S_{\alpha}$ $\uparrow$}
\\ \hline\hline
\small{DUTS-Class} & 0.088 & 0.761 & 0.835 & 0.798 & 0.148 & 0.473 & 0.667 & 0.637
\\
\small{COCO9213}$^{\ddag}$ & 0.089 & 0.762 & 0.840 & 0.794 & 0.147 & 0.513 & 0.683 & 0.654
\\
\cellcolor{mygray}\small{\textbf{CAT}} & \cellcolor{mygray}\textbf{0.068} & \cellcolor{mygray}\textbf{0.791} & \cellcolor{mygray}\textbf{0.866} & \cellcolor{mygray}\textbf{0.811} & \cellcolor{mygray}\textbf{0.108} & \cellcolor{mygray}\textbf{0.529} & \cellcolor{mygray}\textbf{0.734} & \cellcolor{mygray}\textbf{0.673}
\\ \hline\hline
\small{DUTS-Class} & 0.089 & 0.758 & 0.834 & 0.793 & 0.146 & 0.469 & 0.677 & 0.634
\\
\small{COCO9213}$^{\ddag}$ & 0.095 & 0.767 & 0.835 & 0.797 & 0.155 & 0.522 & 0.670 & 0.653
\\
\cellcolor{mygray}\small{\textbf{CAT}} & \cellcolor{mygray}\textbf{0.067} & \cellcolor{mygray}\textbf{0.792} & \cellcolor{mygray}\textbf{0.867} & \cellcolor{mygray}\textbf{0.817} & \cellcolor{mygray}\textbf{0.101} & \cellcolor{mygray}\textbf{0.533} & \cellcolor{mygray}\textbf{0.743} & \cellcolor{mygray}\textbf{0.678}
\\
\bottomrule[1pt]
\end{tabular}}
\vspace{-0.25cm}
\caption{Benchmarking results of different datasets trained by \textit{ICNet} \cite{Model-ICNet} and evaluated on \textit{CoSOD3k} ({left}) and \textit{CoCA} ({right}). {Row 2-4}: SISMs generated by \textit{EGNet} with VGG backbone; {Row 5-7}: SISMs generated by \textit{EGNet} with ResNet backbone.}
\label{tab:icnet}
\end{table}

\begin{table}[t]
\centering
\scalebox{0.656}{
\begin{tabular}{c|cccc|cccc}
\toprule[1pt]
\small{Dataset} & \small{\textit{MAE} $\downarrow$} & \small{$F_{max}$ $\uparrow$} & \small{$E_{avg}$ $\uparrow$} & \small{$S_{\alpha}$ $\uparrow$} & \small{\textit{MAE} $\downarrow$} & \small{$F_{max}$ $\uparrow$} & \small{$E_{avg}$ $\uparrow$} & \small{$S_{\alpha}$ $\uparrow$}
\\ \hline\hline
\small{COCO-SEG} & 0.080 & 0.767 & 0.853 & 0.793 & 0.133 & 0.508 & 0.695 & 0.654
\\
\small{DUTS-Class}$^{\ddag}$ & 0.081 & 0.766 & 0.845 & 0.802 & 0.150 & 0.475 & 0.656 & 0.632
\\
\small{COCO9213} & 0.084 & 0.753 & 0.839 & 0.783 & 0.143 & 0.499 & 0.679 & 0.641
\\
\cellcolor{mygray}\small{\textbf{CAT}} & \cellcolor{mygray}\textbf{0.070} & \cellcolor{mygray}\textbf{0.786} & \cellcolor{mygray}\textbf{0.863} & \cellcolor{mygray}\textbf{0.812} & \cellcolor{mygray}\textbf{0.116} & \cellcolor{mygray}\textbf{0.523} & \cellcolor{mygray}\textbf{0.716} & \cellcolor{mygray}\textbf{0.668}
\\
\bottomrule[1pt]
\end{tabular}}
\vspace{-0.25cm}
\caption{Benchmarking results of different datasets trained by \textit{GICD} \cite{Model-GICD} and evaluated on \textit{CoSOD3k} ({left}) and \textit{CoCA} ({right}).}
\label{tab:gicd}
\end{table}

\begin{table}[t]
\centering
\scalebox{0.66}{
\begin{tabular}{c|cccc|cccc}
\toprule[1pt]
\small{Dataset} & \small{\textit{MAE} $\downarrow$} & \small{$F_{max}$ $\uparrow$} & \small{$E_{avg}$ $\uparrow$} & \small{$S_{\alpha}$ $\uparrow$} & \small{\textit{MAE} $\downarrow$} & \small{$F_{max}$ $\uparrow$} & \small{$E_{avg}$ $\uparrow$} & \small{$S_{\alpha}$ $\uparrow$}
\\ \hline\hline
\small{COCO-SEG} & 0.084 & 0.755 & 0.842 & 0.780 & 0.119 & 0.527 & 0.720 & 0.656
\\
\small{DUTS-Class} & 0.074 & 0.771 & 0.850 & 0.802 & 0.122 & 0.513 & 0.704 & 0.657
\\
\small{COCO9213} & 0.081 & 0.751 & 0.839 & 0.779 & 0.120 & 0.520 & 0.710 & 0.654
\\
\cellcolor{mygray}\small{\textbf{CAT}} & \cellcolor{mygray}\textbf{0.065} & \cellcolor{mygray}\textbf{0.789} & \cellcolor{mygray}\textbf{0.862} & \cellcolor{mygray}\textbf{0.813} & \cellcolor{mygray}\textbf{0.089} & \cellcolor{mygray}\textbf{0.531} & \cellcolor{mygray}\textbf{0.728} & \cellcolor{mygray}\textbf{0.673}
\\
\bottomrule[1pt]
\end{tabular}}
\vspace{-0.25cm}
\caption{Benchmarking results of different datasets trained by \textit{GCoNet} \cite{Model-GCoNet} and evaluated on \textit{CoSOD3k} ({left}) and \textit{CoCA} ({right}).}
\label{tab:gconet}
\end{table}

\begin{table}[t]
\centering
\scalebox{0.8}{
\begin{tabular}{c|ccccccc}
\toprule[1pt]
\small{Method} & \small{\textit{MAE} $\downarrow$} & \small{$F_{max}$ $\uparrow$} & \small{$F_{avg}$ $\uparrow$} & \small{$E_{max}$ $\uparrow$} & \small{$E_{avg}$ $\uparrow$} & \small{$S_{\alpha}$ $\uparrow$}
\\ \hline\hline
\small{RandomFlip} & 0.080 & 0.766 & 0.754 & 0.851 & 0.840 & 0.795
\\
\small{RandomRotate} & 0.103 & 0.729 & 0.718 & 0.831 & 0.819 & 0.767
\\ \hline
\small{AutoAug} & 0.098 & 0.732 & 0.723 & 0.836 & 0.821 & 0.779
\\
\small{CutOut} & 0.110 & 0.715 & 0.707 & 0.812 & 0.807 & 0.762
\\
\small{CutMix} & 0.087 & 0.745 & 0.739 & 0.839 & 0.836 & 0.772
\\
\small{Jigsaw} & 0.079 & 0.770 & 0.763 & 0.848 & 0.845 & 0.797
\\
\cellcolor{mygray}\small{\textbf{GCP}} & \cellcolor{mygray}\textbf{0.073} & \cellcolor{mygray}\textbf{0.802} & \cellcolor{mygray}\textbf{0.785} & \cellcolor{mygray}\textbf{0.866} & \cellcolor{mygray}\textbf{0.849} & \cellcolor{mygray}\textbf{0.804}
\\
\bottomrule[1pt]
\end{tabular}}
\vspace{-0.25cm}
\caption{Comparison among different {augmentation techniques} associated with \textit{GICD} \cite{Model-GICD} and evaluated on \textit{CoSOD3k}.}
\label{tab:aug}
\vspace{-0.25cm}
\end{table}
\begin{table}[ht]
\centering
\scalebox{0.65}{
\begin{tabular}{c|ccc|cccccc}
\toprule[1pt]
\# & Group & Cut & Paste & \textit{MAE} $\downarrow$ & $F_{max}$ $\uparrow$ & $F_{avg}$ $\uparrow$ & $E_{max}$ $\uparrow$ & $E_{avg}$ $\uparrow$ & $S_{\alpha}$ $\uparrow$ 
\\ \hline\hline
\small{A} & $\diamond$ & & & .203 & .419 & .413 & .612 & .601 & .591
\\
\small{B} & \cmark & & & .150 & .472 & .466 & .679 & .663 & .633
\\
\small{C} & $\diamond$ & & $\diamond$ & .178 & .408 & .402 & .631 & .622 & .582
\\
\small{D} & $\diamond$ & \cmark & $\diamond$ & .173 & .417 & .411 & .648 & .635 & .586
\\
\small{E} & \cmark & & $\diamond$ & .139 & .460 & .456 & .699 & .691 & .627
\\
\small{F} & \cmark & \cmark & $\diamond$ & .130 & .474 & .469 & .700 & .692 & .638
\\\hline
\small{Ours} & \cmark & \cmark & \cmark & .102 & .522 & .510 & .721 & .711 & .662
\\
\bottomrule[1pt]
\end{tabular}}
\vspace{-0.25cm}
\caption{{Ablation study} for our proposed GCP approach with \textit{GICD} \cite{Model-GICD} on the \textit{CoCA} dataset. Symbols \cmark and $\diamond$ denote conduct and randomly conduct corresponding operations, respectively.}
\label{tab:ablation}
\vspace{-0.25cm}
\end{table}

\noindent\textbf{Evaluation.}
We adopt the two recent and popular evaluation datasets \textit{CoSOD3k} \cite{Dataset-CoSOD3k} and \textit{CoCA} \cite{Model-GICD} to evaluate model performances. Four conventional metrics are adopted: mean absolute error (\textit{MAE}) \cite{Metrics01}, F-measure \cite{Metrics02}, E-measure \cite{Metrics03}, and structure measure ($S_{\alpha}$) \cite{Metrics04}. For F-measure \cite{Metrics02}, we follow the convention by setting weight ${\beta}^{2}=0.3$ and report both the maximum score ($F_{max}$) and the mean score ($F_{avg}$). We also report both $E_{max}$ and $E_{min}$ for E-measure \cite{Metrics03}. We use symbols $\uparrow$ and $\downarrow$ to denote the metrics which are the higher and the lower the better, respectively. Superscript $^{\ddag}$ denotes the reported results or results reproduced by public checkpoints.

\subsection{Quantitative Analysis}

\noindent\textbf{Longitudinal Comparison.}
From Tables \ref{tab:poolnet} - \ref{tab:gconet}, we observe that all models trained on CAT achieve better results than other datasets in all metrics on both \textit{CoSOD3k} and \textit{CoCA}. This suggests that CAT is agnostic to models and backbones. Specifically, the improvements are typically larger in \textit{MAE}. Significant boosts for on average more than $25\%$ are achieved on \textit{PoolNet}, \textit{EGNet}, \textit{ICNet}, and \textit{GCoNet}, as well as a gain around $18\%$ on \textit{GICD}. Regarding the average improvements on other metrics, we observe that CAT improves more on $F$ ($\sim9\%$), which shows that instead of radically regarding every foreground objects as salient, models trained on CAT tend to make relatively conservative predictions and focus more on regions with co-salient objects only. This supports our analysis in the previous sections. It also provides improvements in $E$ ($\sim7\%$) and $S_{\alpha}$ ($\sim5\%$). Indeed, $S_{\alpha}$ is the hardest among metrics which measures both accuracy and structural similarity. Although models are capable of making more confident predictions now, they cannot capture very detailed information like boundaries. Overall, we believe that the diversity and quality of CAT have laid a solid foundation for more in-depth works in the future.

\noindent\textbf{Horizontal Comparison.}
Changing the direction of comparison, we notice that the evaluation scores on \textit{CoSOD3k} are much higher than \textit{CoCA}. This is in line with the characteristics of these two datasets: although the former is richer in scale, the latter has more complex context and more variant appearance, such as occlusion and object size. Besides, the performances of co-saliency detection models are much better than that of saliency detection models, which emphasize again the importance of specialized modules/techniques designed for identifying co-salient signals and suppressing non-salient ones. As for different augmentation techniques (see Table \ref{tab:aug}), they share a common problem during training: instability. For those without considerations of co-saliency like \textit{AutoAug}, \textit{CutOut}, and \textit{CutMix}, the performances are degraded. Only \textit{Jigsaw} and our GCP procedure surpass the baseline and the latter offers even larger improvements.

\begin{figure*}[t]
\begin{center}
\includegraphics[width=0.9999\textwidth]{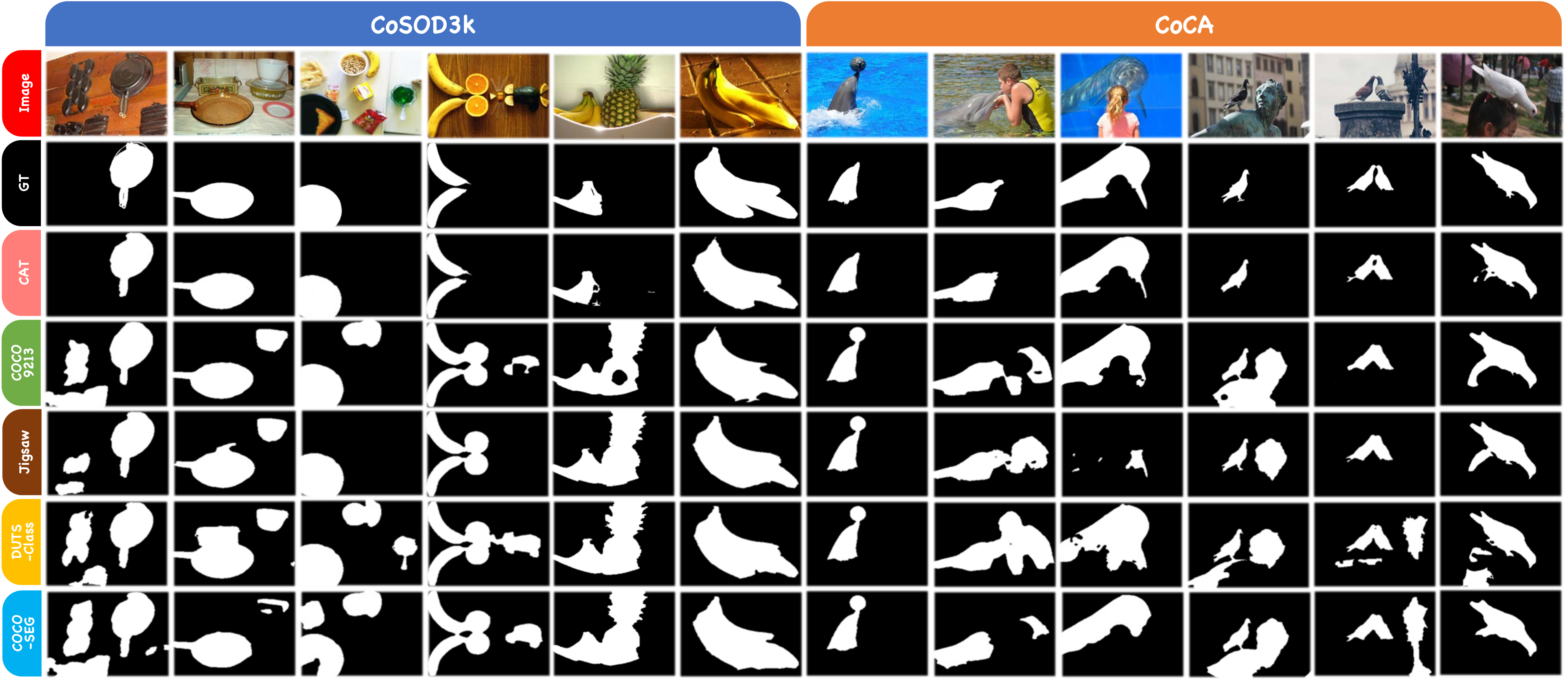}
\end{center}
\vspace{-0.37cm}
\caption{Qualitative results of different datasets trained by \textit{GICD} \cite{Model-GICD}. \textbf{Left}: results on \textit{CoSOD3k} \cite{Dataset-CoSOD3k}; \textbf{Right}: results on \textit{CoCA} \cite{Model-GICD}. Groups from left to right: \textit{frying pan}, \textit{banana}, \textit{dolphin}, and \textit{pigeon}.}
\label{fig:comparison}
\vspace{-0.cm}
\end{figure*}

\noindent\textbf{Ablation Study.}
We conduct an ablation study to verify the effectiveness of our methods (see Table \ref{tab:ablation}). The results for the original 8,375 images are shown in rows A and B. We also provide other combinations for comparisons, e.g., group images w/ classifiers (\textit{vs.} randomly allocate images to the same number of groups), cut candidate objects out before paste (\textit{vs.} do not cut and paste the sampled images directly), and paste objects/images under counterfactual rules (\textit{vs.} no paste or randomly paste w/o limitations). Specifically, ``\textit{Group}" offers proper inter-saliency signals, which is essential for models to find the common salient objects between image groups (A \textit{vs.} B, C \textit{vs.} E). ``\textit{Cut}" gives fine-grained category information and further improves performance (C \textit{vs.} D, E \textit{vs.} F). As the key part, ``\textit{Paste}" provides intra-saliency signals, which determines the quality of the collaborative learning paradigm. A proper introduction of such signals significantly boosts the performance (Ours \textit{vs.} F), and vice versa (A \textit{vs.} C, B \textit{vs.} E).

\subsection{Qualitative Analysis}
Figure \ref{fig:comparison} illustrates some visualizations among different competitors on \textit{CoSOD3k} and \textit{CoCA} generated by \textit{GICD}. It can be easily observed that the quality of a dataset plays an important role for model training. Models encoded with features from high-quality dataset like CAT tends to focus more on the true co-salient signals while suppressing the non-salient ones. Taking the \textit{frying pan} group as an example. Both semantic segmentation and saliency detection datasets fail to teach the model ``what" and ``where" to focus and thus let it mistakenly regards items besides \textit{frying pans} as co-salient. In contrast, the model trained on CAT can well distinguish co-salient objects from non-salient clusters even when they share similar shapes, textures, colors, etc. We refer to the supplementary for more examples. Another point worthy of attention is that the existing models cannot well-restore the details (e.g., the boundaries) of objects, especially that of small objects. For example, the contours of the \textit{pigeons} in the last two columns of Figure \ref{fig:comparison} are not well drawn. A possible reason for this is that most current co-saliency models adopt VGG-16 \cite{VGG} as their backbones and rely on low-resolution inputs (e.g., 224 $\times$ 224). Some minor but important features are missing during the down-sampling. With the emergence of more and more deep and efficient backbones, we encourage future works to explore more possibilities.

\section{Conclusion}
In this work, we addressed the inconsistency problem in co-saliency detection. Borrowing the idea from cause and effect, we proposed a counterfactual training framework to formalize the co-salient signals during training. A group-cut-paste procedure is introduced to leverage existing data samples and makes ``cost-free" augmentation by context adjustment. Follow this procedure, we collected a large-scale dataset call Context Adjustment Training. The abundant semantics and high-quality annotations of our dataset liberates models from the impediment caused by spurious variations and biases and significantly improves model performances. As moving forward, we are going to take a deeper look at the co-salient signals both across image groups and within each sample, and improve the quality and scale of our dataset accordingly.
\section{Appendix}

In this appendix, we further detail the following aspects to supplement the main body of this paper.

\begin{itemize}
    \item \textbf{Section \textcolor{red}{A}: Dataset Characteristics}, which includes additional characteristic details of our proposed dataset and more technical details of our GCP procedure.
    \item \textbf{Section \textcolor{red}{B}: Experimental Details}, which includes additional details for the training and evaluations and more qualitative results of our benchmark experiments.
    \item \textbf{Section \textcolor{red}{C}: Benchmark Study}, which includes the complete results and analysis of our benchmark experiments, more information for our benchmark study, and discussions for the future directions.
\end{itemize}
\section{Dataset Characteristics}
\label{sec:dataset}

In this section, we further supplement additional statistical information and examples for the superclass and subclass of our dataset and extensive details for the data collection and cleaning of our GCP procedure.

\begin{figure}[!ht]
\begin{center}
\includegraphics[width=0.474\textwidth]{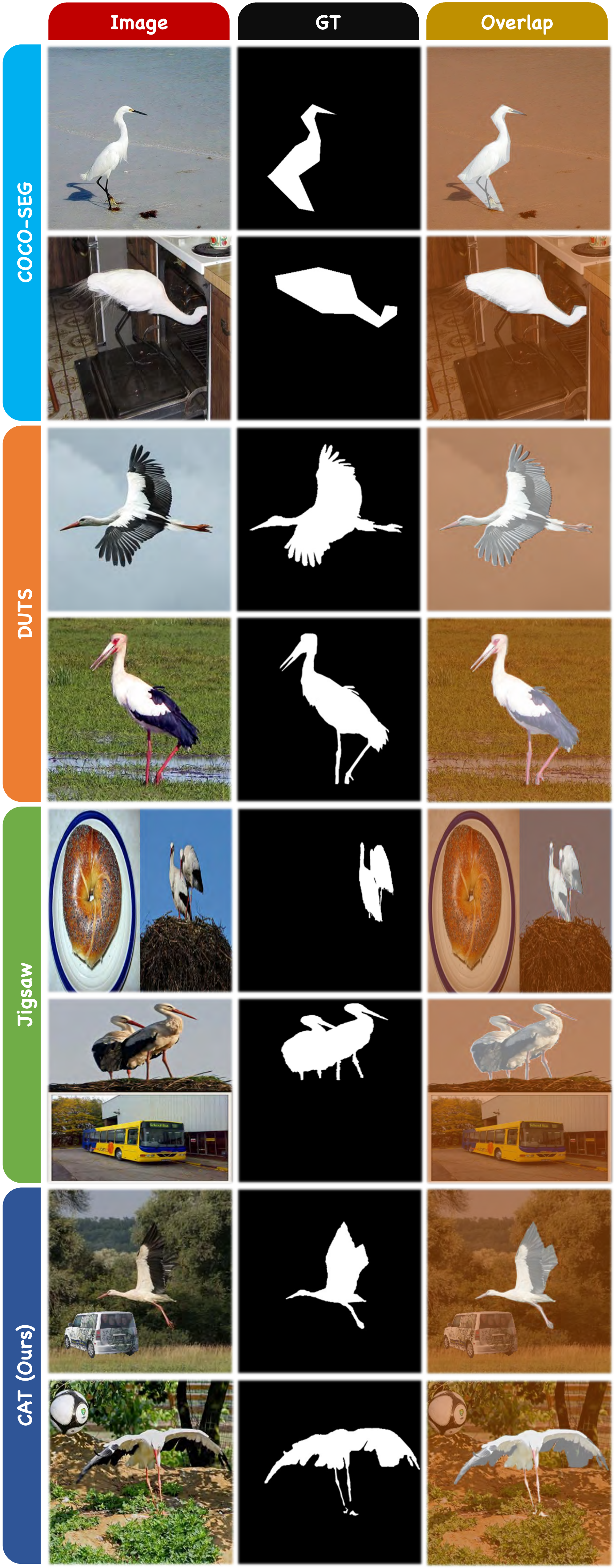}
\end{center}
\vspace{-0.3cm}
\caption{Comparison among different datasets. Semantic group: \textit{white stork}.}
\label{fig:mask_comparison_1}
\vspace{-0.5cm}
\end{figure}

\subsection{Dataset Comparison}

As mentioned in Sections \textcolor{blue}{1} (Introduction) and \textcolor{blue}{2} (Related Work) of the main body, current co-saliency detection models suffer from the inconsistency among training distribution $\mathcal{D}$, testing distribution $\mathcal{E}$, and true distribution $\mathcal{T}$. Although recent works \cite{Dataset-CoSOD3k,Model-GICD} help $\mathcal{E}$ to move one-step closer towards $\mathcal{T}$, a dataset that can properly define $\mathcal{D}$ is still missing. Most works have to rely instead on semantic segmentation datasets (e.g., \textit{COCO-SEG} \cite{Dataset-COCO-SEG} and \textit{COCO9213} \cite{Dataset-COCO}) and saliency detection datasets (e.g., \textit{DUTS} \cite{Dataset-DUTS}) for model training, which exacerbates the distribution gap.

\begin{figure}[!ht]
\begin{center}
\includegraphics[width=0.474\textwidth]{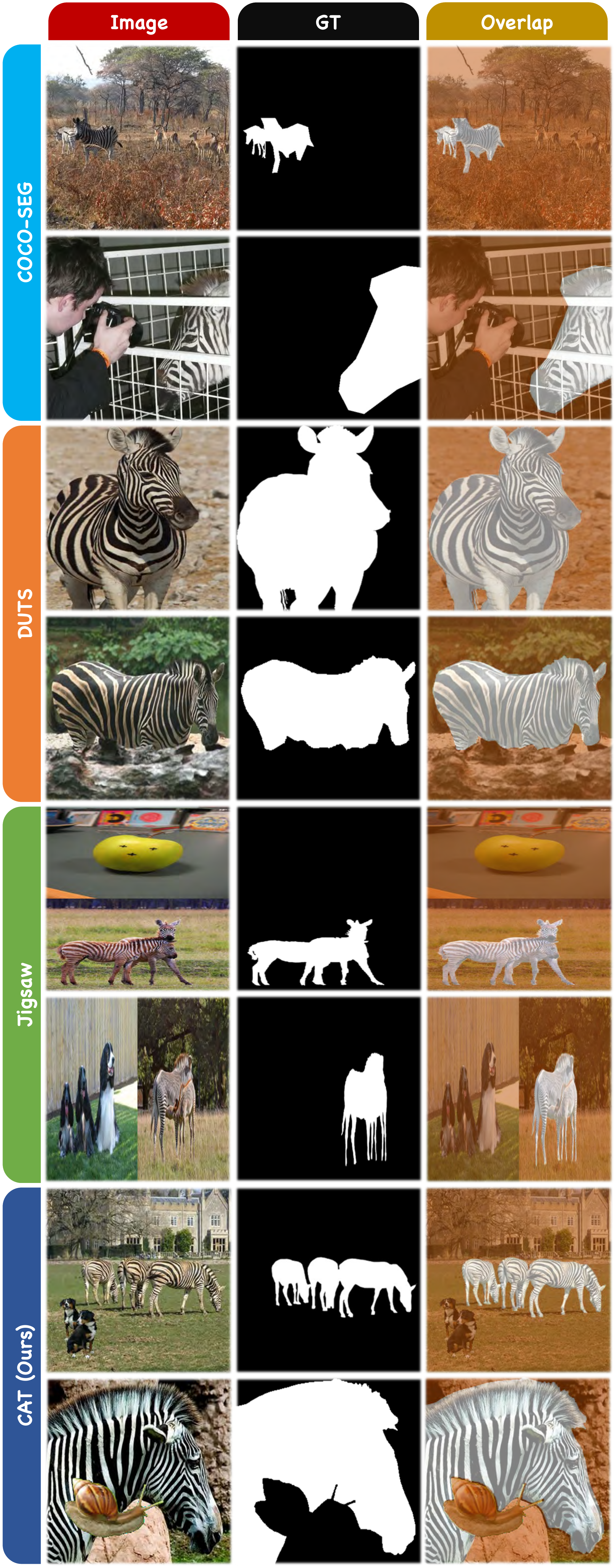}
\end{center}
\vspace{-0.3cm}
\caption{Comparison among different datasets. Semantic group: \textit{zebra}.}
\label{fig:mask_comparison_2}
\vspace{-0.5cm}
\end{figure}

Figures \ref{fig:mask_comparison_1} and \ref{fig:mask_comparison_2} show some typical examples from \textit{COCO-SEG} \cite{Dataset-COCO-SEG}, \textit{DUTS} \cite{Dataset-DUTS}, \textit{Jigsaw} \cite{Model-GICD}, and our proposed dataset. Specifically, although there are rich contexts, the annotations of \textit{COCO-SEG} \cite{Dataset-COCO-SEG} are coarse. Unlike pixel-by-pixel labeling, the annotations of \textit{COCO-SEG} \cite{Dataset-COCO-SEG} are composed of polygons and triangles delimited by points. The fine-grained part of an object (e.g., the \texttt{legs} and \texttt{tails} of the \textit{while stork}) and the object contours (e.g., the \textit{zebra} behind \texttt{shrubs} and \texttt{railings}) cannot be appropriately depicted. In contrast, saliency detection datasets like \textit{DUTS} \cite{Dataset-DUTS} have sophisticated mask annotations. Parts such as the wingtip \texttt{feathers} and the slender \texttt{legs} of the \textit{white stork} and the \texttt{ears} of the \textit{zebra} are well-annotated. The reason why these examples are so different is that these two tasks have different purposes. While the former pays more attention to the segmentation of multiple objects in complex scenes, the latter emphasizes more on the high-precision detection of salient objects. However, the uni-object distribution of \textit{DUTS} \cite{Dataset-DUTS} is not in line with the objective of co-saliency detection \cite{Model-CoEGNet}, i.e., to identify co-salient signals from multi-foreground visual contexts in a group-by-group manner. 

Recent work \cite{Model-GICD} proposes to concatenate two images horizontally or vertically to form a multi-object jigsaw. This includes two resize processes: (1) resize candidate images to squares before concatenation; and (2) resize jigsawed images to squares before feeding into the network. Although this technique introduces co-salient signals, it also leads to severe distortion and incoherent concatenation boundaries. As shown in Figures \ref{fig:mask_comparison_1} and \ref{fig:mask_comparison_2}, the jigsawed images become cluttered, and the objects in some examples (e.g., the \textit{zebra} underneath the green \textit{apple} and beside three \textit{dogs}) are difficult to be recognized even by human eyes. This deviates from the original intention of co-saliency detection, i.e., to imitate human visual attentions \cite{Model-CoSaliency}. On the contrary, the images in our dataset have clear visual contexts, carefully defined co-saliency, as well as high-quality annotations.

\subsection{Data Collection}
As mentioned in Section \textcolor{blue}{3.2} (Group-Cut-Paste) of the main body, our GCP procedure leverages images from existing datasets for context adjustment. This means that the quality of the canvas determines the quality of the entire adjustment process. Therefore, we strictly followed the following criteria when selecting candidate images:
\begin{itemize}
    \item Candidate images should have clear semantic meanings.
    \item Candidate images should have correct and sophisticated annotations.
\end{itemize}

\begin{figure}[t]
\begin{center}
\includegraphics[width=0.474\textwidth]{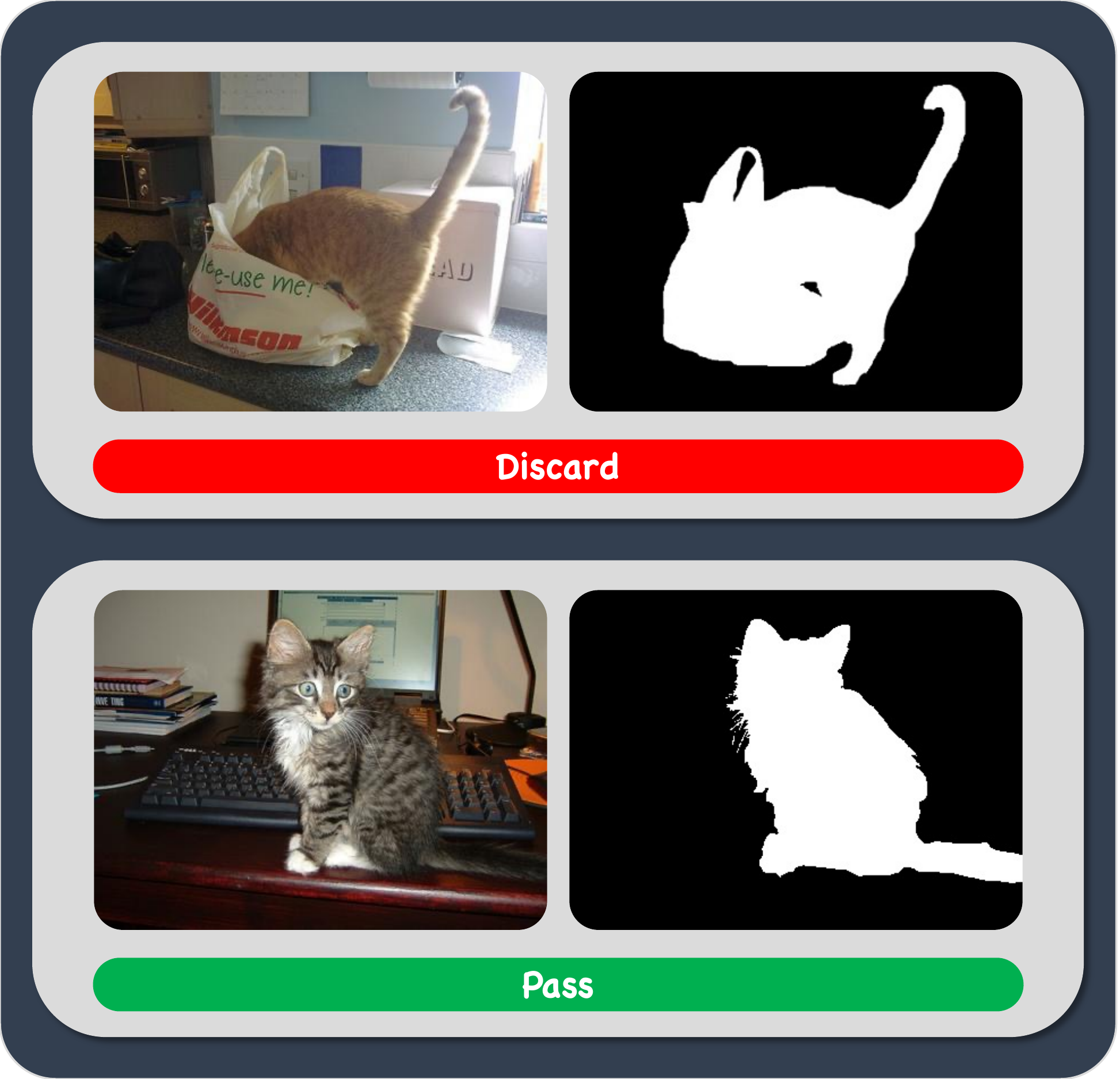}
\end{center}
\vspace{-0.3cm}
\caption{Data collection examples. Images with unclear/clear semantics are discarded/passed.}
\label{fig:data_collection_1}
\vspace{-0.cm}
\end{figure}

\begin{figure}[t]
\begin{center}
\includegraphics[width=0.474\textwidth]{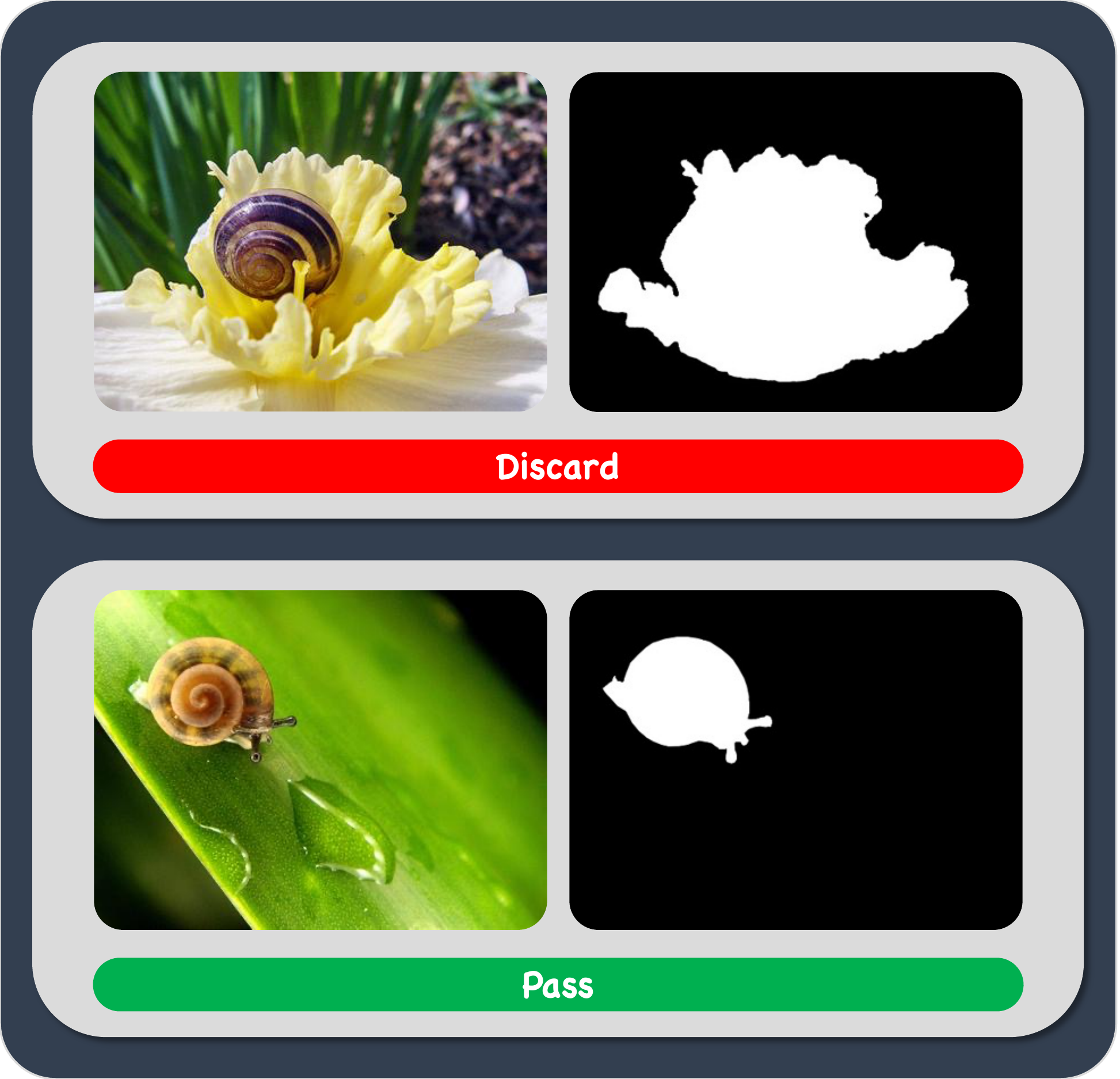}
\end{center}
\vspace{-0.3cm}
\caption{Data collection examples. Images with incorrect/correct annotations are discarded/passed.}
\label{fig:data_collection_2}
\vspace{-0.cm}
\end{figure}

Figures \ref{fig:data_collection_1} and \ref{fig:data_collection_2} show some discarded and passed cases (in terms of both image appearances and mask annotations). Specifically, we notice that there are cases that the key part of an object (e.g., the \texttt{ears} and \texttt{nose} of a \textit{cat}) is missing or the corresponding mask annotation of an image is incorrect. This is because some samples in saliency detection datasets do not contain distinguished semantic information and thus treat all the regions in the foreground as salient and label them accordingly. Obviously, these examples do not meet our requirements, so we choose to pick them out of the candidate pool. Following the above criteria, we collected 8,375 candidate images with clear saliency \cite{Saliency} and sophisticated annotations for further GCP processing.

\subsection{More Superclass Details}
\begin{figure*}[ht]
\begin{center}
\includegraphics[width=0.999\textwidth]{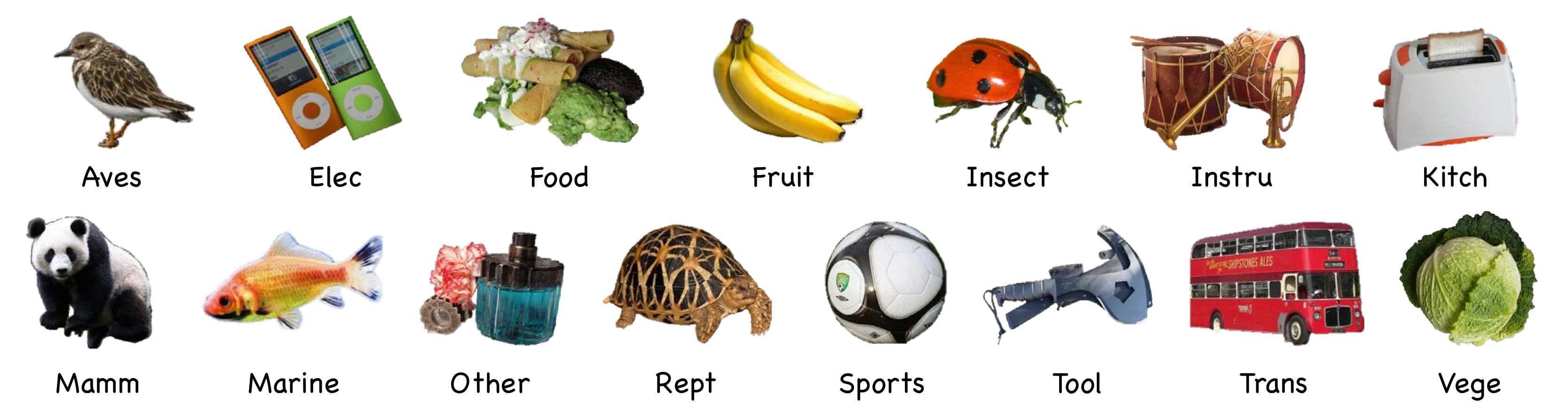}
\end{center}
\vspace{-0.3cm}
\caption{Representative example of each superclass in our dataset. Best viewed in color and zoomed-in for details.}
\label{fig:example15}
\vspace{-0.cm}
\end{figure*}

\begin{figure*}[!ht]
\begin{center}
\includegraphics[width=0.85\textwidth]{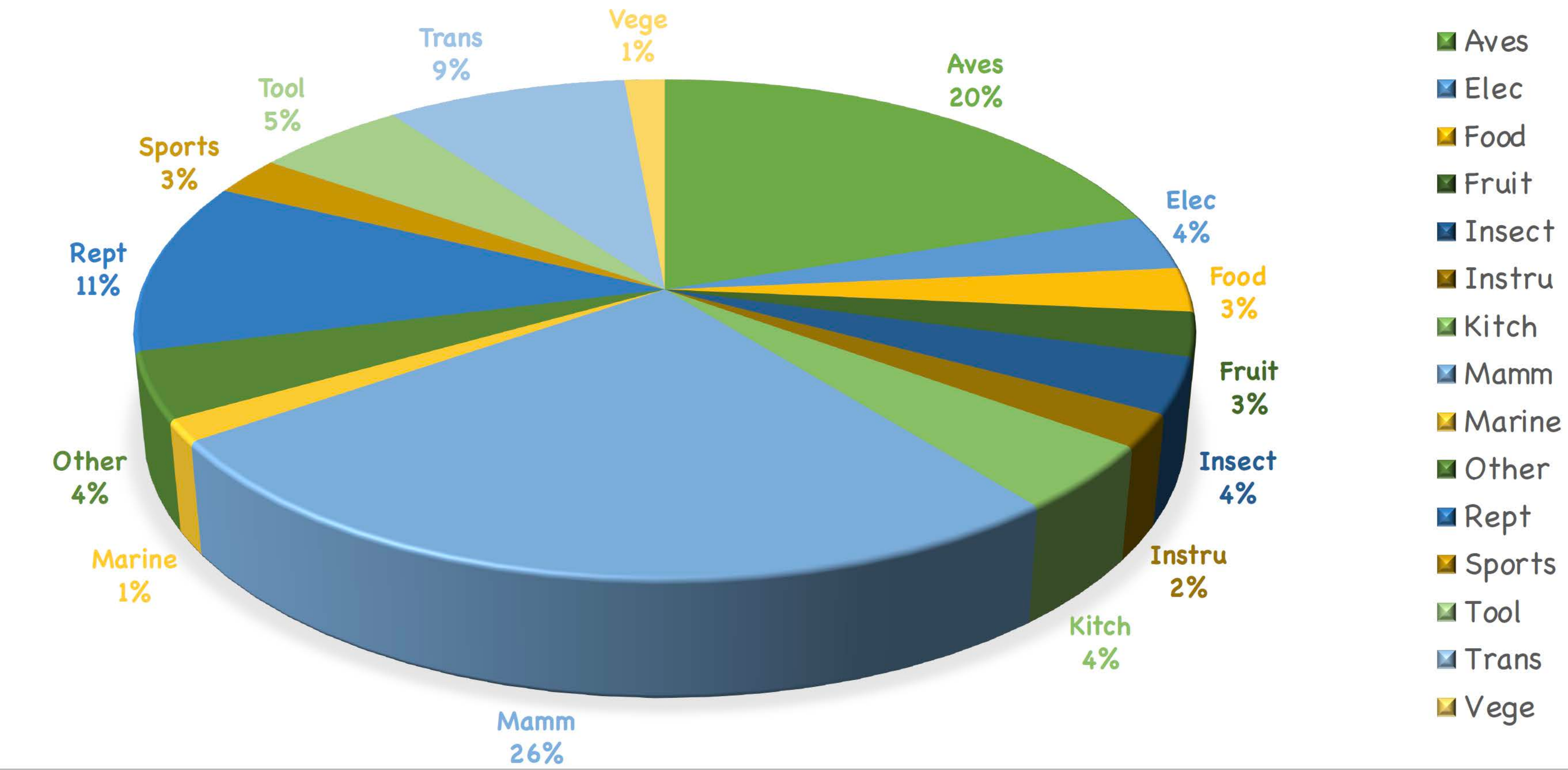}
\end{center}
\vspace{-0.3cm}
\caption{Pie chart for the superclass distribution of our dataset.}
\label{fig:pie}
\vspace{-0.cm}
\end{figure*}

\begin{figure*}[!ht]
\begin{center}
\includegraphics[width=0.999\textwidth]{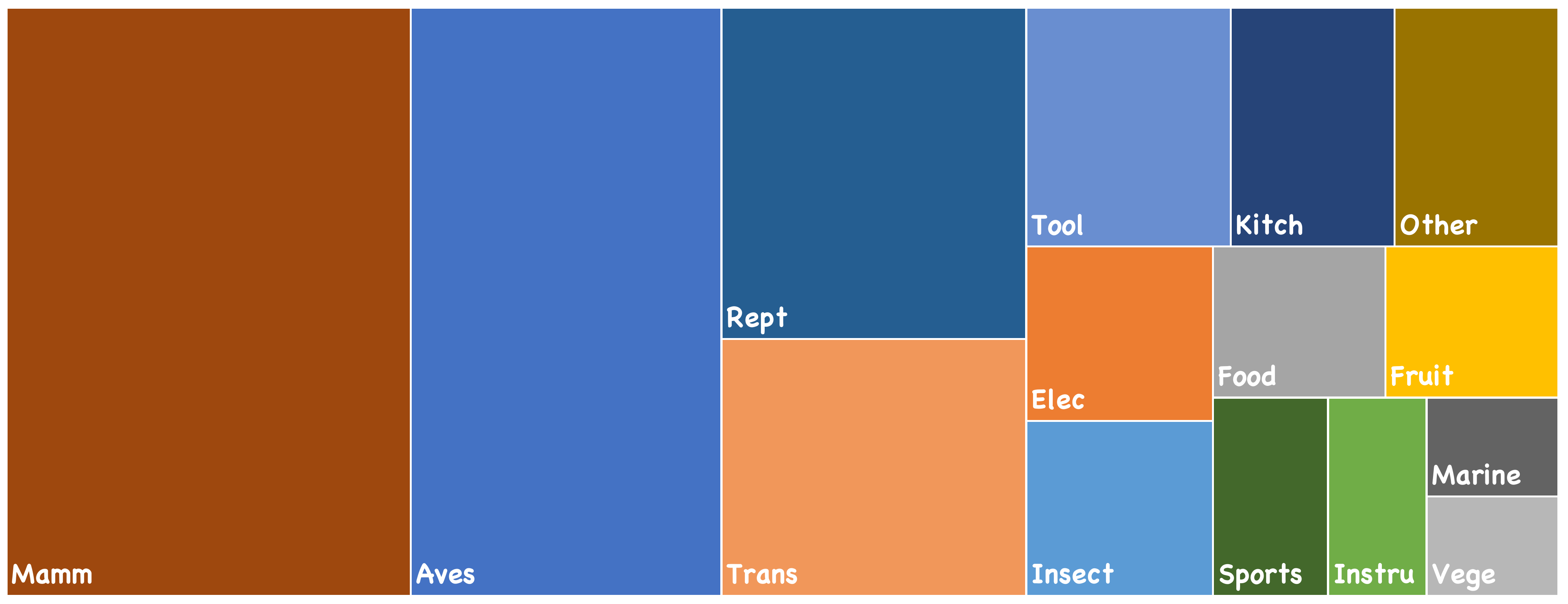}
\end{center}
\vspace{-0.3cm}
\caption{Tree chart for the superclass distribution of our dataset. The larger the area, the more semantic groups it contains.}
\label{fig:tree}
\vspace{-0.cm}
\end{figure*}

As mentioned in Section \textcolor{blue}{3.3} (Constructing the Dataset) of the main body, our dataset consists of 15 superclasses, i.e., aves, electronic product (elec.), food, fruit, insect, instrument (instru.), kitchenware (kitch.), mammal (mamm.), marine, other, reptile (rept.), sports, tool, transportation (trans.), and vegetable (vege.). Figure \ref{fig:example15} shows the representative example of each superclass. It can be seen that our dataset is diverse in semantics, which covers a large number of categories. The superclass distribution is shown in the pie chart (see Figure \ref{fig:pie}) and tree chart (see Figure \ref{fig:tree}), respectively. Specifically, mammal occupies 73 of the 280 semantic groups (23$\%$), which is the largest among the 15 superclasses. The second-largest is the aves, which consists of 56 semantic groups (20$\%$). The remaining superclasses occupy a total of 54$\%$ subclasses, of which the smallest two are marine and vegetable, each containing four semantic groups. In general, the 15 superclasses in our dataset cover a large number of categories in different fields. We believe that such diversity and comprehensiveness can benefit future works in co-saliency detection and beyond.

\subsection{More Subclass Details}

\begin{figure}[t]
\begin{center}
\includegraphics[width=0.43\textwidth]{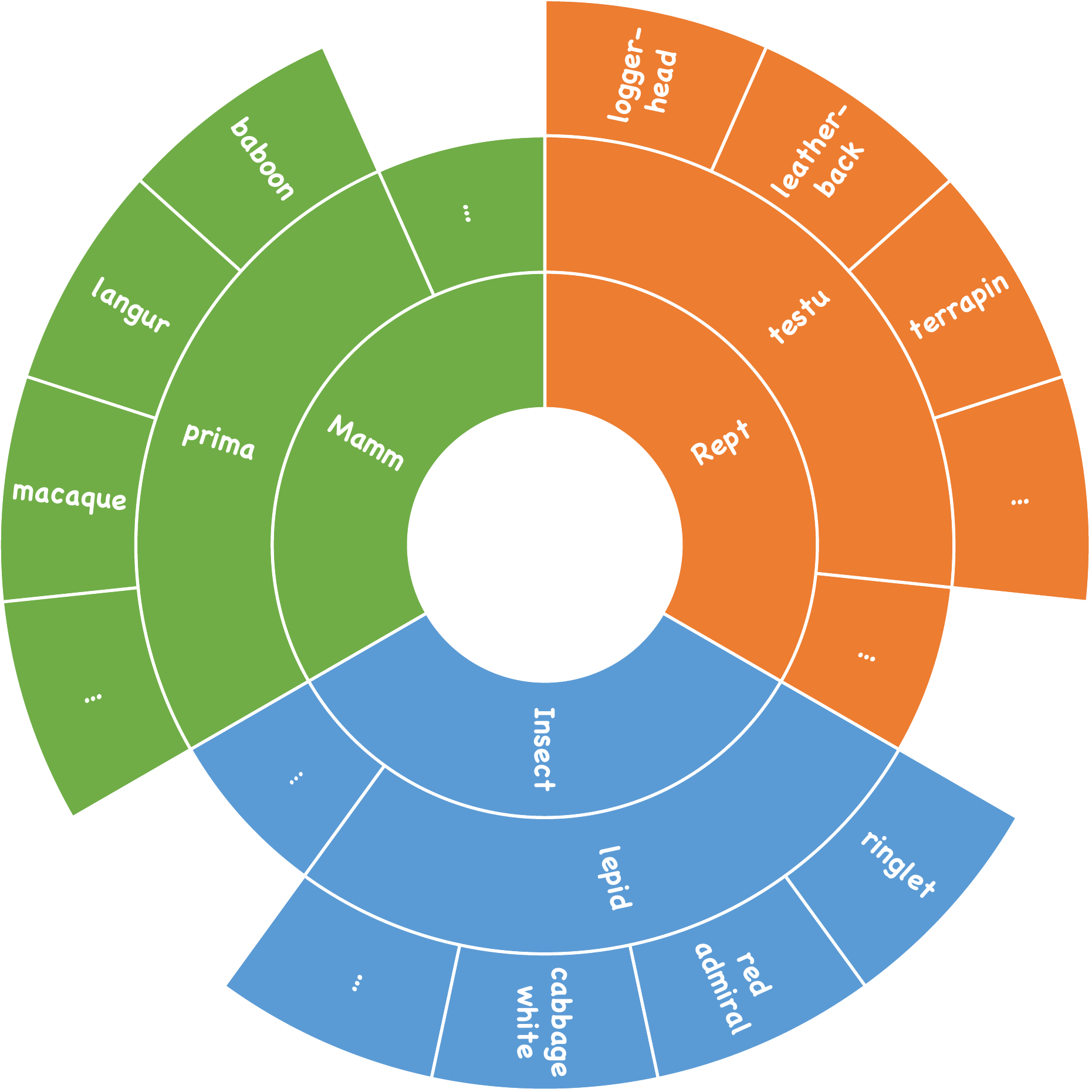}
\end{center}
\vspace{-0.3cm}
\caption{Examples of the ``semantic branches" in our dataset. ``lepid": Lepidoptera; ``testu": Testudines; ``prima": Primates.}
\label{fig:branch}
\vspace{-0.cm}
\end{figure}

\begin{figure*}[!ht]
\begin{center}
\includegraphics[width=0.999\textwidth]{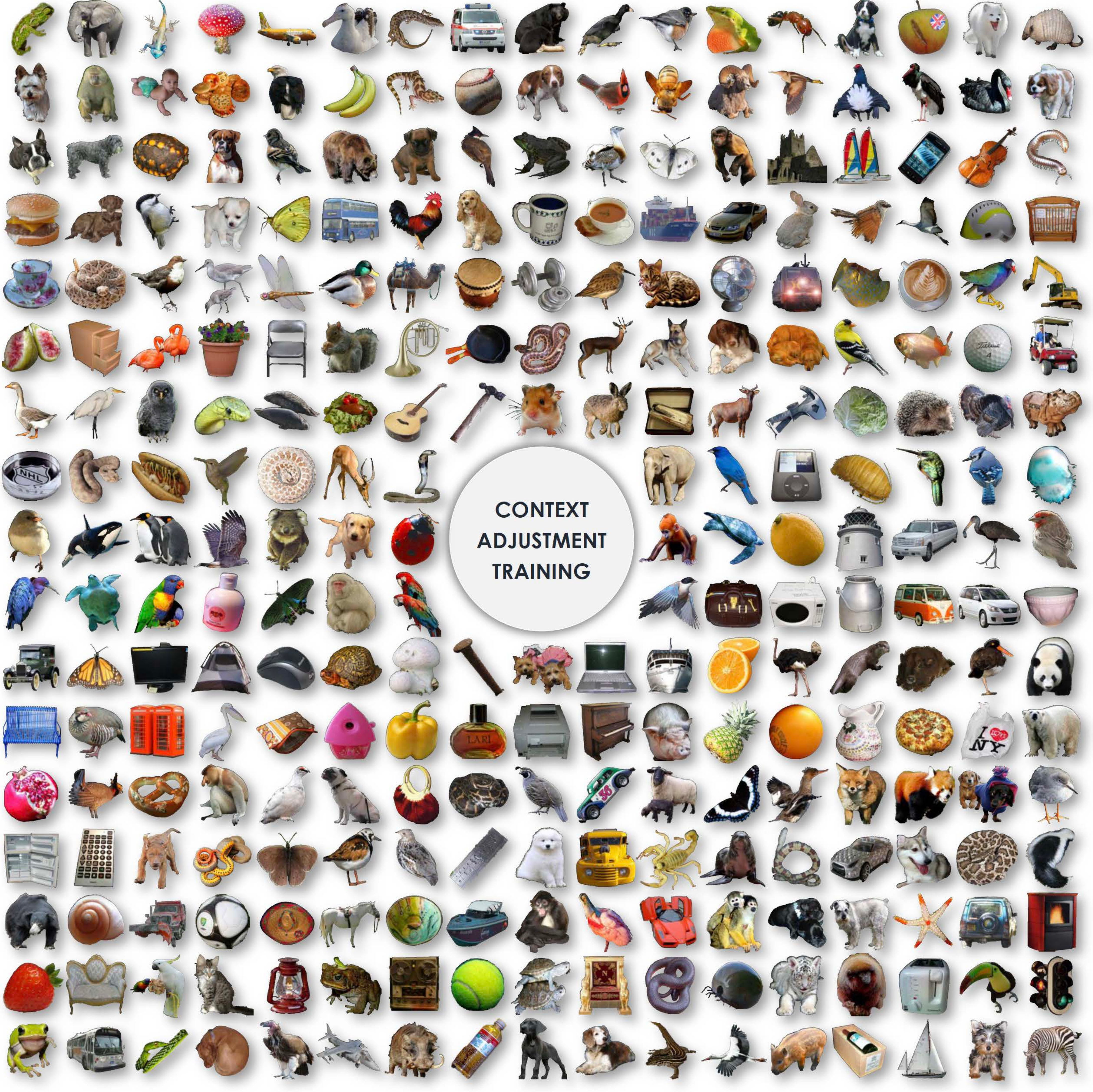}
\end{center}
\vspace{-0.3cm}
\caption{Representative example of each semantic group in our dataset. Best viewed in color and zoomed-in for details.}
\label{fig:example280}
\vspace{-0.1cm}
\end{figure*}

\begin{figure*}[!ht]
\begin{center}
\includegraphics[width=0.999\textwidth]{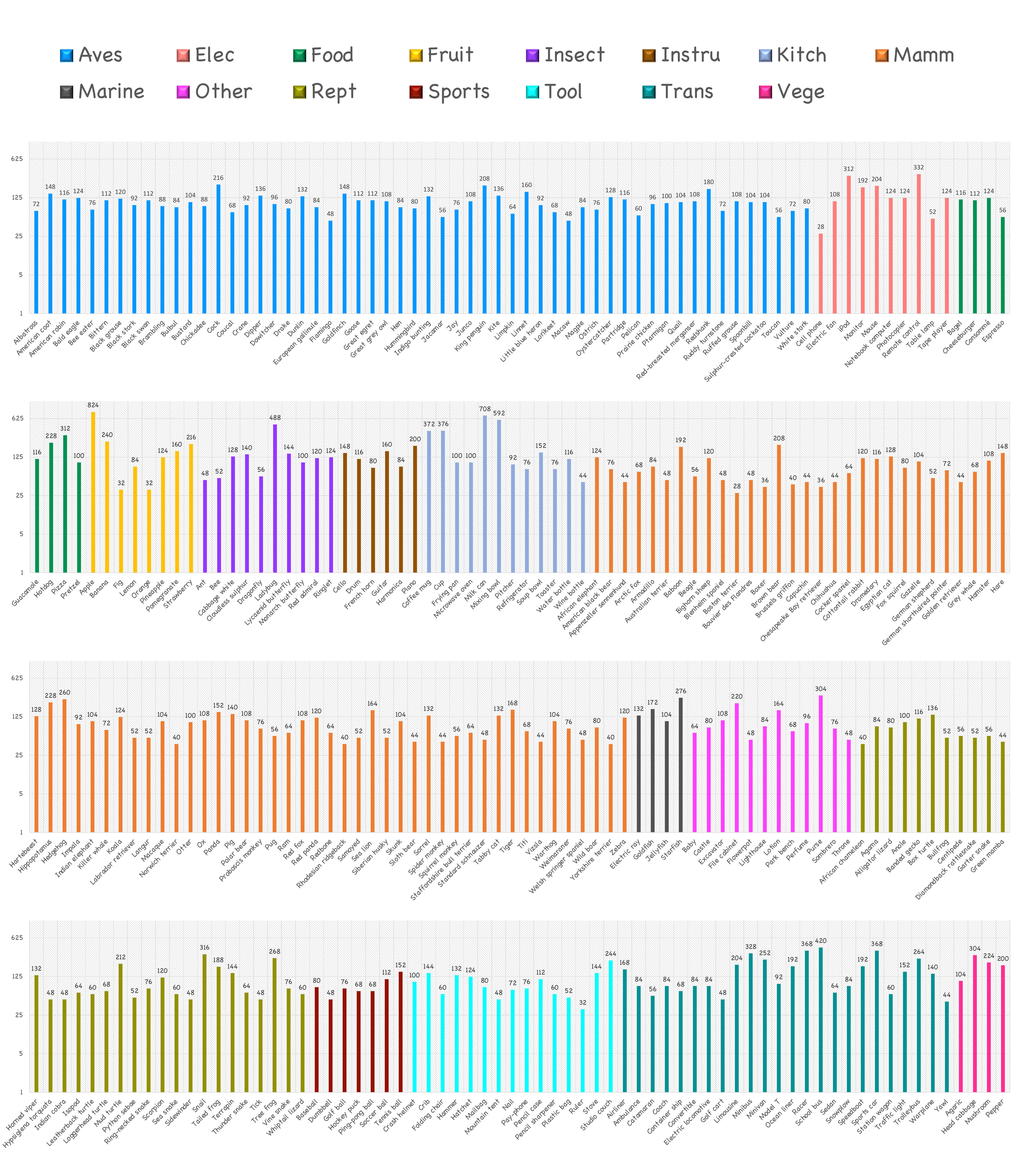}
\end{center}
\vspace{-0.6cm}
\caption{The distribution for the 280 semantic groups in CAT. Subclasses belong to the same superclass are highlighted with unified color. Best viewed in color and zoomed-in for details.}
\label{fig:distribution}
\vspace{-0.1cm}
\end{figure*}

As mentioned in Section \textcolor{blue}{3.3} (Constructing the Dataset) of the main body, our dataset contains 280 subclasses (semantic groups), of which range from biological species and daily necessities. Specifically, as shown in Figure \ref{fig:example280}, each semantic group in our dataset is a collection of one specific species. These species usually have a uniform appearance, such as a specific shape, color, texture, etc. Thanks to the powerful pre-training model \cite{Pretrain-WSL-Images}, these species are automatically classified to form diverse semantic groups. Figure \ref{fig:distribution} shows the distribution (w.r.t. the number of images per group) of all 280 semantic groups in CAT. The average number of images per group is 119.6. Among all the semantic groups, \textit{apple} is the largest one with 824 images, while \textit{boston terrier} and \textit{cell phone} are the smallest, which contain 28 images. Such a distribution is in line with that of the recent evaluation datasets, i.e., \textit{CoSOD3k} \cite{Dataset-CoSOD3k} and \textit{CoCA} \cite{Model-GICD}. The maximum/minimum numbers of images per group for these two datasets are 30/4 and 40/8, respectively. Another thing worth noting is the semantic hierarchy. Thanks to the abundant semantic information, our dataset offers diverse ``semantic branches" (see Figure \ref{fig:branch}), which are similar to the ``subtrees" of WordNet \cite{WordNet}. 

\begin{figure*}[!ht]
\begin{center}
\includegraphics[width=0.71\textwidth]{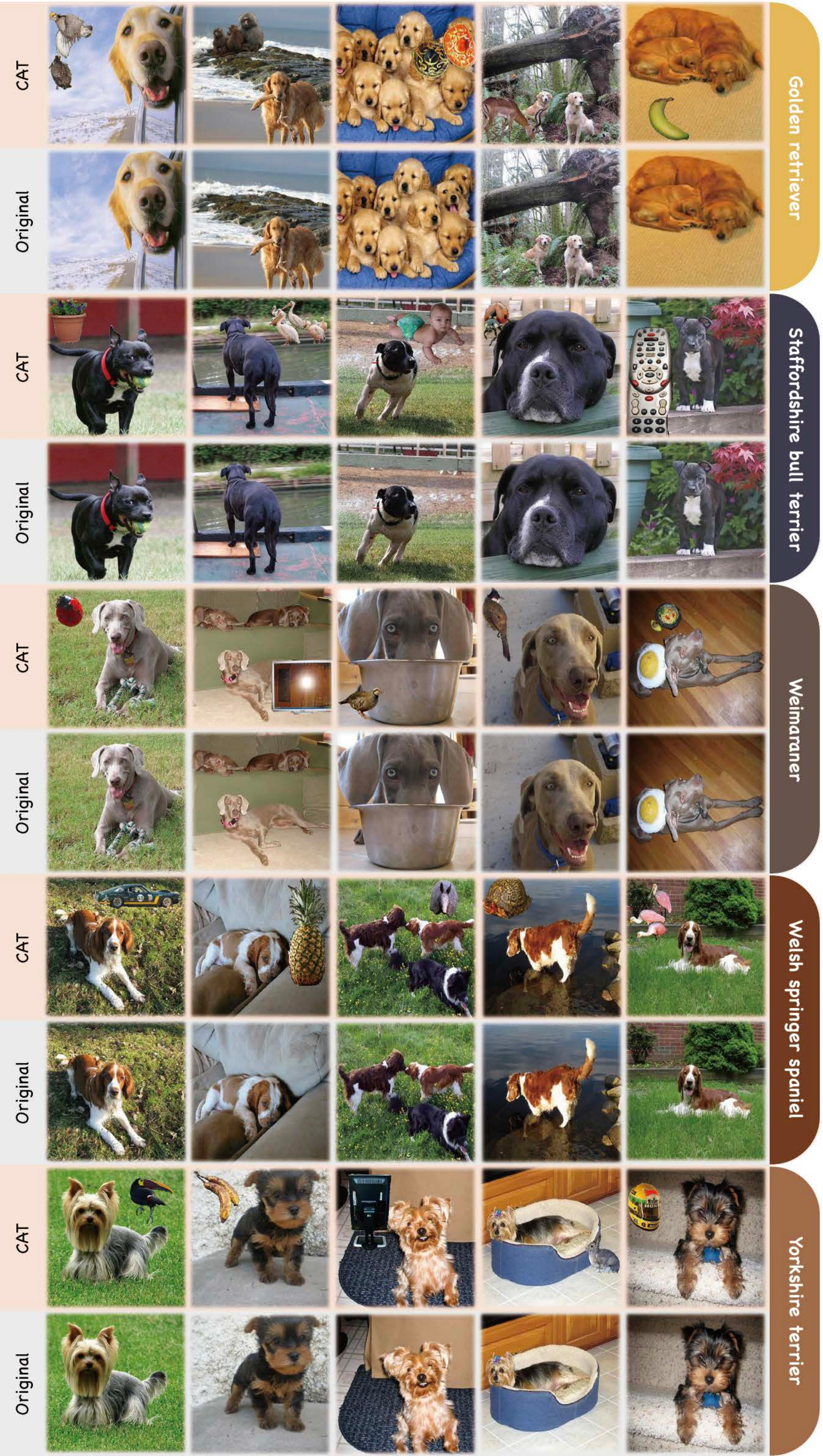}
\end{center}
\vspace{-0.3cm}
\caption{Examples of the ``semantic branches" in our dataset. Best viewed in color and zoomed-in for details.}
\label{fig:species}
\vspace{-0.5cm}
\end{figure*}

Along the backbones of the ``semantic tree", such as \textit{dog}, \textit{cat}, and \textit{bird}, our dataset has extended many ``branches" that contain sophisticated species meanings. An example of the ``branches" for \textit{dog} is shown in Figure \ref{fig:species}. Specifically, instead of regarding and aggregating every \textit{dog}-related samples into a single group, we sincerely maintain their basic semantic hierarchy and divide them into different groups, i.e., \textit{golden retriever}, \textit{staffordshire bull terrier}, \textit{weimaraner}, \textit{welsh springer spaniel}, and \textit{yorkshire terrier}. Our motives here are twofold. First of all, many current one-stage models rely too much on extracting high-level semantic information through pre-training and mapping this information into soft attention to achieve the extraction of co-salient signals. This solves the problem of ``where to look" to a certain extent, but it is also restricted by the quality of pre-training and cannot make reasonable judgments about examples that have not been seen in the pre-training dataset. The second reason is that many potential application scenarios, such as online surveillance, have high requirements for the accuracy of identifying and distinguishing objects from the same roots. For example, continuously identifying the co-salient signal of a specific car from a surveillance video of the street scene requires the model to have the ability to distinguish it from other vehicles. Based on the above reasons, we construct these ``semantic branches". We hope that future models can consider both high-level semantic information and low-level appearance to achieve correct detection and segmentation of co-salient objects, especially those that have never been seen in the pre-training.

\begin{figure}[t]
\begin{center}
\includegraphics[width=0.474\textwidth]{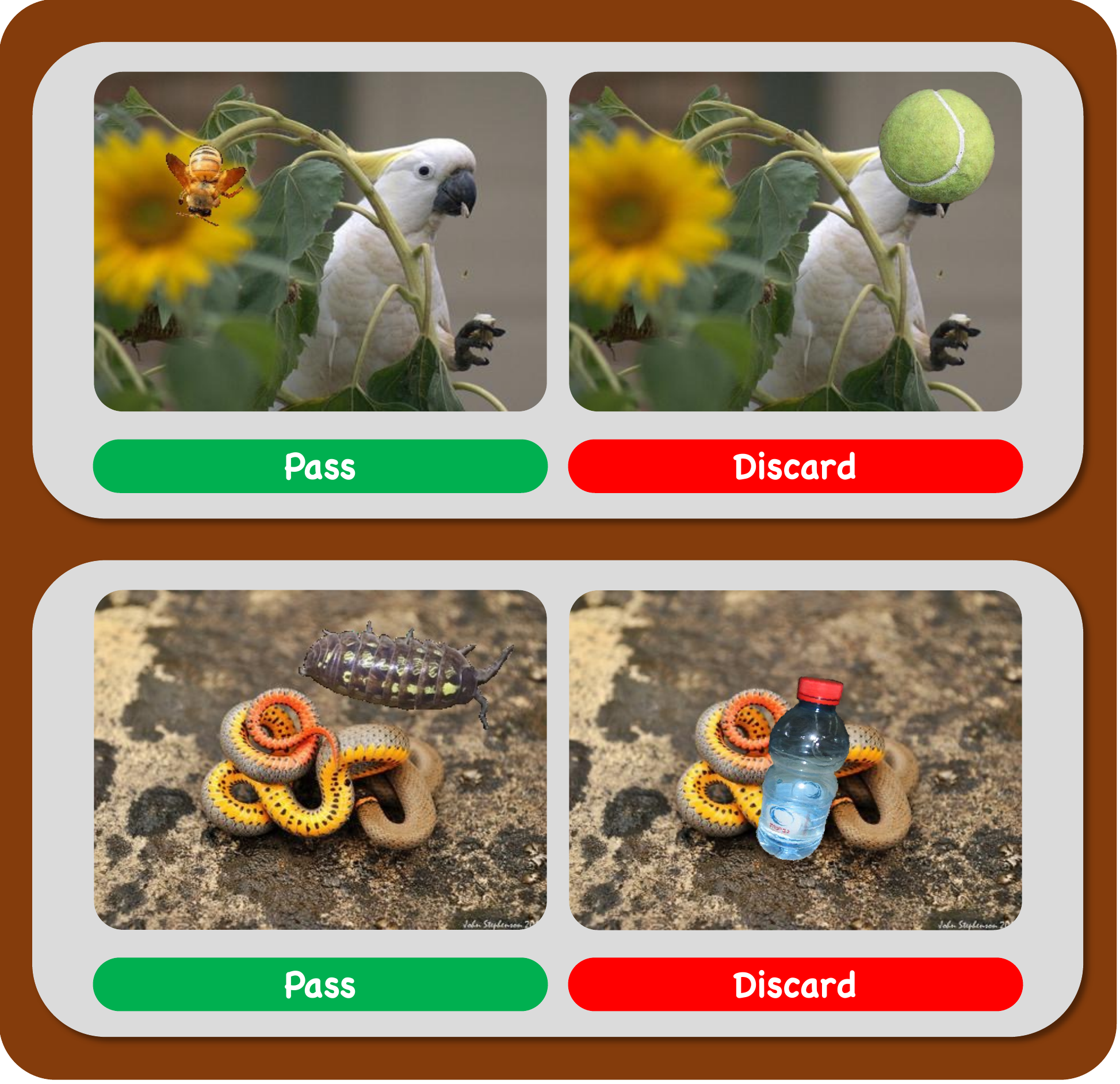}
\end{center}
\vspace{-0.3cm}
\caption{Quality control examples. Images with severe occlusions are discarded.}
\label{fig:quality_control}
\vspace{-0.cm}
\end{figure}

\subsection{Quality Control}

As mentioned in Section \textcolor{blue}{3.3} (Constructing the Dataset) of the main body, to ensure the high-quality of the dataset, we re-sampled some unsatisfactory examples (around 10$\%$). Here we give some examples of our selection criteria. As shown in Figure \ref{fig:quality_control}, objects with slender or meandering shapes, such as creatures like \textit{birds}, \textit{centipedes}, and \textit{snakes}, and items like \textit{hammers}, \textit{hatchets}, and \textit{nails}, can be occluded by objects pasted into the images, resulting in the loss of corresponding semantics. Therefore, we pick these examples out from the candidate pool and re-sample them via GCP.

\subsection{Complete Group Patterns}

\begin{figure*}[!ht]
\begin{center}
\includegraphics[width=0.952\textwidth]{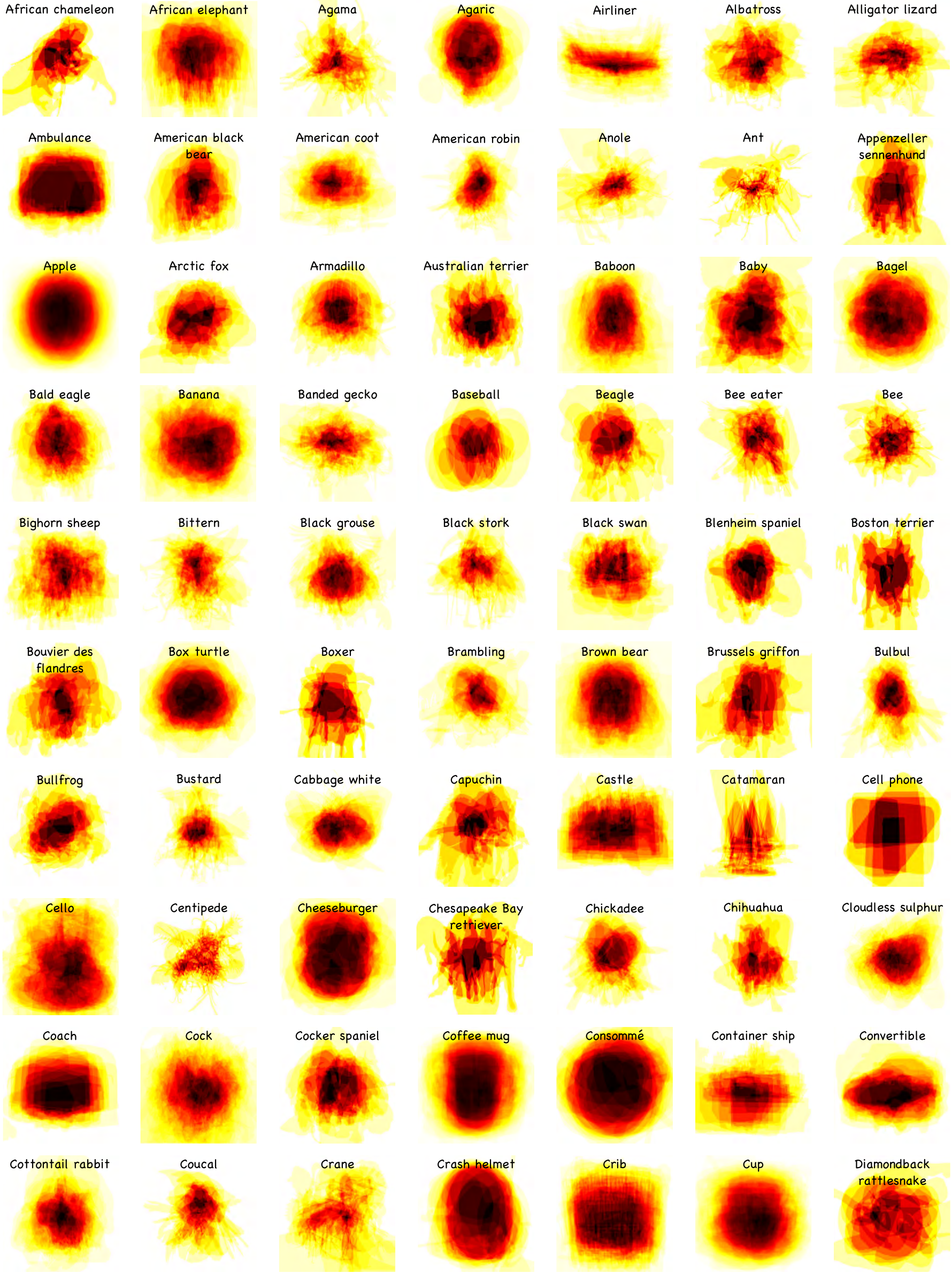}
\end{center}
\vspace{-0.2cm}
\caption{Patterns for the 280 groups in our dataset (1$\sim$70). Best viewed in color and zoomed-in for details.}
\label{fig:patterns_1}
\vspace{-0.6cm}
\end{figure*}

\begin{figure*}[!ht]
\begin{center}
\includegraphics[width=0.952\textwidth]{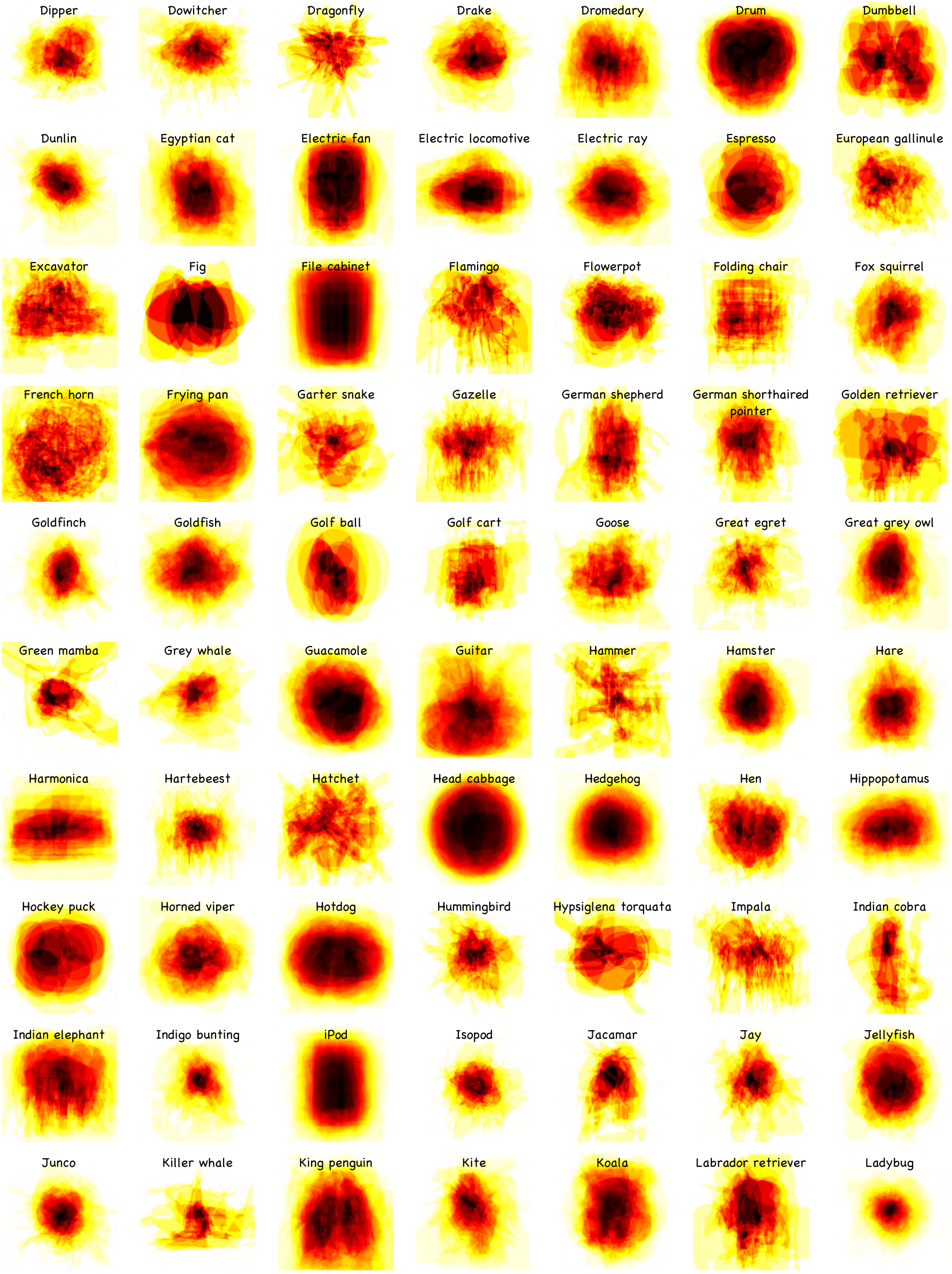}
\end{center}
\vspace{-0.2cm}
\caption{Patterns for the 280 groups in our dataset (71$\sim$140). Best viewed in color and zoomed-in for details.}
\label{fig:patterns_2}
\vspace{-0.6cm}
\end{figure*}

\begin{figure*}[!ht]
\begin{center}
\includegraphics[width=0.952\textwidth]{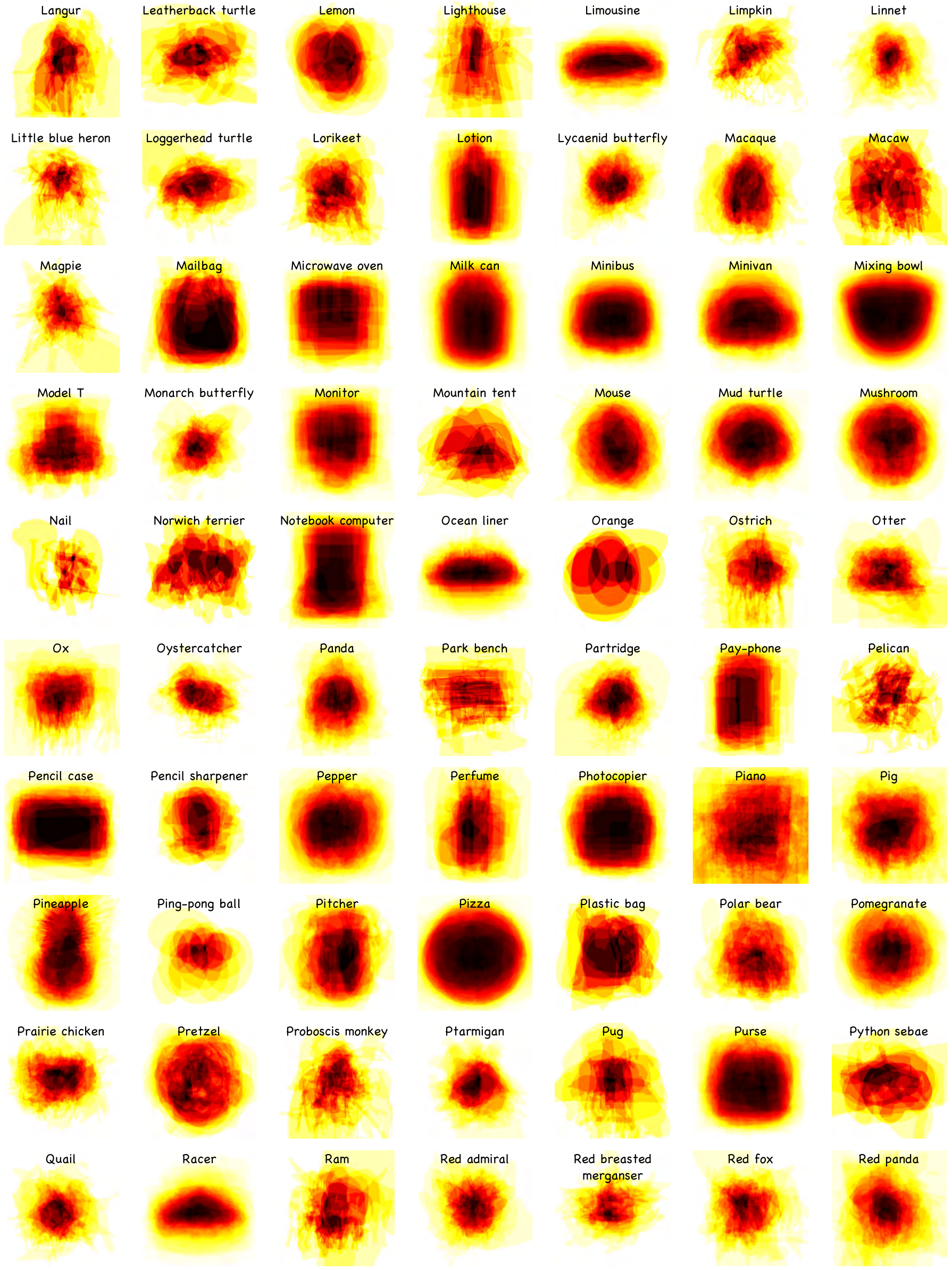}
\end{center}
\vspace{-0.2cm}
\caption{Patterns for the 280 groups in our dataset (141$\sim$210). Best viewed in color and zoomed-in for details.}
\label{fig:patterns_3}
\vspace{-0.6cm}
\end{figure*}

\begin{figure*}[!ht]
\begin{center}
\includegraphics[width=0.952\textwidth]{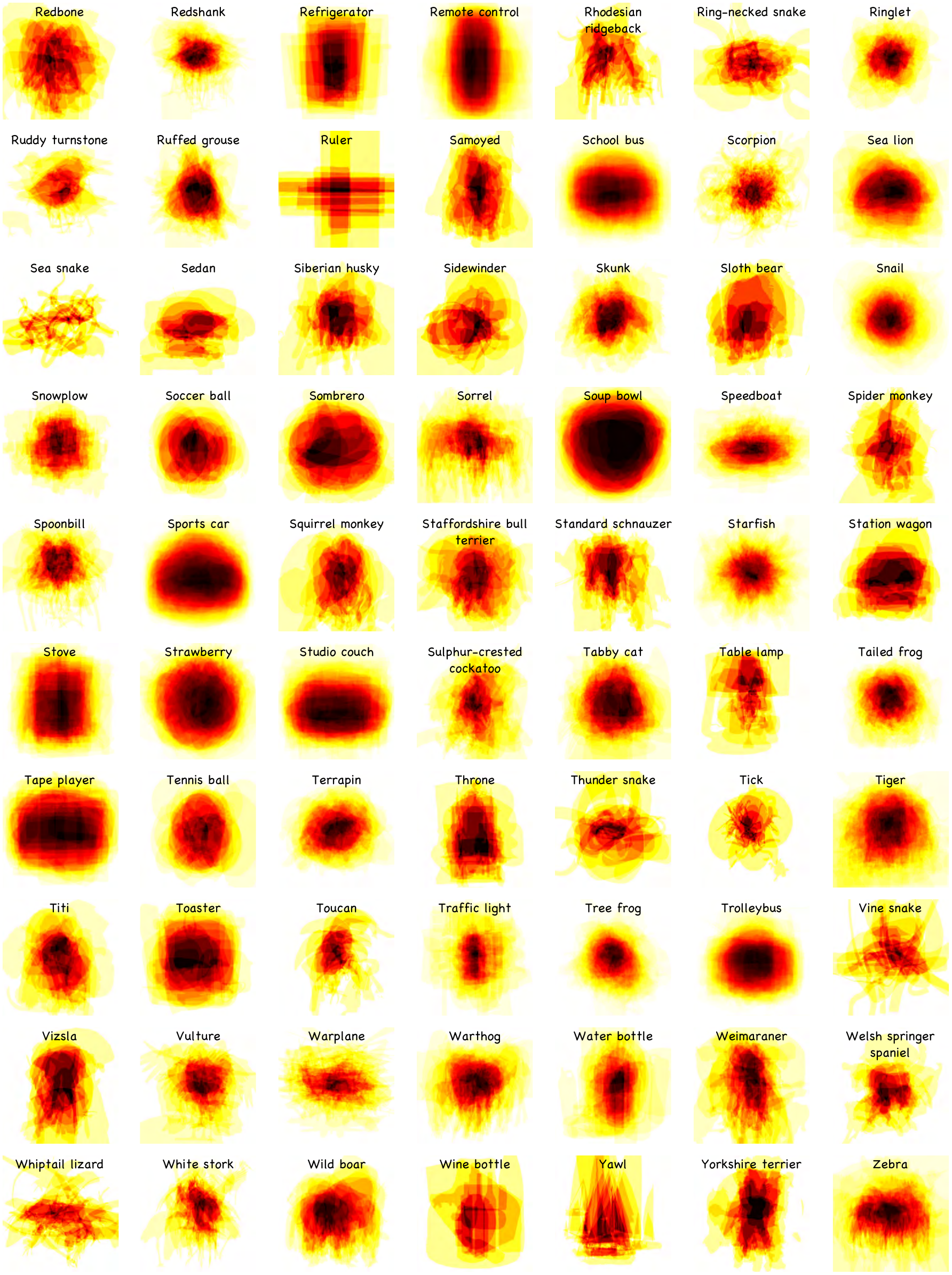}
\end{center}
\vspace{-0.22cm}
\caption{Patterns for the 280 groups in our dataset (211$\sim$280). Best viewed in color and zoomed-in for details.}
\label{fig:patterns_4}
\vspace{-0.6cm}
\end{figure*}

As mentioned in Section \textcolor{blue}{3.3} (Constructing the Dataset) of the main body, the patterns (calculated by averaging overlapping masks) of semantic groups in our dataset tend to have shape-biased appearances, which is consistent with the idea of saliency \cite{Saliency}. This is also the major difference between co-saliency detection \cite{Model-CoEGNet} and image co-segmentation \cite{Survey-Co-Segmentation}. Here we show the complete group patterns for all 280 semantic groups in our dataset in Figures \ref{fig:patterns_1}, \ref{fig:patterns_2}, \ref{fig:patterns_3}, and \ref{fig:patterns_4}.

\subsection{Summary}
After supplementing more statistical and technical details, we hereby summarize the features of our dataset as follows.

\begin{itemize}
    \item \textbf{Scale.} CAT aims to provide comprehensive coverage for co-saliency detection in the deep learning era. The current version of our dataset consists of 33,500 images with high-quality annotations, making it the largest specialized co-saliency detection dataset so far.
    \item \textbf{Diversity.} CAT organizes all image samples in a dense and diverse semantic hierarchy. Rooted from 15 superclasses, the 280 semantic groups in our dataset are like branches, which together construct a ``data tree" covering a large number of categories.
    \item \textbf{Uniqueness.} CAT is constructed with the goal that images should have complex visual contexts with multiple foreground objects. To achieve this, we introduce the idea of counterfactual training and propose a GCP procedure to synthesize training samples without the need for labor-intensive pixel-level labeling.
    \item \textbf{Quality.} We would like to offer a clean dataset with diverse visual contexts and accurate annotations. Therefore, we have carefully considered the technical requirements in each step, formulated corresponding criteria, and strictly followed them. The experimental results also confirmed the reliability of our dataset.
\end{itemize}

\clearpage

\clearpage

\clearpage
\section{Experimental Details}
\label{sec:evaluation}

In this section, we further supplement some detailed information for our benchmark experiment and extensive qualitative results for a more comprehensive visual comparison.

\subsection{Evaluation Metrics}
As mentioned in Section \textcolor{blue}{4.1} (Implementation Details) of the main body, we follow the convention by adopting the following four metrics to evaluate model performances in our benchmark experiment: mean absolute error (\textit{MAE}) \cite{Metrics01}, F-measure \cite{Metrics02}, E-measure \cite{Metrics03}, and structure measure ($S_{\alpha}$) \cite{Metrics04}. Denote  $\mathcal{S}\in\mathbb{R}^{w\times h}$ and  $\mathcal{G}\in\mathbb{R}^{w\times h}$ as the predicted saliency map and the ground-truth with width $w$ and height $h$, respectively. Here we further detail the characteristics of these metrics.

\begin{itemize}
    \item \textbf{Precision \& Recall}\\The basic building blocks for evaluations. Denote true-positive, true-negative, false-positive, and false-negative as \textit{TP}, \textit{TN}, \textit{FP}, and \textit{FN}, respectively. Precision ($\mathcal{P}$) and recall ($\mathcal{R}$) are calculated based on the binarized saliency map $\mathcal{S}$ and the ground-truth $\mathcal{G}$ as follows:
    \begin{equation}
        \mathcal{P} = \frac{\textit{TP}}{\textit{TP}+\textit{FP}}, ~~~~~~~~
        \mathcal{R} = \frac{\textit{TP}}{\textit{TP}+\textit{FN}}.
    \label{pr}
    \end{equation}
    \item \textbf{MAE} \cite{Metrics01}\\Metric for measuring the average pixel-wise absolute error between saliency map $\mathcal{S}$ and ground-truth $\mathcal{G}$:
    \begin{equation}
        \textit{MAE} = \frac{1}{w\times h}\sum_{i=1}^{w}\sum_{j=1}^{h}\mid\mathcal{S}(i,j)-\mathcal{G}(i,j)\mid, 
    \label{mae}
    \end{equation}
    where $i$ and $j$ are the indexes.
    \item \textbf{F-measure} \cite{Metrics02}\\Metric for weighting precision ($\mathcal{P}$) and recall ($\mathcal{R}$), which is defined by the weighted harmonic mean as follows:
    \begin{equation}
        F_{\beta} = \frac{(1+\beta^{2})\mathcal{P}\times\mathcal{R}}{\beta^{2}\mathcal{P}+\mathcal{R}},
    \label{f_measure}
    \end{equation}
    where coefficient $\beta^{2}$ is empirically set as $0.3$ to emphasize more on precision ($\mathcal{P}$) than recall ($\mathcal{R}$). Here we consider both the maximum value $F_{max}$ and the mean value $F_{avg}$, which is calculated by averaging the binarized saliency map $\mathcal{S}$ with an adaptive threshold.
    \item \textbf{E-measure} \cite{Metrics03}\\Metric for measuring both the pixel-level matching and the image-level statistics for the binarized saliency map $\mathcal{S}$, i.e.,
    \begin{equation}
        Q_{\phi} = \frac{1}{w\times h}\sum_{i=1}^{w}\sum_{j=1}^{h}\phi(i,j),
    \label{e_measure}
    \end{equation}
    where $\phi$ denotes the enhanced alignment matrix \cite{Metrics03}, which reflects the correlation between normalized saliency map $\mathcal{S}$ and ground-truth $\mathcal{G}$.
    \item \textbf{Structure measure} \cite{Metrics04}\\Metric for measuring the structural similarity between saliency map $\mathcal{S}$ and ground-truth $\mathcal{G}$. Denote $S_{o}$ and $S_{r}$ as the object-aware similarity and the region-aware similarity \cite{Metrics04}, respectively. The structure measure is calculated as follows:
    \begin{equation}
        S_{\alpha} = (1-\alpha)\times S_{o}+\alpha\times S_{r},
    \label{s_measure}
    \end{equation}
    where coefficient $\alpha$ is empirically set as $0.5$.
\end{itemize}

It worth noting that the aforementioned metrics are from saliency detection \cite{Survey-SOD-DL} and are widely adopted by co-saliency detection works \cite{Model-CoEGNet}. Specifically, \textit{MAE} \cite{Metrics01} and F-measure \cite{Metrics02} focus on measuring pixel-level errors, while E-measure \cite{Metrics03} and structure measure \cite{Metrics04} consider structural similarities. Although F-measure \cite{Metrics02} fails to consider \textit{TN} pixels, \textit{MAE} \cite{Metrics01} makes up for this vacancy. 

Overall, these metrics are complementary to each other. Together, they build a comprehensive assessment for evaluating model performances. However, so far there is no evaluation metric specifically designed for co-saliency detection. We hope that researchers in this area can design a suitable one that considers both co-saliency and group-wise effect in the near future.

\subsection{Details of Loss Functions}
As mentioned in Section \textcolor{blue}{4.1} (Implementation Details) of the main body, we use five state-of-the-art models for the benchmark experiments, i.e., \textit{PoolNet} \cite{Model-PoolNet}, \textit{EGNet} \cite{Model-EGNet}, \textit{ICNet} \cite{Model-ICNet}, \textit{GICD} \cite{Model-GICD}, and \textit{GCoNet} \cite{Model-GCoNet}. 

Specifically, \textit{PoolNet} \cite{Model-PoolNet} adopts the \textit{standard binary cross-entropy loss} for saliency detection and the \textit{balanced binary cross-entropy loss} \cite{Loss-Func-Balanced-Cross-Entropy} for edge detection. \textit{EGNet} \cite{Model-EGNet} adopts the \textit{standard binary cross-entropy loss} for both saliency detection and edge detection. \textit{ICNet} \cite{Model-ICNet} and \textit{GICD} \cite{Model-GICD} adopt the \textit{IoU loss} \cite{Loss-Func-IoU} and the \textit{soft IoU loss} \cite{Loss-Func-Soft-IoU-1,Loss-Func-Soft-IoU-2} for co-saliency detection, respectively. For \textit{GCoNet} \cite{Model-GCoNet}, three loss functions are aggregated, i.e., \textit{soft IoU loss} \cite{Loss-Func-Soft-IoU-1,Loss-Func-Soft-IoU-2}, \textit{focal loss} \cite{Loss-Focal}, and \textit{cross-entropy loss}. Several weights are applied to balance these three losses during training.

\subsection{Details of Training Configurations}
The detailed training configurations for models mentioned in Section \textcolor{blue}{4.1} (Implementation Details) of the main body are summarized as follows.

\begin{itemize}
    \item \textbf{PoolNet} \cite{Model-PoolNet}\\The Res2Net \cite{Res2Net} version of \textit{PoolNet} \cite{Model-PoolNet} is adopted in our experiments. The model is trained via the Adam optimizer \cite{Adam} with learning rate $=4e-5$ and weight decay $=5e-4$. We run this model for 50 epochs and evaluate the checkpoints from the last three epochs.
    \item \textbf{EGNet} \cite{Model-EGNet}\\For this model, we adopt the one with the VGG backbone in our experiments. We follow the default setting by using the Adam optimizer \cite{Adam} and assigning the learning rate as $2e-5$, weight decay as $5e-4$, and loss weight as $1$. Before the training, the edge maps for the training samples are prepared via the corresponding mask annotations of each dataset. We train \textit{EGNet} \cite{Model-EGNet} for 30 epochs and divide the learning rate by 10 after 15 epochs. For each run, we evaluate the last three checkpoints of this model and report the average results.
    \item \textbf{ICNet} \cite{Model-ICNet}\\For this model, we adopt both the VGG \cite{VGG} and ResNet \cite{ResNet} versions of \textit{EGNet} \cite{Model-EGNet} to prepare the single-image-saliency-maps (SISMs). The Adam optimizer \cite{Adam} with learning rate $=8e-6$ is adopted. In total, we run this model for 80 epochs and evaluate the checkpoints from the last three epochs.
    \item \textbf{GICD} \cite{Model-GICD}\\We keep the original training paradigm of \textit{GICD} \cite{Model-GICD} in our experiments, which randomly selects at most 20 samples from each image group to extract the group saliency. Adam optimizer \cite{Adam} is adopted. The initial learning rate is set as $2e-4$. We train this model for 100 epochs and reduce the learning to $80\%$ every 20 epochs. The \textit{CoSal2015} dataset \cite{Dataset-CoSal2015} is used as the validation set for picking the best possible checkpoints.
    \item \textbf{GCoNet} \cite{Model-GCoNet}\\We adopt the default model architecture of \textit{GCoNet} \cite{Model-GCoNet}, which is a VGG-based \cite{VGG} feature pyramid network. Similar to \textit{GICD} \cite{Model-GICD}, the group saliency is calculated by selecting and aggregating group samples. For each training episode, 16 samples from two different groups are picked. This model also uses the Adam optimizer \cite{Adam} with learning rate $=2e-4$ and a total number of 50 epochs are trained. We use \textit{CoSal2015} \cite{Dataset-CoSal2015} as the validation set for picking the best possible checkpoints.
\end{itemize}
\subsection{More Qualitative Results}
\begin{figure*}[t]
\begin{center}
\includegraphics[width=0.999\textwidth]{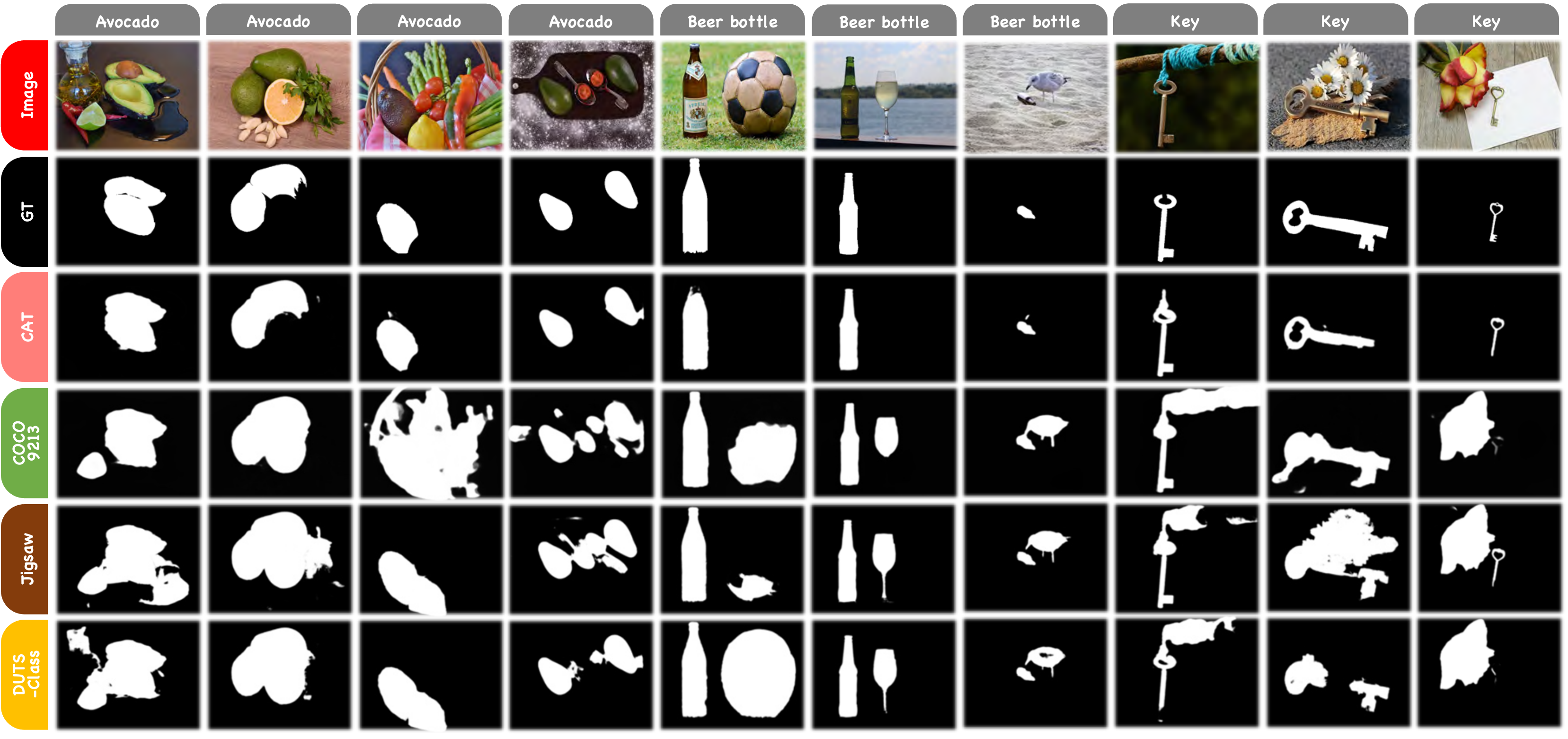}
\end{center}
\vspace{-0.3cm}
\caption{Qualitative comparison among different training datasets. Model: \textit{ICNet} \cite{Model-ICNet} (SISMs generated by pretrained \textit{EGNet} with ResNet \cite{ResNet} backbone). Evaluation dataset: \textit{CoCA} \cite{Model-GICD}. Semantic groups from left to right: \textit{avocado}, \textit{beer bottle}, and \textit{key}.}
\label{fig:comparison_icnet_res_coca_1}
\vspace{-0.0cm}
\end{figure*}

\begin{figure*}[t]
\begin{center}
\includegraphics[width=0.999\textwidth]{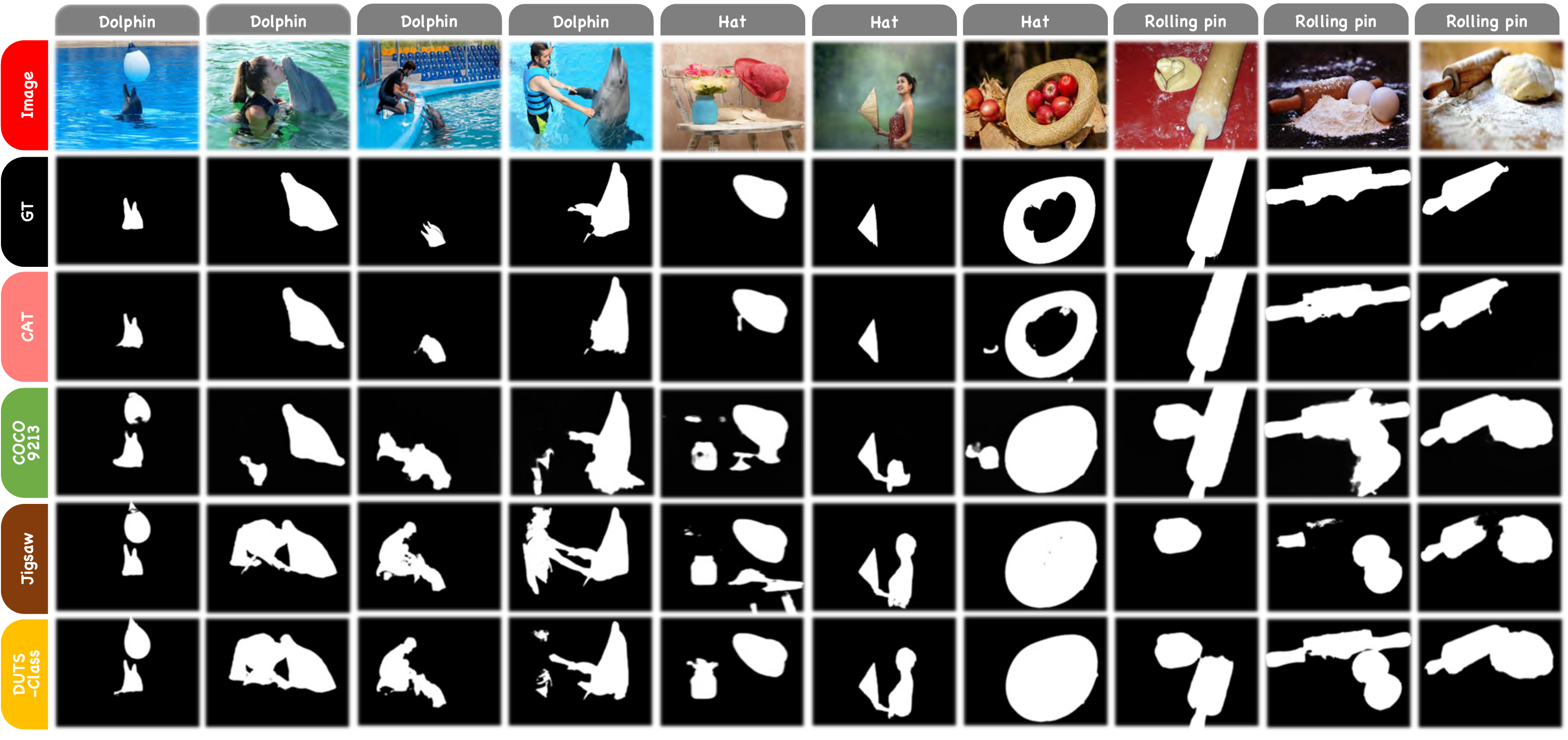}
\end{center}
\vspace{-0.3cm}
\caption{Qualitative comparison among different training datasets. Model: \textit{ICNet} \cite{Model-ICNet} (SISMs generated by pretrained \textit{EGNet} with ResNet \cite{ResNet} backbone). Evaluation dataset: \textit{CoCA} \cite{Model-GICD}. Semantic groups from left to right: \textit{dolphin}, \textit{hat}, and \textit{rolling pin}.}
\label{fig:comparison_icnet_res_coca_2}
\vspace{-0.cm}
\end{figure*}

\begin{figure*}[t]
\begin{center}
\includegraphics[width=0.999\textwidth]{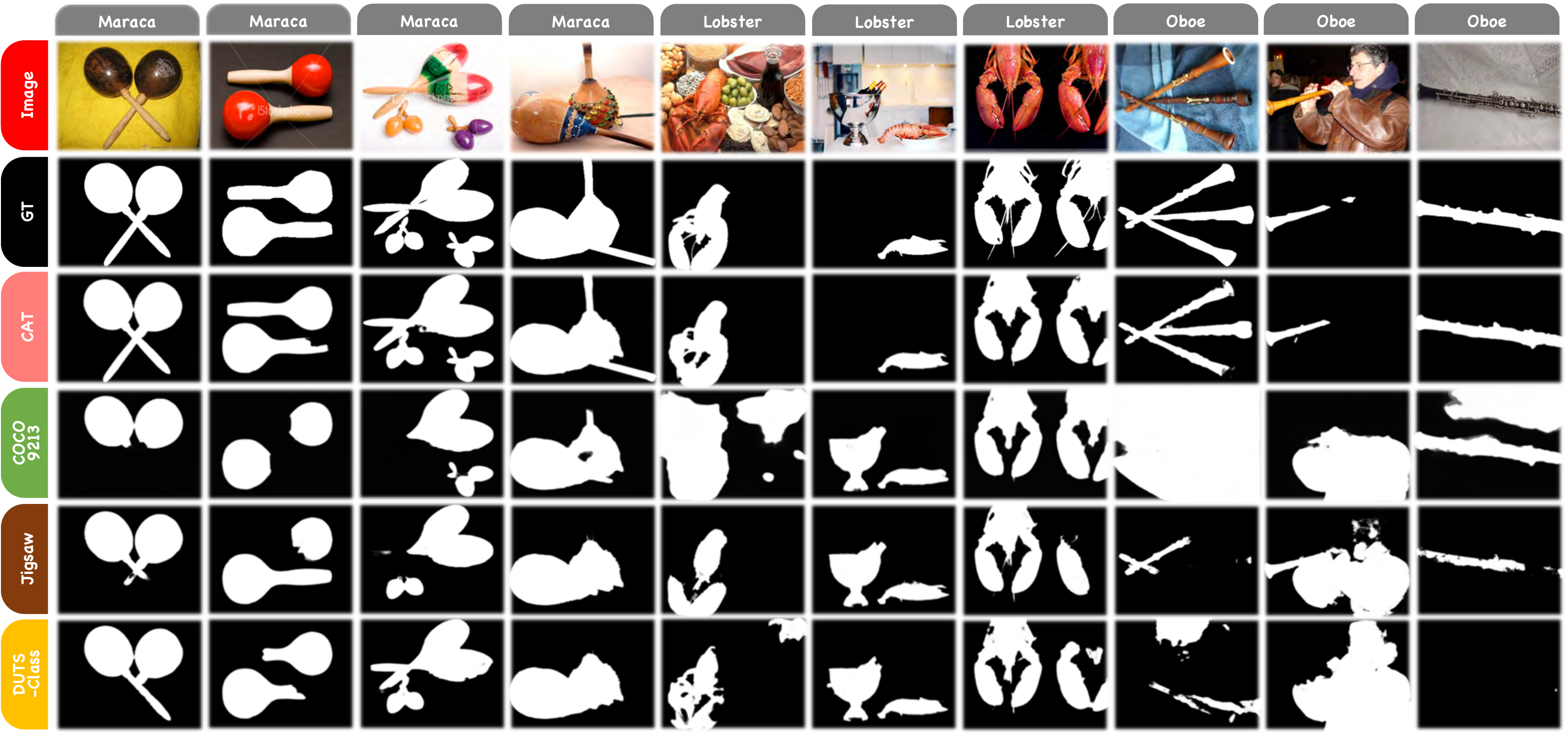}
\end{center}
\vspace{-0.3cm}
\caption{Qualitative comparison among different training datasets. Model: \textit{ICNet} \cite{Model-ICNet} (SISMs generated by pretrained \textit{EGNet} with ResNet \cite{ResNet} backbone). Evaluation dataset: \textit{CoSOD3k} \cite{Dataset-CoSOD3k}. Semantic groups from left to right: \textit{maraca}, \textit{lobster}, and \textit{oboe}.}
\label{fig:comparison_icnet_res_cosod3k_1}
\vspace{-0.cm}
\end{figure*}

\begin{figure*}[t]
\begin{center}
\includegraphics[width=0.999\textwidth]{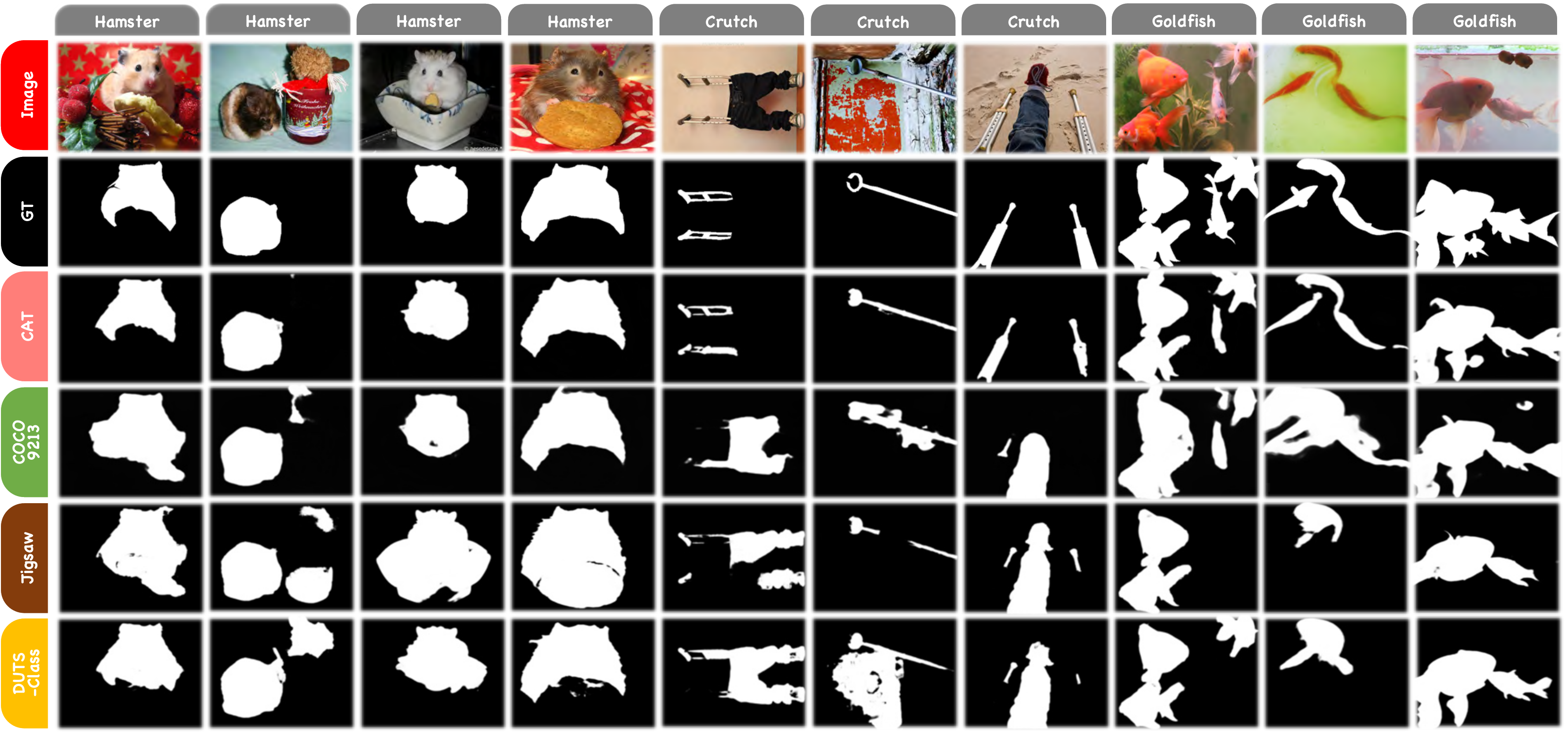}
\end{center}
\vspace{-0.3cm}
\caption{Qualitative comparison among different training datasets. Model: \textit{ICNet} \cite{Model-ICNet} (SISMs generated by pretrained \textit{EGNet} with ResNet \cite{ResNet} backbone). Evaluation dataset: \textit{CoSOD3k} \cite{Dataset-CoSOD3k}. Semantic groups from left to right: \textit{hamster}, \textit{crutch}, and \textit{goldfish}.}
\label{fig:comparison_icnet_res_cosod3k_2}
\vspace{-0.cm}
\end{figure*}

\begin{figure*}[t]
\begin{center}
\includegraphics[width=0.999\textwidth]{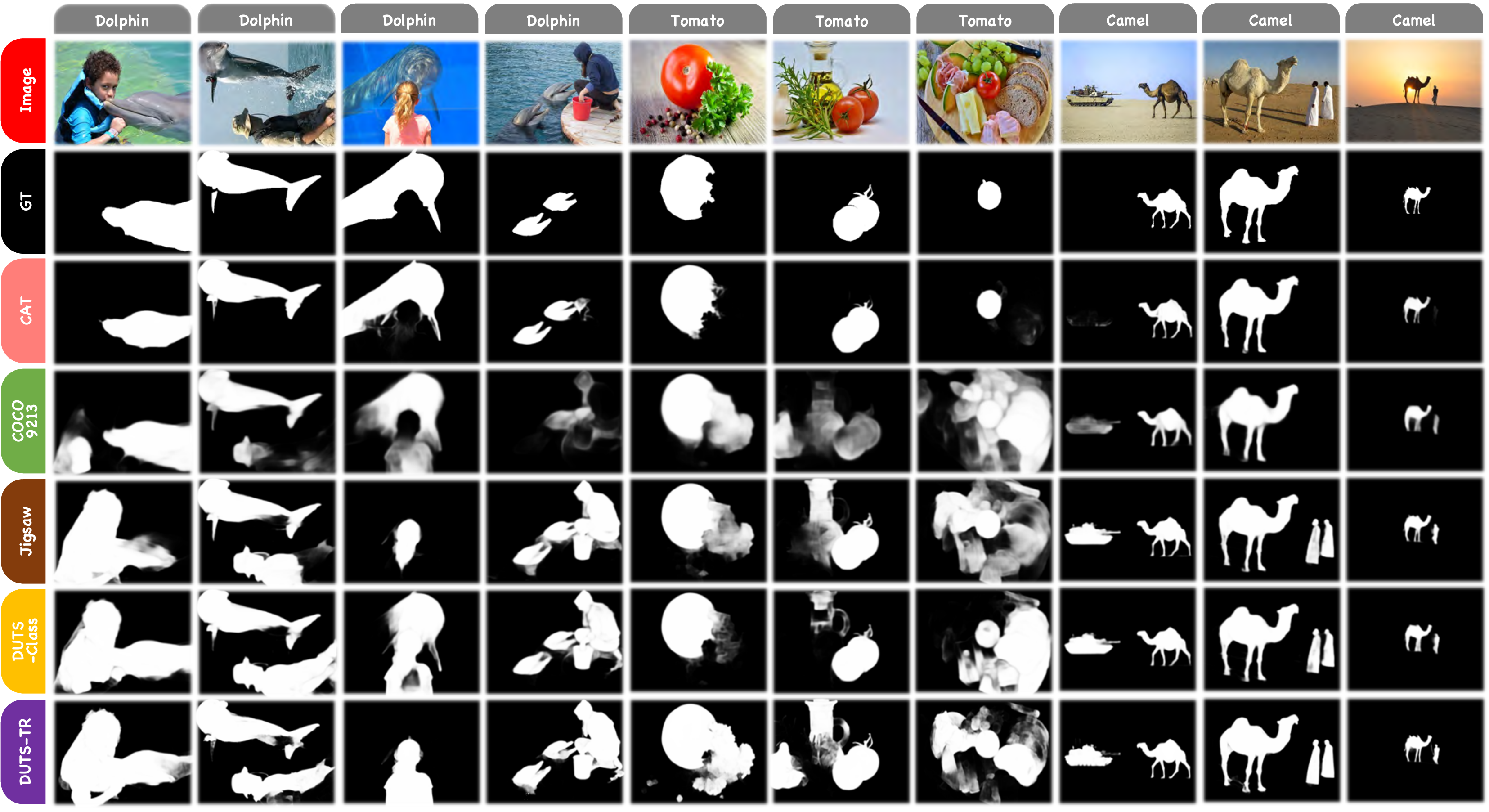}
\end{center}
\vspace{-0.3cm}
\caption{Qualitative comparison among different training datasets. Model: \textit{EGNet} \cite{Model-EGNet}. Evaluation dataset: \textit{CoCA} \cite{Model-GICD}. Semantic groups from left to right: \textit{dolphin}, \textit{tomato}, and \textit{camel}.}
\label{fig:comparison_egnet_coca_1}
\vspace{-0.cm}
\end{figure*}

\begin{figure*}[t]
\begin{center}
\includegraphics[width=0.999\textwidth]{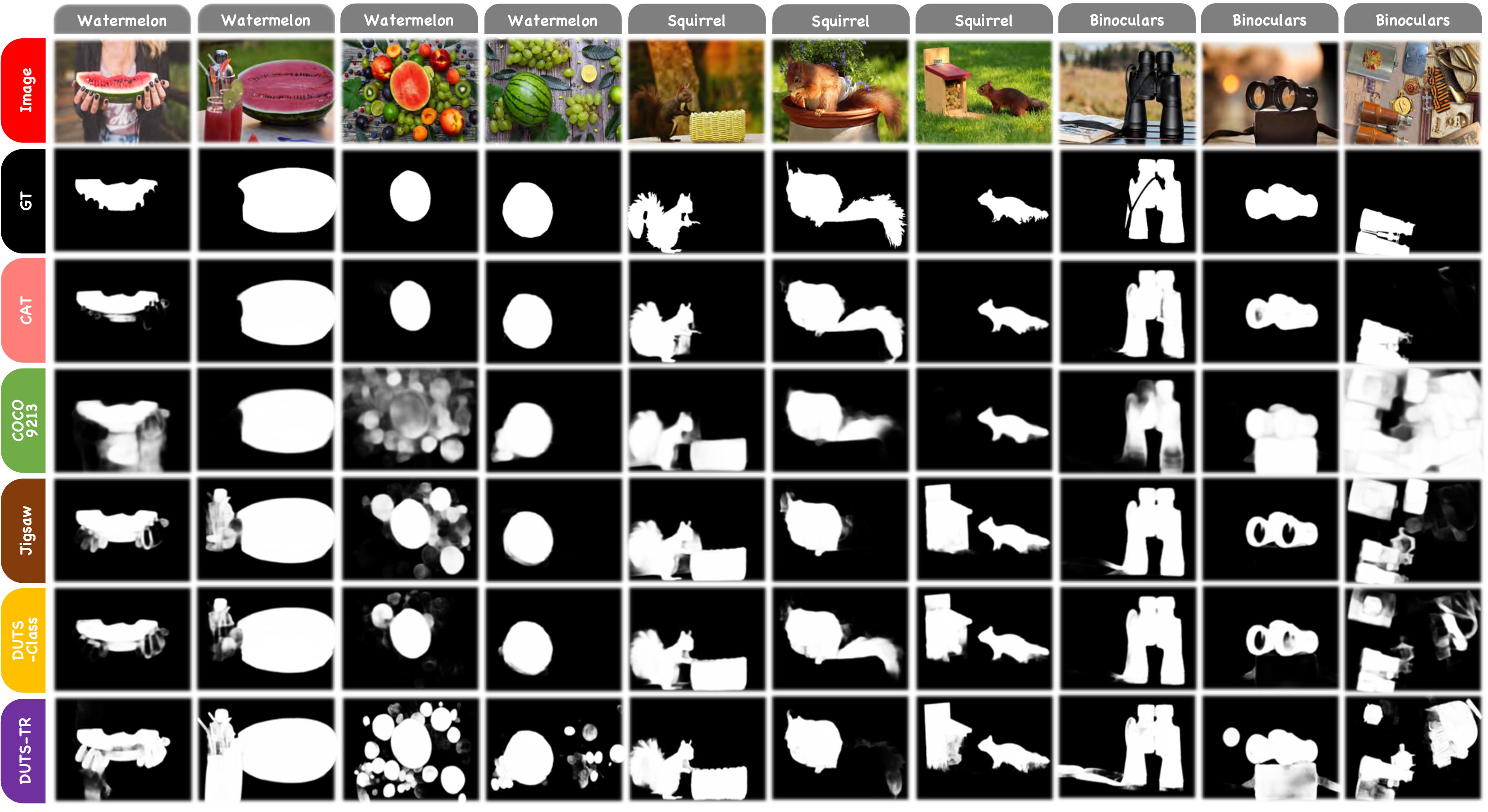}
\end{center}
\vspace{-0.3cm}
\caption{Qualitative comparison among different training datasets. Model: \textit{EGNet} \cite{Model-EGNet}. Evaluation dataset: \textit{CoCA} \cite{Model-GICD}. Semantic groups from left to right: \textit{watermelon}, \textit{squirrel}, and \textit{binoculars}.}
\label{fig:comparison_egnet_coca_2}
\vspace{-0.cm}
\end{figure*}

\begin{figure*}[t]
\begin{center}
\includegraphics[width=0.999\textwidth]{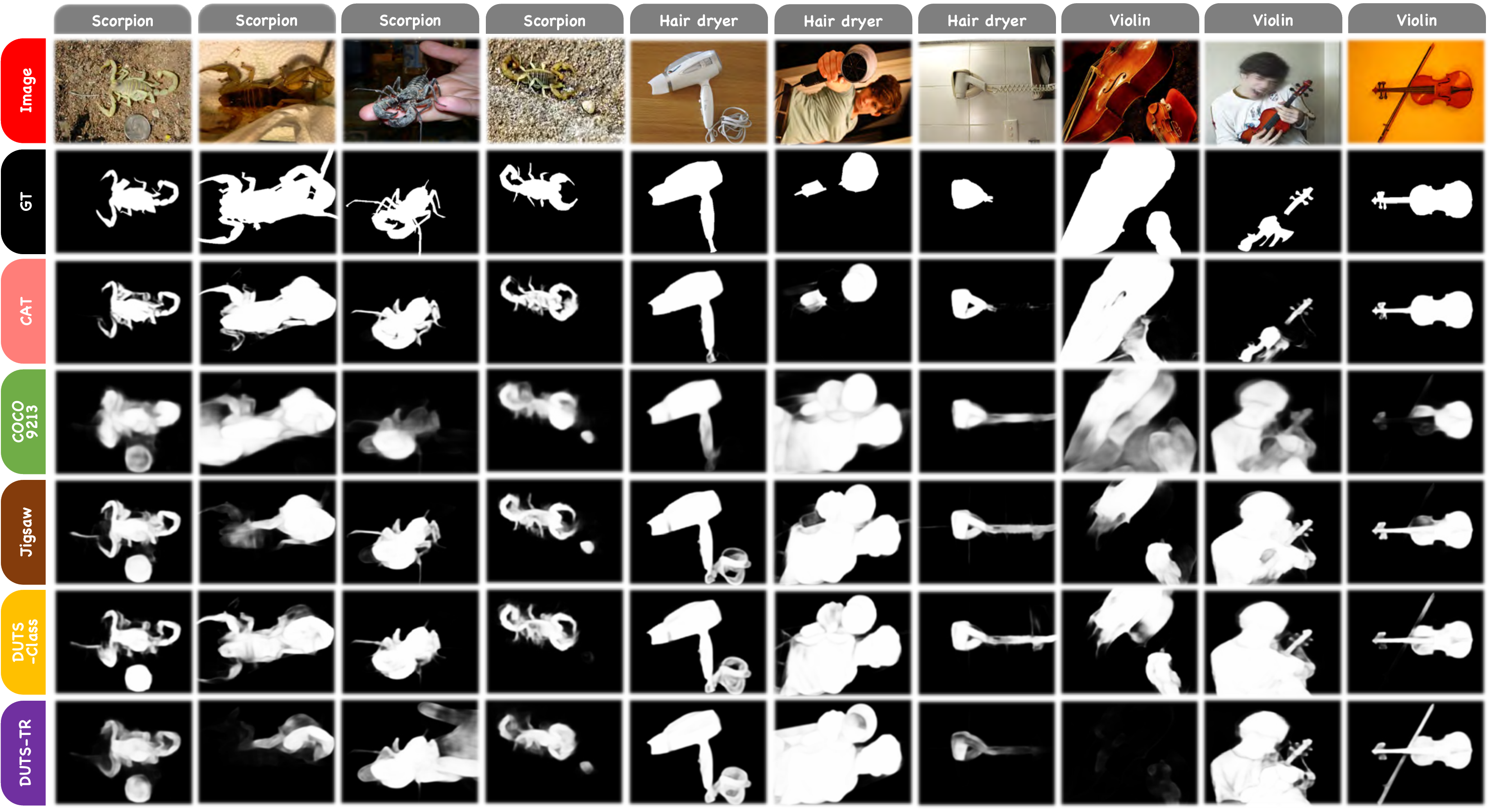}
\end{center}
\vspace{-0.3cm}
\caption{Qualitative comparison among different training datasets. Model: \textit{EGNet} \cite{Model-EGNet}. Evaluation dataset: \textit{CoSOD3k} \cite{Dataset-CoSOD3k}. Semantic groups from left to right: \textit{scorpion}, \textit{hair dryer}, and \textit{violin}.}
\label{fig:comparison_egnet_cosod3k_1}
\vspace{-0.cm}
\end{figure*}

\begin{figure*}[t]
\begin{center}
\includegraphics[width=0.999\textwidth]{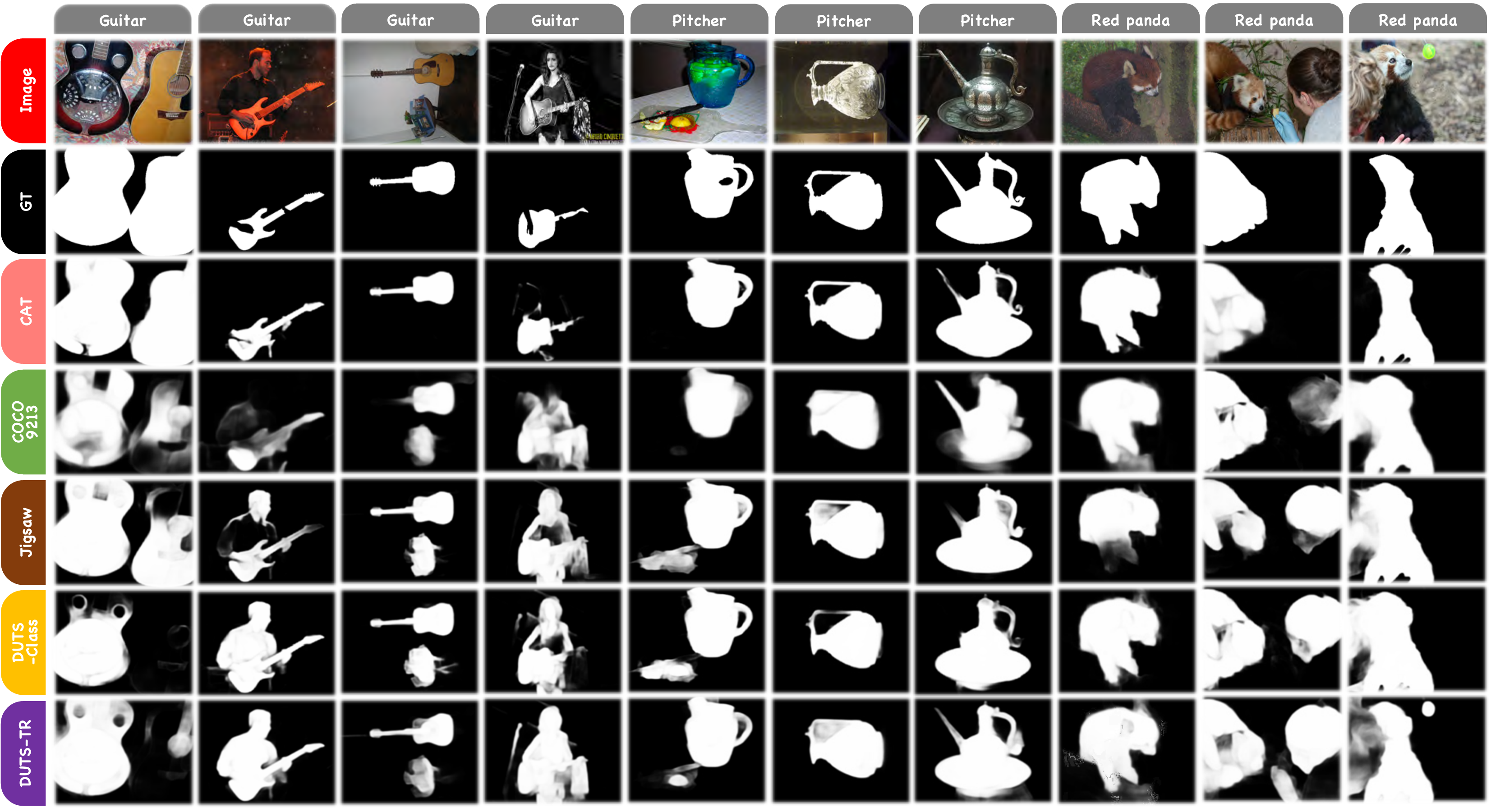}
\end{center}
\vspace{-0.3cm}
\caption{Qualitative comparison among different training datasets. Model: \textit{EGNet} \cite{Model-EGNet}. Evaluation dataset: \textit{CoSOD3k} \cite{Dataset-CoSOD3k}. Semantic groups from left to right: \textit{guitar}, \textit{pitcher}, and \textit{red panda}.}
\label{fig:comparison_egnet_cosod3k_2}
\vspace{-0.0cm}
\end{figure*}

In Section \textcolor{blue}{4.3} (Qualitative Analysis) of the main body, we show some qualitative results of different training datasets on \textit{GICD} \cite{Model-GICD}. To offer a comprehensive and precise evaluation among different competitors, we show more visualization results here. The qualitative results for different datasets trained by \textit{ICNet} \cite{Model-ICNet} are shown in Figures \ref{fig:comparison_icnet_res_coca_1}, \ref{fig:comparison_icnet_res_coca_2}, \ref{fig:comparison_icnet_res_cosod3k_1}, and \ref{fig:comparison_icnet_res_cosod3k_2}. Furthermore, for \textit{EGNet} \cite{Model-EGNet}, the qualitative results are shown in Figures \ref{fig:comparison_egnet_coca_1}, \ref{fig:comparison_egnet_coca_2}, \ref{fig:comparison_egnet_cosod3k_1}, and \ref{fig:comparison_egnet_cosod3k_2}.

\clearpage

\clearpage

\clearpage
\section{Benchmark Study}
\label{sec:benchmark}

In this section, to provide a more comprehensive benchmark study, we review the growth of co-saliency detection in chronological order and then discuss the corresponding works on datasets and models.

\subsection{Chronology}

\begin{figure}[t]
\begin{center}
\includegraphics[width=0.474\textwidth]{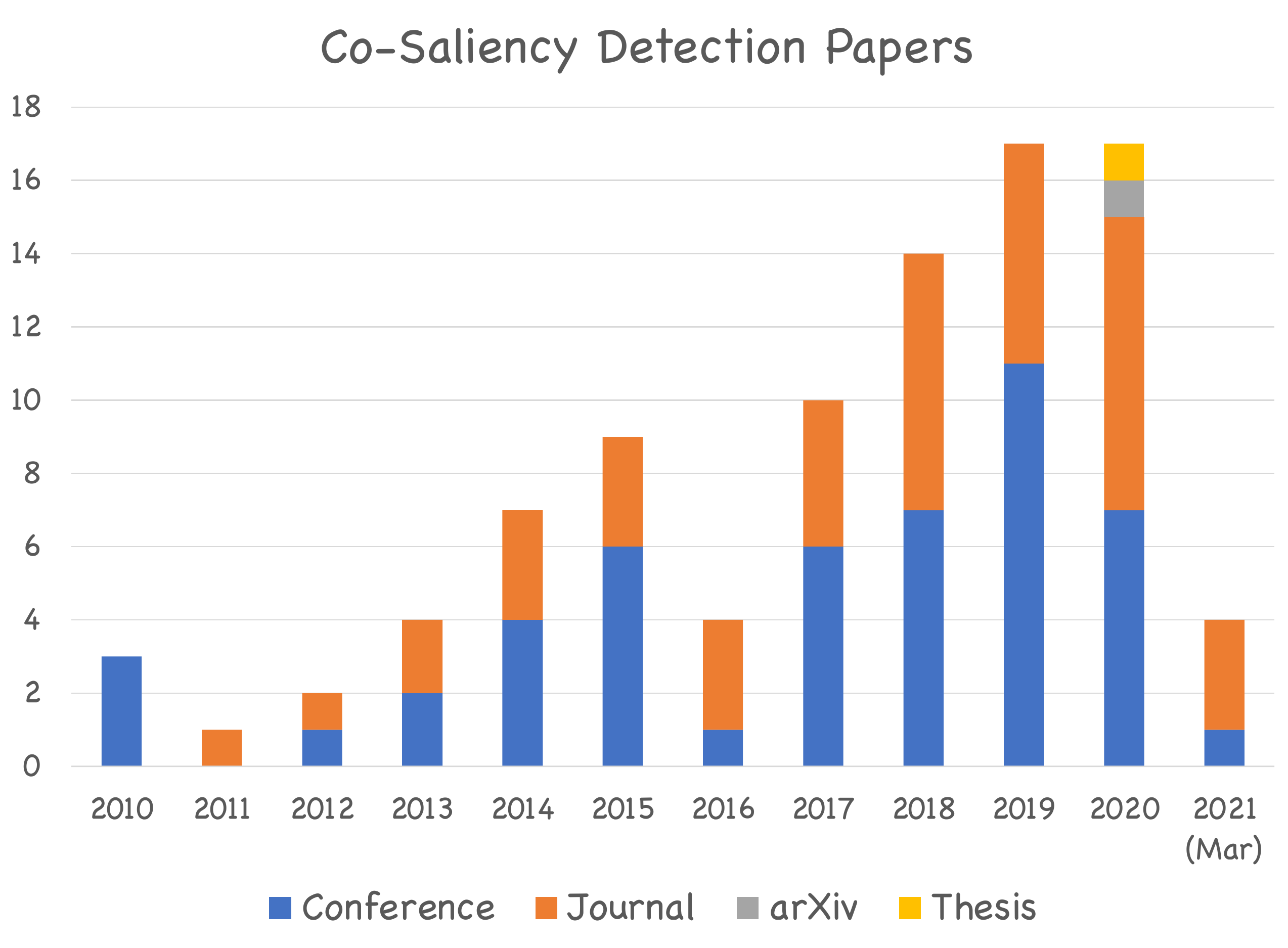}
\end{center}
\vspace{-0.3cm}
\caption{The number of co-saliency detection related papers published each year.}
\label{fig:trend}
\vspace{-0.cm}
\end{figure}

Co-saliency detection is quite young compared to other conventional computer vision tasks like image classification and object detection, but more and more attention has been gained in recent years. As shown in Figure \ref{fig:trend}, there is a clear upward trend for research in co-saliency detection and its related fields.

\begin{figure*}[t]
\begin{center}
\includegraphics[width=0.999\textwidth]{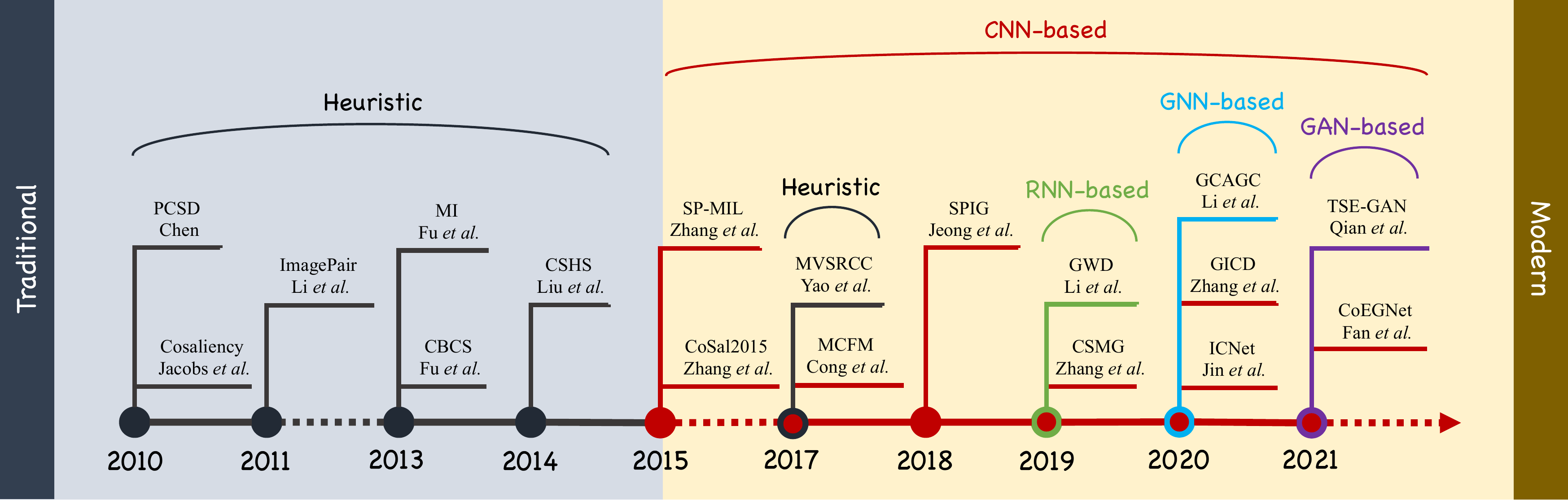}
\end{center}
\vspace{-0.3cm}
\caption{The chronology of co-saliency detection. Representative works for each year are listed and divided by types.}
\label{fig:history}
\vspace{-0.cm}
\end{figure*}

Figure \ref{fig:history} shows the chronological structure of co-saliency detection. Specifically, Like most popular computer vision tasks, the evolution of co-saliency detection can be divided into two stages: traditional and modern. \cite{Model-CoSaliency} is the very beginning of this task. Since then, researchers have begun to pay attention to identify co-saliency signals in images groups. Traditional works \cite{Model-PCSD,Model-CBCS,Model-MI,Model-CSHS} depend on certain heuristics like the color contrast and background prior to extract intra-image and inter-image cues, while modern methods \cite{Model-SP-MIL1,Model-MVSRCC,Model-MCFM} mainly adopt variant network structures to learn patterns from data, such as convolutional neural networks \cite{Model-SPIG,Model-CSMG,Model-GICD,Model-ICNet,Model-CoEGNet,Model-GCoNet}, recurrent neural networks \cite{Model-GWD}, graph neural networks \cite{Model-GCAGC1}, and generative adversarial networks \cite{Model-TSE-GAN}.

\subsection{Dataset Zoo}

\begin{figure}[t]
\begin{center}
\includegraphics[width=0.474\textwidth]{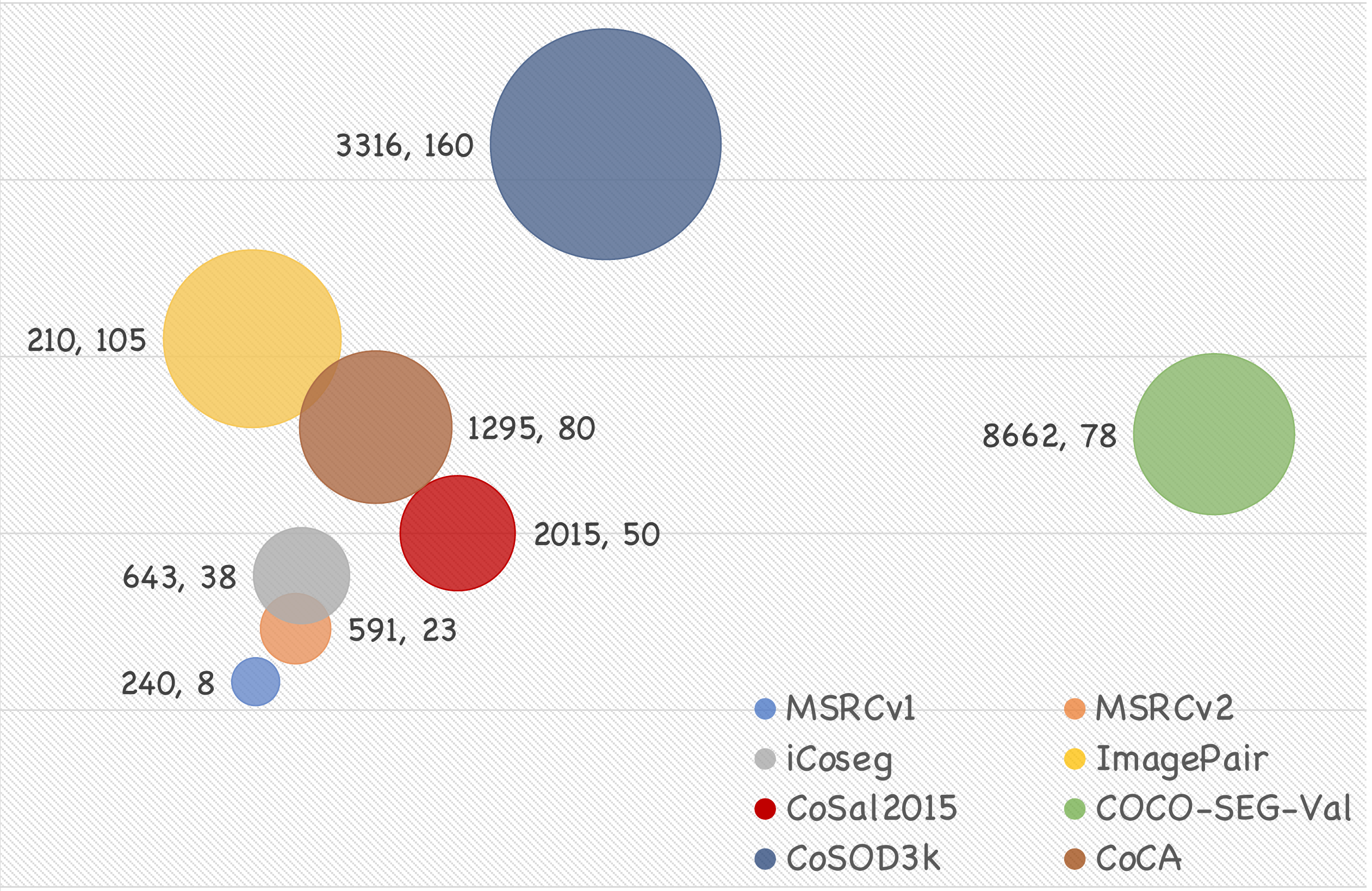}
\end{center}
\vspace{-0.3cm}
\caption{Bubble chart for evaluation datasets. Horizontal axis: number of images; Vertical axis: number of groups. The bigger the bubble, the more semantic groups it contains.}
\label{fig:bubble_eval}
\vspace{-0.cm}
\end{figure}

In Section \textcolor{blue}{2} (Related Work) of the main body, we give a comprehensive analysis for the 10 training datasets in co-saliency detection. Here we adopt a similar procedure and show the characteristics of the datasets used to evaluate model performances, i.e., \textit{MSRC} \cite{Dataset-MSRC}, \textit{iCoseg} \cite{Dataset-iCoseg}, \textit{ImagePair} \cite{Dataset-ImagePair}, \textit{CoSal2015} \cite{Dataset-CoSal2015}, \textit{WISD} \cite{Dataset-WISD}, \textit{COCO-SEG-Val} \cite{Dataset-COCO-SEG}, \textit{CoSOD3k} \cite{Dataset-CoSOD3k}, and \textit{CoCA} \cite{Model-GICD}. The statistics of these evaluation datasets are shown in Table \ref{tab:evaluation_datasets} and Figure \ref{fig:bubble_eval}.

Specifically, small-scale datasets like \textit{MSRC} \cite{Dataset-MSRC} and \textit{iCoseg} \cite{Dataset-iCoseg} are popular among works from early years only. \textit{COCO-SEG-Val} \cite{Dataset-COCO-SEG} is the largest one among all nine datasets, which contains 8,662 images from 78 categories. However, as we mentioned before, the mask annotations for semantic segmentation datasets like \textit{COCO-SEG} \cite{Dataset-COCO-SEG} are coarse compared with both saliency and co-saliency detection datasets; thus, it may not be accurate enough for model evaluations. \textit{CoSal2015} \cite{Dataset-CoSal2015} is a more reliable choice in recent years and has been adopted as both the training and evaluation dataset for different works. However, the context and diversity of this dataset are getting insufficient with the deepening of related research. Recent works \cite{Model-GCoNet,Model-CoEGNet} have achieved nearly 0.85 in F-measure and structure measure and 0.9 in E-measure on \textit{CoSal2015} \cite{Dataset-CoSal2015}. The two datasets proposed last year (i.e., \textit{CoSOD3k} \cite{Dataset-CoSOD3k} and \textit{CoCA} \cite{Model-GICD}) improved the quality of visual contexts, increased the dataset scales, and introduced more diversity. Therefore, we adopt them in our benchmark experiment to evaluate performances.

\begin{table*}[t]
\centering
\scalebox{0.91}{
\begin{tabular}{p{0.3cm}<{\centering}|p{2.4cm}<{\centering}|p{1.cm}<{\centering}|p{1.cm}<{\centering}p{1.cm}<{\centering}p{1.cm}<{\centering}p{1.cm}<{\centering}p{1.cm}<{\centering}|p{0.7cm}<{\centering}p{0.7cm}<{\centering}p{0.7cm}<{\centering}p{0.7cm}<{\centering}|p{2.2cm}<{\centering}}
\toprule[2pt]
\textbf{\#} & \textbf{Dataset}  & \textbf{Year} & \textbf{\#Img.} & \textbf{\#Cat.} & \textbf{\#Avg.} & \textbf{\#Max.} & \textbf{\#Min.} & \textbf{Mul.} & \textbf{Sal.} & \textbf{Larg.} & \textbf{H.Q.} & \textbf{Inputs}
\\ \hline\hline
1 & MSRCv1 & 2005 & 240 & 8 & 30.0 & 30 & 30 & \xmark & \cmark & \xmark & \xmark & Group images
\\
2 & MSRCv2 & 2005 & 591 & 23 & 25.7 & 34 & 21 & \xmark & \cmark & \xmark & \xmark & Group images
\\
3 &iCoseg & 2010 & 643 & 38 & 16.9 & 41 & 4 & \xmark & \cmark & \xmark & \cmark & Group images
\\
4 & ImagePair & 2011 & 210 & 105 & 2.0 & 2 & 2 & \xmark & \cmark & \xmark & \xmark & Pair images
\\
5 & CoSal2015 & 2015 & 2,015 & 50 & 40.3 & 52 & 26 & \cmark & \cmark & \cmark & \cmark & Group images
\\
6 & WISD & 2019 & 2,019 & - & - & - & - & \cmark & - & \cmark & - & Group images
\\
7 & COCO-SEG-Val & 2019 & 8,662 & 78 & 111.1 & 2,107 & 6 & \cmark & \xmark & \cmark & \xmark & Group images
\\
8 & CoSOD3k & 2020 & 3,316 & 160 & 20.7 & 30 & 4 & \cmark & \cmark & \cmark & \cmark & Group images
\\
9 & CoCA & 2020 & 1,295 & 80 & 16.2 & 40 & 8 & \cmark & \cmark & \cmark & \cmark & Group images
\\
\bottomrule[2pt]
\end{tabular}}
\vspace{-0.2cm}
\caption{Datasets used for evaluating co-saliency detection models. {\#Img.}: Number of images; {\#Cat.}: Number of categories; {\#Avg.}: Average number of images per category; {\#Max.}: Maximum number of images per group; {\#Min.}: Minimum number of images per group. {Mul.}: Whether contains multiple foreground objects or not; {Sal.}: Whether maintains saliency or not; {Larg.}: Whether large-scale (more than 1k images) or not; {H.Q.}: Whether has high-quality annotations or not. ``-" denotes ``not available".}
\label{tab:evaluation_datasets}
\vspace{-0.cm}
\end{table*}

\begin{table*}[t]
\centering
\scalebox{0.99}{
\begin{tabular}{c|c|c|cccccccc}
\toprule[2pt]
\textbf{\#} & \textbf{Model} & \textbf{Venue} & {\textit{MAE} $\downarrow$} & {$F_{max}$ $\uparrow$} & {$F_{avg}$ $\uparrow$} & {$E_{max}$ $\uparrow$} & {$E_{avg}$ $\uparrow$} & {$S_{\alpha}$ $\uparrow$}
\\ \hline\hline
1 & {CBCS} \cite{Model-CBCS} & TIP'13 & 0.228 & 0.468 & 0.322 & 0.633 & 0.509 & 0.529
\\
2 & {CSHS} \cite{Model-CSHS} & SPL'14 & 0.309 & 0.484 & - & 0.656 & - & 0.563
\\
3 & {ESMG} \cite{Model-ESMG} & SPL'14 & 0.239 & 0.364 & - & - & 0.615 & 0.532
\\
4 & {CODR} \cite{Model-CODR} & SPL'15 & 0.229 & 0.458 & - & - & 0.645 & 0.630
\\
5 & {GW} \cite{Model-GW} & IJCAI'17 & 0.147 & 0.649 & - & 0.777 & - & 0.716
\\
6 & {PoolNet} \cite{Model-PoolNet} & CVPR'19 & 0.171 & 0.561 & 0.528 & 0.751 & 0.689 & 0.636
\\
7 & {CSMG} \cite{Model-CSMG} & CVPR'19 & 0.141 & 0.729 & 0.596 & 0.821 & 0.675 & 0.727
\\
8 & {BASNet} \cite{Model-BASNet} & CVPR'19 & 0.122 & 0.696 & - & - & 0.791 & 0.753
\\
9 & {EGNet$^{\bigstar}$} \cite{Model-EGNet} & ICCV'19 & 0.088 & 0.745 & 0.735 & 0.840 & 0.830 & 0.791
\\
10 & {SCRN} \cite{Model-SCRN} & ICCV'19 & 0.118 & 0.717 & - & - & 0.806 & 0.773
\\
11 & {RCAN} \cite{Model-RCAN} & IJCAI'19 & 0.130 & 0.688 & - & 0.808 & - & 0.744
\\
12 & {SSNM} \cite{Model-SSNM} & AAAI'20 & 0.120 & 0.675 & - & - & 0.756 & 0.726
\\
13 & {F3Net} \cite{Model-F3Net} & AAAI'20 & 0.114 & 0.717 & - & 0.802 & - & 0.772
\\
14 & {GCAGC} \cite{Model-GCAGC1} & CVPR'20 & 0.100 & 0.740 & - & - & 0.831 & 0.778
\\
15 & {MINet} \cite{Model-MINet} & CVPR'20 & 0.122 & 0.707 & - & 0.782 & - & 0.754
\\
16 & {ICNet-V$^{\bigstar}$} \cite{Model-ICNet} & NeurIPS'20 & \textcolor{orange}{0.068} & \textcolor{blue}{0.791} & \textcolor{orange}{0.782} & 0.870 & \textcolor{blue}{0.866} & 0.811
\\
17 & {ICNet-R$^{\bigstar}$} \cite{Model-ICNet} & NeurIPS'20 & \textcolor{blue}{0.067} & \textcolor{red}{0.792} & \textcolor{blue}{0.783} & \textcolor{orange}{0.872} & \textcolor{red}{0.867} & \textcolor{blue}{0.817}
\\
18 & {CoADNet} \cite{Model-CoADNet} & NeurIPS'20 & 0.078 & 0.786 & - & \textcolor{blue}{0.874} & - & \textcolor{red}{0.822}
\\
19 & {GICD$^{\bigstar}$} \cite{Model-GICD} & ECCV'20 & 0.070 & 0.786 & 0.779 & \textcolor{red}{0.886} & \textcolor{orange}{0.863} & 0.812
\\
20 & {CoEGNet} \cite{Model-CoEGNet} & TPAMI'21 & 0.092 & 0.736 & - & 0.825 & - & 0.762
\\
21 & {DeepACG} \cite{Model-DeepACG} & CVPR'21 & 0.089 & 0.756 & - & - & 0.838 & 0.792
\\
22 & {GCoNet$^{\bigstar}$} \cite{Model-GCoNet} & CVPR'21 & \textcolor{red}{0.065} & \textcolor{orange}{0.789} & \textcolor{red}{0.784} & 0.865 & 0.862 & \textcolor{orange}{0.813}
\\
\bottomrule[2pt]
\end{tabular}}
\vspace{-0.1cm}
\caption{Benchmarking results on \textit{CoSOD3k} \cite{Dataset-CoSOD3k}. Symbols $\uparrow$ and $\downarrow$ denote the metrics which are the higher and the lower the better, respectively. Models trained by our CAT are highlighted with superscript ${\bigstar}$. ``-V" and ``-R" denote models with VGG and ResNet backbones, respectively. The \textcolor{red}{best}, \textcolor{blue}{second best}, and \textcolor{orange}{third best} results for each metric are highlighted with \textcolor{red}{red}, \textcolor{blue}{blue}, and \textcolor{orange}{orange} colors.}
\label{tab:benchmark1}
\vspace{-0.cm}
\end{table*}

\begin{table*}[t]
\centering
\scalebox{0.99}{
\begin{tabular}{c|c|c|cccccccc}
\toprule[2pt]
\textbf{\#} & \textbf{Model} & \textbf{Venue} & {\textit{MAE} $\downarrow$} & {$F_{max}$ $\uparrow$} & {$F_{avg}$ $\uparrow$} & {$E_{max}$ $\uparrow$} & {$E_{avg}$ $\uparrow$} & {$S_{\alpha}$ $\uparrow$}
\\ \hline\hline
1 & {CBCS} \cite{Model-CBCS} & TIP'13 & 0.180 & 0.313 & 0.230 & 0.641 & 0.531 & 0.523
\\
2 & {CSHS} \cite{Model-CSHS} & SPL'14 & - & - & - & - & - & -
\\
3 & {ESMG} \cite{Model-ESMG} & SPL'14 & - & - & - & - & - & -
\\
4 & {CODR} \cite{Model-CODR} & SPL'15 & - & - & - & - & - & -
\\
5 & {GW} \cite{Model-GW} & IJCAI'17 & 0.166 & 0.408 & - & 0.701 & - & 0.602
\\
6 & {PoolNet} \cite{Model-PoolNet} & CVPR'19 & 0.181 & 0.346 & 0.323 & 0.707 & 0.625 & 0.561
\\
7 & {CSMG} \cite{Model-CSMG} & CVPR'19 & 0.124 & 0.508 & - & \textcolor{orange}{0.735} & - & 0.632
\\
8 & {BASNet} \cite{Model-BASNet} & CVPR'19 & 0.195 & 0.397 & - & - & 0.623 & 0.589
\\
9 & {EGNet$^{\bigstar}$} \cite{Model-EGNet} & ICCV'19 & 0.143 & 0.449 & 0.439 & 0.691 & 0.678 & 0.626
\\
10 & {SCRN} \cite{Model-SCRN} & ICCV'19 & 0.166 & 0.416 & - & - & 0.658 & 0.610
\\
11 & {RCAN} \cite{Model-RCAN} & IJCAI'19 & 0.160 & 0.422 & - & 0.702 & - & 0.616
\\
12 & {SSNM} \cite{Model-SSNM} & AAAI'20 & 0.116 & 0.482 & - & - & 0.741 & 0.628
\\
13 & {F3Net} \cite{Model-F3Net} & AAAI'20 & 0.178 & 0.437 & - & 0.678 & - & 0.614
\\
14 & {GCAGC} \cite{Model-GCAGC1} & CVPR'20 & 0.111 & 0.523 & - & - & \textcolor{blue}{0.754} & 0.669
\\
15 & {MINet} \cite{Model-MINet} & CVPR'20 & 0.221 & 0.387 & - & 0.634 & - & 0.550
\\
16 & {ICNet-V$^{\bigstar}$} \cite{Model-ICNet} & NeurIPS'20 & 0.108 & 0.529 & \textcolor{orange}{0.522} & \textcolor{blue}{0.744} & 0.734 & \textcolor{orange}{0.673}
\\
17 & {ICNet-R$^{\bigstar}$} \cite{Model-ICNet} & NeurIPS'20 & \textcolor{blue}{0.101} & \textcolor{blue}{0.533} & \textcolor{blue}{0.526} & \textcolor{red}{0.751} & \textcolor{orange}{0.743} & \textcolor{blue}{0.678}
\\
18 & {CoADNet} \cite{Model-CoADNet} & NeurIPS'20 & - & - & - & - & - & -
\\
19 & {GICD$^{\bigstar}$} \cite{Model-GICD} & ECCV'20 & 0.116 & 0.523 & 0.516 & 0.727 & 0.716 & 0.668
\\
20 & {CoEGNet} \cite{Model-CoEGNet} & TPAMI'21 & 0.104 & 0.499 & - & 0.717 & - & 0.616
\\
21 & {DeepACG} \cite{Model-DeepACG} & CVPR'21 & \textcolor{orange}{0.102} & \textcolor{red}{0.552} & - & - & \textcolor{red}{0.771} & \textcolor{red}{0.688}
\\
22 & {GCoNet$^{\bigstar}$} \cite{Model-GCoNet} & CVPR'21 & \textcolor{red}{0.089} & \textcolor{orange}{0.531} & \textcolor{red}{0.528} & 0.730 & 0.728 & \textcolor{orange}{0.673}
\\
\bottomrule[2pt]
\end{tabular}}
\vspace{-0.1cm}
\caption{Benchmarking results on \textit{CoCA} \cite{Model-GICD}. Symbols $\uparrow$ and $\downarrow$ denote the metrics which are the higher and the lower the better, respectively. Models trained by our CAT are highlighted with superscript ${\bigstar}$. ``-V" and ``-R" denote models with VGG and ResNet backbones, respectively. The \textcolor{red}{best}, \textcolor{blue}{second best}, and \textcolor{orange}{third best} results for each metric are highlighted with \textcolor{red}{red}, \textcolor{blue}{blue}, and \textcolor{orange}{orange} colors.}
\label{tab:benchmark2}
\vspace{-0.cm}
\end{table*}

\subsection{Model Zoo}
In Section \textcolor{blue}{4} (Benchmark Experiment) of the main body, we have evaluated five state-of-the-art co-saliency detection models in the benchmark experiment part of the main body, i.e., \textit{PoolNet} \cite{Model-PoolNet}, \textit{EGNet} \cite{Model-EGNet}, \textit{ICNet} \cite{Model-ICNet}, \textit{GICD} \cite{Model-GICD}, and \textit{GCoNet} \cite{Model-GCoNet}. A more comprehensive benchmarking study of 22 models are shown in Tables \ref{tab:benchmark1} and \ref{tab:benchmark2}. For the continuous growth of this task, we will keep updating the evaluation results of more models to the benchmark.


{\small
\bibliography{aaai22}
}

\end{document}